# Language models guide symbolic equation discovery by controlling search

Zikai Xie[1,2,†,*], Wenmei Li[1,†], Man Luo[1], Jun Jiang[1,2,*] and Linjiang Chen[1,2,3,*]

[1] State Key Laboratory of Precision and Intelligent Chemistry, Hefei National Research Center for Physical Sciences at the Microscale, School of Chemistry and Materials Science, University of Science and Technology of China, Hefei, China.
[2] Center for Scientific Intelligence Innovation, Hefei, China
[3] School of Chemistry, School of Computer Science, University of Birmingham, Birmingham, UK
[†] These authors contributed equally: Zikai Xie, Wenmei Li
* Corresponding authors: linjiangchen@ustc.edu.cn, jiangj1@ustc.edu.cn, zikaix@ustc.edu.cn

## Abstract
Scientific equation discovery must combine broad domain priors with strict numerical testing. Symbolic regression supplies numerical grounding but faces a combinatorial search space, whereas many language-model systems ask the model to propose or select formulas directly. We test a different division of labour. We compare role specifications in which the language model acts as equation author, candidate decider or search controller, alongside end-to-end language-model and purely numerical baselines. In the controller setting we propose here, implemented as LLM-PySR, language models specify variables, operators, transformations and search depth; symbolic regression enumerates and fits expressions; and deterministic metrics govern retention. Across 74 AI-Feynman equations and seven complex formula-recovery tasks, search control achieved the strongest observed balance of accuracy, complexity, stability and cost. On an independent battery dataset, LLM-PySR identified a compact piecewise-linear relation between early voltage-curve displacement and cycle life. The results suggest that language models should shape hypothesis exploration rather than decide which equations survive.

## Main
Inferring compact, mechanistically meaningful equations from data has long been a hallmark of scientific discovery, and is also a central goal of scientific machine learning.[1] Unlike conventional machine-learning models that primarily optimize predictive accuracy, scientific equation discovery seeks compact analytical relations that can be inspected, tested and connected to mechanism.[2] Symbolic regression (SR) is one of the main approaches to this problem: it searches over variables, operators and constants to identify closed-form expressions that explain observed data. This makes SR especially attractive in physics, chemistry, materials science and engineering, where an explicit formula can provide insight beyond numerical prediction.[3–7] However, the same flexibility that makes SR powerful also makes it difficult: the space of possible expressions grows rapidly with the number of variables, operators and allowed expression computation complexity.[8]

Recent large-language-model (LLM) based SR methods have improved equation discovery by injecting scientific priors, semantic information and code-generation ability into the search process.[9–12] Compared with traditional search-based approaches, these approaches provide a hybrid mechanism that couples semantic reasoning with symbolic search.[13–18] A common strategy is agentification: to prompt an LLM with data summaries, variable names, scientific context and previous feedback, and ask it to generate candidate equation skeletons, which are then numerically fitted and evaluated.[19] This paradigm has shown that LLMs can provide useful high-level priors for SR. However, in such skeleton-generation pipelines, the LLM does more than provide priors: it effectively pre-selects the structural search space by deciding which equation forms are passed to numerical fitting. Although numerical optimization can fit constants after a skeleton is proposed, the structural search itself remains limited by the LLM's prompt-level interpretation of the data and context. As a result, direct equation generation can be expensive, because many candidate formulas require repeated LLM calls, and fragile, because generated skeletons may reflect memorized priors, semantic associations or hallucinated relationships rather than fine-grained numerical structure.[20–22]

We argue that LLMs should guide where SR searches, not decide what equation science has discovered. LLMs are valuable because they provide broad scientific priors, integrate heterogeneous information, reason over variable semantics and propose useful analogies. However, their weakness is that they are not inherently reliable numerical search engines: they can be insensitive to small residual patterns, over-rely on familiar scientific forms and produce plausible but poorly grounded structural suggestions. We therefore hypothesize that separating the semantic and numerical roles is essential for reliable equation discovery. In this proposed division of labour, the LLM is restricted to controlling the search space—selecting variables, operators, engineered features, transformations and expression-complexity schedules, whereas the SR engine performs the actual combinatorial search, constant optimization. Ultimately, semantic priors shape hypothesis exploration, but strict numerical evidence retains authority over which equations survive.

To systematically test this hypothesis, we evaluated five distinct role specifications for LLM-assisted equation discovery: equation author, candidate decider, search controller (instantiated here as LLM-PySR), full-stack LLM search, and no-LLM numerical search. By allocating equivalent computational budgets across physics-derived benchmarks and complex symbolic-regression tasks, we found that the controller configuration provides the most favourable joint trade-off between predictive accuracy, expression complexity, and stability. Crucially, it fundamentally reduces the computational cost of AI-driven discovery: on the AI-Feynman benchmark, the search controller achieves state-of-the-art predictive accuracy while consuming merely 4.1% and 3.2% of the LLM tokens required by leading full-stack LLM methods (LLM-SR and DrSR, respectively).[9,10,23] Beyond synthetic benchmarks, we demonstrated the controller's scientific utility on an independent lithium-ion battery dataset.[24] The system autonomously extracted a compact, piecewise-linear empirical law that links early voltage-curve displacement to cycle life—a finding consistent with established electrochemical degradation signatures. Together, these results validate that a numerically grounded division of labour provides a highly efficient, interpretability-first blueprint for the cognitive architecture of scientific AI agents.

## Results

To systematically delineate the boundaries of artificial agency, we conceptualize scientific equation discovery as a closed-loop system comprising three core functional blocks: search-space planning, expression enumeration, and candidate acceptance. By shifting the computational authority over these functions between LLMs and deterministic numerical procedures, we formalize a spectrum of five distinct cognitive role specifications (Fig. 1).

In our proposed search controller paradigm instantiated here as the LLM-PySR framework: the LLM is strictly restricted to high-level search-space planning (proposing variable subsets, operators, and complexity constraints); the structural enumeration is left entirely to a symbolic regression engine (PySR), and final formula retention is governed strictly by deterministic numerical metric gates.[15] The equation author setting relaxes this by allowing the LLM to directly dictate mathematical skeletons while delegating only baseline constant optimization to a numerical solver. Conversely, the candidate decider configuration permits a numerical engine to enumerate formulas, but grants the LLM the ultimate authority to evaluate and filter which candidates survive. To complete this spectrum, we included two established baseline extremes: the full-stack LLM search, represented by two state-of-the-art generative methods (LLM-SR and DrSR), and the no-LLM numerical search, represented by pure PySR.[9,10] Detailed role specifications and controlled variants are introduced in Supplementary Note 2.

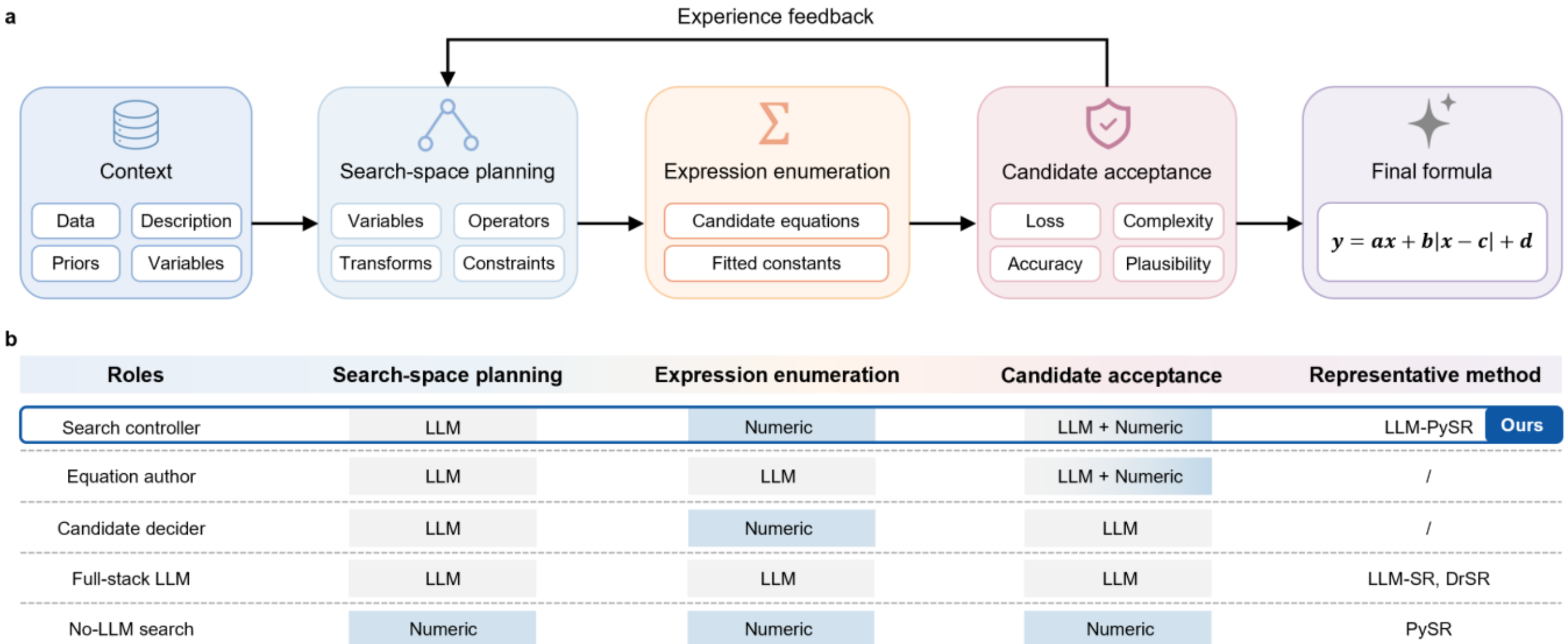

**Fig. 1 | Role specifications for LLM-assisted symbolic regression. a,** Scientific equation discovery is represented as a closed-loop process in which task context informs search-space planning, a symbolic-regression engine enumerates and fits candidate expressions, and candidate acceptance provides feedback for subsequent planning. This abstraction separates three decision points: where to search, which expressions to generate and which candidates to retain. **b,** Five role specifications are obtained by assigning authority over these decision points to language models, numerical procedures or both. In the search-controller setting, implemented here as LLM-PySR, the language model proposes search-space constraints, whereas PySR performs expression enumeration and deterministic numerical criteria govern candidate retention. The remaining settings test alternative allocations in which language models author equations, decide candidate acceptance, conduct full-stack search or are removed entirely.

We evaluated these five role specifications across 74 AI-Feynman equations and seven complex symbolic-regression cases spanning physical, biological, and engineering relations.[23] On the large-scale AI-Feynman benchmark, both the search controller and candidate decider demonstrated a profound enhancement in predictive accuracy ($\mathrm{ACC}_{0.1}$) and normalized mean squared error (NMSE) compared to the remaining three configurations (Fig. 2a). This systemic divergence underscores that leveraging LLMs to guide numerical search spaces is fundamentally more effective than relying entirely on generative language models or unguided numerical exploration. Crucially, the search controller marginally outperformed the candidate decider, a result we attribute to the rigorous numerical stability of deterministic metric gates over LLM-based judgment. Language models frequently suffer from hallucinations, stochastic volatility, and numerical insensitivity when evaluating complex expressions; this mechanistic vulnerability is empirically confirmed by the significantly lower inter-run variance exhibited by the search controller relative to the candidate decider. Statistics for AI-Feynman benchmark results can be found in Extended Data Table 1. Detailed results including recovered formulas can be found in Supplementary Note 4.

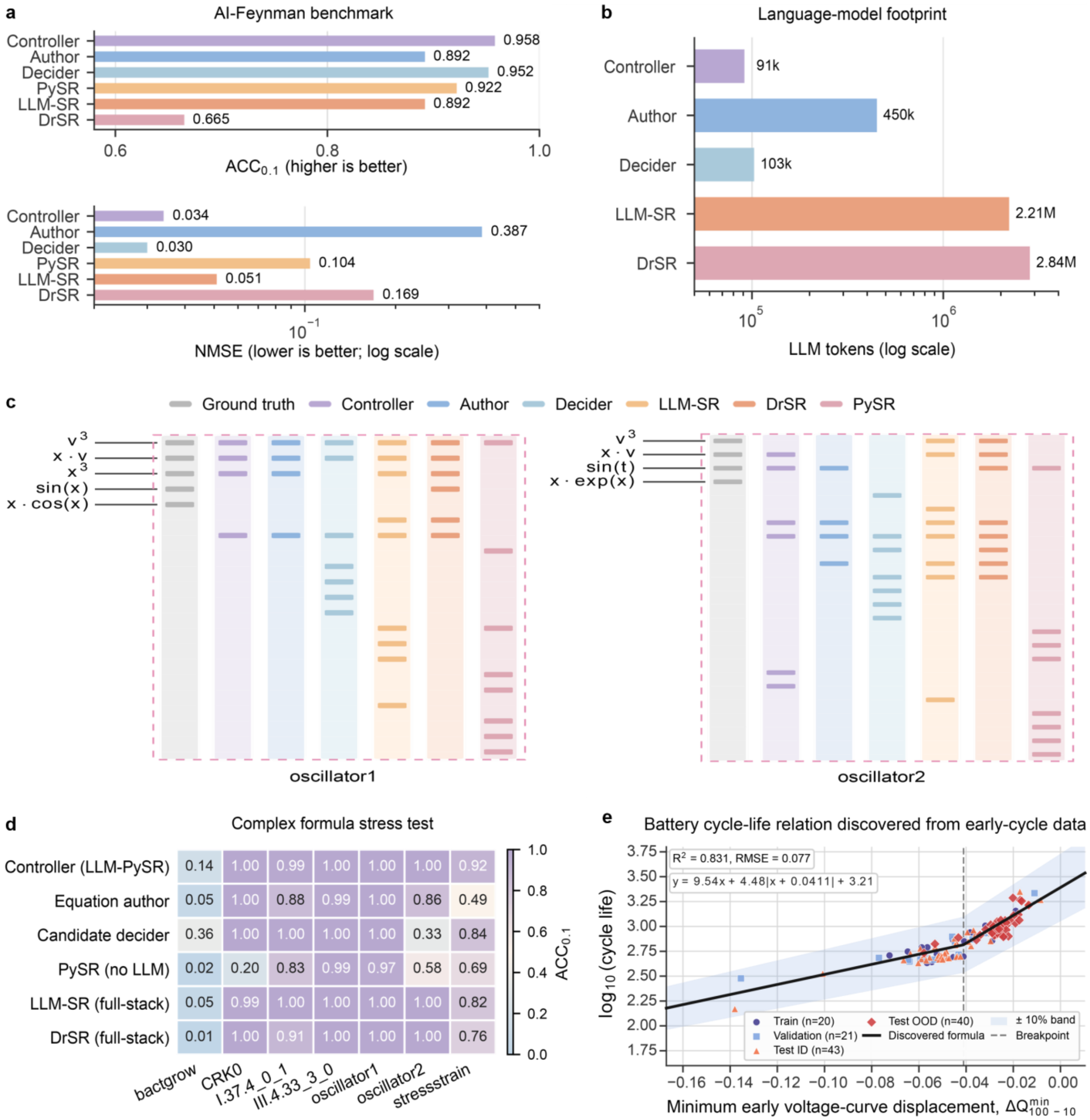

**Fig. 2 | Role-dependent performance in symbolic equation discovery. a,** Predictive performance of the role specifications on 74 AI-Feynman equations, measured by $ACC_{0.1}$ and normalized mean squared error (NMSE). $ACC_{0.1}$ denotes the fraction of test predictions within 10% relative error, with higher values indicating better bounded-error prediction; lower NMSE indicates better overall fit. **b,** Total LLM token consumption on the same benchmark, summed across all LLM calls and shown on a logarithmic scale. The search-controller role combines high predictive accuracy with substantially lower LLM use than full-stack LLM search. **c,** Term-level structural recovery for the oscillator1 and oscillator2 stress-test equations. Target terms are listed on the left, and coloured marks indicate terms recovered by each role specification, showing both recovery of ground-truth components and generation of spurious terms. **d,** Formula-level $ACC_{0.1}$ heat map across seven complex symbolic-regression stress tests spanning biological, chemical, physical and engineering relations. **e,** Application of the search-controller framework to lithium-ion battery cycle-life prediction. The discovered hinge-like expression relates early voltage-curve displacement, $\Delta Q_{100-10,\min}$ , to $\log_{10}$ cycle life. Points denote training, validation, in-distribution test and out-of-distribution test cells; the black curve shows the discovered expression, the vertical dashed line marks the fitted breakpoint, and the shaded region denotes the ±10% relative-error band.

Remarkably, this architectural superiority does not incur a higher computational premium; rather, the search controller and candidate decider fundamentally compressed the LLM footprint. The search controller proved to be the most token-efficient configuration, consuming only 4.1% and 3.2% of the tokens required by full-stack LLM methods, while the equation author configuration demanded orders of magnitude more resource overhead. This massive disparity highlights a straightforward operational reality: token consumption is driven primarily by the verbosity of the delegated task. The computational bottleneck in generative approaches resides in the equation formulation phase, where LLMs must continuously synthesize expansive code structures or Python-like symbolic functions. By restricting the LLM to compact, high-leverage boundary formulation and delegating the verbose structural assembly to numerical procedures, the search controller achieves an optimal alignment of efficiency and precision.

The seven highly complex, domain-specific equations further exposed the distinct vulnerabilities of granting LLMs direct generative or evaluative authority (Fig. 2d). For the generative roles (equation author and full-stack LLM), performance was strictly bounded by familiarity from the base LLM. They achieved near-perfect accuracy on standard scientific structures (such as CRK0 or oscillator1) by retrieving memorized priors, but suffered catastrophic collapse on atypical, highly nonlinear formulations like bactgrow and stressstrain. This failure mode indicates that unconstrained generative LLMs prematurely narrow the exploration space, forcing data to fit familiar but mathematically incorrect templates.

Conversely, the candidate decider configuration exposed a different vulnerability: evaluative instability. Although a numerical engine generated the candidates, empowering the LLM to make the final selection led to surprisingly poor performance on standard physical relations like oscillator2. This anomaly highlights that LLMs lack the fine-grained numerical sensitivity required to reliably distinguish between structurally similar candidates with minute residual differences.

In contrast, the search controller is the sole architecture that isolated the benefits of semantic priors without inheriting either generative or evaluative fragility. By strictly delegating both the combinatorial assembly and the final candidate retention to deterministic numerical procedures, the search controller reliably navigated both familiar and atypical mathematical topologies with robust confidence intervals. This fine-grained performance gap validates a critical boundary in artificial scientific agency: while language models are exceptional at mapping the topology of a hypothesis space, numerical evidence must provide the ultimate selection principle rather than semantic plausibility to ensure robustness across diverse scientific domains.

Having established the robust performance of the search controller on synthetic benchmarks, we sought to determine whether this role separation yields physically meaningful relations in real-world scenarios lacking a known ground-truth equation. To this end, we deployed our instantiated framework, LLM-PySR, to the early prediction of lithium-ion battery lifetime—a critical challenge where degradation labels are intrinsically delayed and obscured by high-dimensional measurement noise.[24] Given only raw discharge-voltage measurements and semantic context, LLM-PySR autonomously pinpointed the minimum change in the discharge curve between cycles 10 and 100 as the governing coordinate, returning a compact analytical relationship:

$$log_{10}(N_{life}) = 9.535x + 4.475|x + 0.0411| + 3.208, \mathrm{x} = \Delta\mathrm{Q}_{100-10,\mathrm{min}} \quad (1)$$

where x is the minimum value of $\Delta\mathrm{Q}_{100-10,\mathrm{min}} = Q_{100}(V) - Q_{10}(V)$ , which is typically negative. On a 40-cell independent test set, this relation achieved an $RMSE$ of 0.0772 and $R^2$ = 0.8314. To render the physical meaning explicit, we define the displacement magnitude $d = -\Delta Q_{100-10,\min}$, representing the magnitude of the most negative early cycle voltage-curve shift. The equation can be rewritten as:

$$log_{10}(N_{life}) = 3.208 - 9.535d + 4.475|0.0411 - d|. \quad (2)$$

The equation reveals a piecewise-linear decay, monotonically predicting shorter lifetimes for larger early voltage-curve shifts. This autonomous mathematical formulation aligns elegantly with established electrochemical domain knowledge: while early capacity fade is often negligible and visually imperceptible, underlying degradation modes such as the loss of active material in the negative electrode may silently shift the voltage curves long before overt capacity loss appears.[25,26] Consequently, we interpret the discovered threshold at $d \approx 0.0411\ Ah$ as a data-driven empirical breakpoint marking the transition between mild and severe early degradation regimes. This result demonstrates that LLM-PySR, by strictly guiding numerical search without forcing structural generation, can distil highly predictive and mechanistically interpretable laws directly from complex experimental observations.

Together, the experiments identify role specification as a central design choice in hybrid scientific systems. The result is not that language models should be excluded from hypothesis formation. Rather, their broad but fallible priors are most useful when expressed as revisable constraints on a transparent numerical search. This separation also makes failure easier to diagnose: poor performance can be attributed to an unsuitable search specification, insufficient numerical exploration or an overly permissive acceptance rule, rather than to an opaque sequence of generated formulas. The present evidence suggests a testable principle for their design: learned models may guide where hypotheses are sought, while explicit numerical procedures determine which hypotheses survive.

## Discussion

These results identify role specification as a central design choice in language-model-assisted equation discovery. Rather than asking only whether language models should be added to symbolic regression, our comparisons ask which decisions should be assigned to them. The answer that emerges from the benchmarks is clear: language models are useful when their semantic priors are expressed as constraints on the search space, but less reliable when they are asked to author equations directly or decide which candidates should be retained. In LLM-PySR, the language model proposes variables, transformations, operators and complexity schedules, whereas PySR performs expression enumeration and deterministic numerical criteria govern candidate retention. Under the tested configurations, this controller role provided the strongest overall balance among predictive accuracy, expression complexity, stability and language-model cost, while using less than one-twentieth of the tokens required by full-stack generative baselines. The battery case suggests that this division of labour can also be useful beyond synthetic formula recovery. The controller selected early voltage-curve displacement and returned a compact hinge-like relation with cycle life. Its value is that the recovered relation is inspectable, quantitatively testable and linked to a physically meaningful coordinate. More generally, these results support a design principle for scientific AI: learned models may guide where hypotheses are sought, but explicit numerical procedures should determine which hypotheses survive.

## Methods

### Role specifications and common pipeline

To systematically evaluate the cognitive boundaries of large language models in equation discovery, we establish the search controller configuration (the LLM-PySR framework) as our foundational pipeline. This framework conceptualizes the SR process through five primary components:

(1) Search-space planning and sanitization: An LLM agent interprets the scientific context and historical memory to propose structured search boundaries (e.g., selected variables, operators, and complexity constraints), which are subsequently sanitized by deterministic validity checks.
(2) Numerical search and unified evaluation: A deterministic SR engine (PySR) executes low-level symbolic enumeration and constant optimization within the constrained search space, returning a pool of candidate expressions that are re-evaluated uniformly.
(3) Candidate review and metric-based gating: An LLM reviewer provides auxiliary semantic and interpretability scores for the candidates, but the final positive/negative feedback labels are strictly assigned by deterministic numerical metric gates.
(4) Experience memory: An LLM agent synthesizes the deterministic feedback and candidate evaluations into a structured experience memory to inform the next planning iteration.
(5) Formula simplification and final model selection: A deterministic post-hoc ablation and pruning step greedily simplifies subexpressions to ensure the final mathematical formula is maximally compact without degrading performance.

Building upon this modular framework, we define two distinct generative role specifications by shifting the computational authority of specific deterministic steps to the LLM. Equation author alters the second step: instead of delegating structural enumeration to the numerical engine, the LLM directly generates and proposes mathematical equation skeletons based on the context and memory. The numerical solver is restricted to baseline constant optimization. Candidate decider alters the third step: we remove the deterministic metric-based gating, granting the LLM reviewer the ultimate authority to filter, evaluate, and dictate which candidate equations survive based on its holistic assessment of numerical metrics and semantic plausibility.

To contextualize these architectural variations, we included two baseline extremes utilizing complete, external frameworks. No-LLM numerical search: We employed pure PySR to represent an unguided, purely deterministic combinatorial search devoid of semantic priors or LLM intervention. Full-stack LLM search: We selected two state-of-the-art LLM-based SR methods, LLM-SR and DrSR, to represent end-to-end generative paradigms. LLM-SR utilizes an evolutionary framework where an LLM iteratively proposes, evaluates, and mutates candidate equations. Conversely, DrSR employs advanced search-based prompting strategies (such as tree-search reasoning) to guide the LLM through step-by-step hypothesis generation and refinement.

Full details of the LLM-PySR framework are introduced in Extended Data Fig. 1, Supplementary Algorithm 1 and Supplementary Note 1. Hyperparameter configurations of LLM-PySR are introduced in Supplementary Note 3. Detailed prompt templates used in LLM-PySR are introduced in Supplementary Note 5.

**Benchmarks, controls and data splits**

We evaluated the five roles using two synthetic datasets to benchmark the overall performance and efficiency:

(1) AI-Feynman dataset.[23] The AI-Feynman dataset comprises 100 physics equations derived from the Feynman Lectures on Physics. These equations represent compact analytical relationships commonly encountered in physical sciences. To ensure sufficient training data for symbolic regression, equations with fewer than 500 samples were removed, leaving 74 equations for evaluating all compared methods.
(2) DrSR-aligned complex formula suite.[10] To facilitate direct comparison with full-stack LLM search methods, we further evaluated the five roles on the representative formula-recovery tasks adopted in the DrSR study. These tasks cover 7 nontrivial symbolic expressions from physics, biology, chemistry, and materials science, including nonlinear oscillator dynamics, bacterial growth, stress--strain behaviour, and representative LSR-Transform and LSR-Synth cases from LLMSR-bench.[27] Because this suite contains only a small number of representative cases, we use it as a formula-level stress test for complex symbolic recovery rather than as a large-scale statistical benchmark.

For each benchmark equation, at most 500 samples were randomly selected for training, and all remaining samples were used for testing. In addition, for search controller, equation author and candidate decider that follows our controlled pipeline， the 500 training samples were further split into 250 training samples and 250 validation samples. The validation set was used only for search feedback, model selection and experience accumulation, and the final reported metrics were computed on the held-out test set.

To ensure a fair comparison for full-stack LLM search, we adopted the default hyperparameter settings recommended in the original papers whenever possible, rather than enforcing an identical number of outer-loop iterations across methods: LLM-SR used a maximum of 2500 generated samples, while DrSR was configured with 10 iterations and 4 samples per iteration. For search controller, LLM-PySR was run for at most 10 iterations with a PySR population size of 128. The initial PySR search budget was set to 200 iterations and adaptively adjusted within the range of 100--500 iterations during the search process. Equation author, candidate decider and no-LLM PySR follows the same settings. All LLM-based methods used GPT-5.1 as the underlying language model.[28]

To further showcase the scientific discovery ability of LLM-PySR, we applied it to a real-world problem with limited data: Battery life prediction. Severson et al. generated a dataset containing 124 commercial lithium iron phosphate/graphite cells cycled under fast-charging conditions, with widely varying cycle lives ranging from 150 to 2,300 cycles.[24] For feature engineering, 11 key statistical indicators (such as variance, minimum, and mean) were extracted from the evolution of discharge voltage curves $Q(V)$ over the first 100 cycles (e.g.,$\Delta Q_{100-10}(V)$), alongside physical measurements including capacity degradation, internal resistance, and temperature. The target output feature was the cycle life, defined as the number of cycles until the cell reached 80% of its nominal capacity. The original study rigorously partitioned the 124 cells into a training set of 41 cells, a primary test set of 43 cells (which we repurposed as our validation set), and a secondary test set of 40 cells used for independent evaluation. Consequently, our experiments strictly adhere to this data splitting protocol. In the LLM-PySR framework, the 41-sample training set was utilized for PySR's low-level symbolic space exploration and constant optimization. The 43-sample validation set guided the LLM agents' experience feedback and deterministic gating selection. Finally, the 40-sample independent test set was employed to evaluate the generalization performance of the discovered equations.

**Evaluation metrics**

Following DrSR, we evaluated symbolic regression performance using Accuracy under Error Tolerance ($ACC_\tau$) and Normalized Mean Squared Error ($NMSE$). In all experiments, we set $\tau = 0.1$. Given test predictions $y_{\text{pred}}$ and ground-truth values $y_{\text{true}}$, $ACC_\tau$ is defined as:

$$ACC_\tau = \frac{1}{n}\sum_{i=1}^{n} \mathrm{I}\left(\left|y_{\text{pred},i} - y_{\text{true},i}\right| \le \tau\left|y_{\text{true},i}\right|\right) \tag{3}$$

where $\mathrm{I}(\cdot)$ is the indicator function. $ACC_\tau$ measures the fraction of predictions whose relative error is bounded by the prescribed tolerance and therefore reflects the ability of a discovered equation to provide bounded-error predictions.

$NMSE$ measures the overall prediction error normalized by the variance of the target variable:

$$NMSE = \frac{\frac{1}{n}\sum_{i=1}^{n}(y_{\text{pred},i} - y_{\text{true},i})^2}{\frac{1}{n}\sum_{i=1}^{n}(y_{\text{true},i} - \overline{y_{\text{true}}})^2}. \quad (4)$$

Lower $NMSE$ indicates better predictive accuracy, whereas higher $ACC_\tau$ indicates a larger fraction of predictions within the tolerance interval. To evaluate computational efficiency and keep a fair comparison between different algorithm configurations, we additionally record the overall LLM token consumption, which is computed as the sum of all input and output tokens across all LLM calls made during the search.

While exact equation recovery is a standard metric for synthetic benchmarks, we intentionally prioritize prediction-based metrics. Real-world scientific discovery rarely provides predefined analytical expressions, and multiple mathematically distinct expressions can often produce nearly identical predictions under finite, noisy observations.[29] Therefore, we rely on $ACC_\tau$ and $NMSE$ for quantitative evaluation, while assessing the scientific validity of discovered formulas based on compactness and consistency with domain knowledge.

**Theoretical analysis**

We provide a statistical interpretation of the role separation used in this research to clarify the potential trade-offs from different LLM roles and actions.

Let $\mathcal{D} = \{(x_i, y_i)\}_{i=1}^{n}$ be sampled from an unknown distribution $P$, and let $R(f) = E_{(x,y)\sim P}[L(f(x), y)]$ denote the population risk under a bounded loss $L$. Let $\hat{R}_n(f)$ be the empirical risk. In a standard PAC-style uniform convergence bound, for a fixed symbolic hypothesis class $\mathcal{F}$, with probability at least $1 - \delta$,

$$\sup_{f\in\mathcal{F}}\left|R(f) - \hat{R}_n(f)\right| \lesssim \sqrt{\frac{C(\mathcal{F}) + \log(1/\delta)}{n}} \quad (5)$$

where $C(\mathcal{F})$ denotes an appropriate statistical complexity measure of the class, such as pseudo-dimension, covering complexity or a Rademacher-complexity-based term. If an algorithm returns an approximate empirical-risk minimizer $\hat{f} \in \mathcal{F}$ satisfying

$$\hat{R}_n(\hat{f}) \leq \inf_{f\in\mathcal{F}} \hat{R}_n(f) + \eta, \quad (6)$$

then the corresponding excess risk over the best function in $\mathcal{F}$ obeys

$$R(\hat{f}) - \inf_{f\in\mathcal{F}} R(f) \lesssim \sqrt{\frac{C(\mathcal{F}) + \log(1/\delta)}{n}} + \eta. \quad (7)$$

In symbolic regression, the additional term $\eta$ is important. Unlike convex empirical-risk minimization, symbolic regression relies on finite-budget combinatorial search and constant optimization; hence the returned expression need not be the empirical optimum of the specified symbolic class. We therefore interpret $\eta$ as a finite-budget search error.

Let $\mathcal{F}_0$ denote the full symbolic class accessible to an unconstrained PySR run under the global variable set, operator set and maximal expression complexity. At an LLM-PySR iteration, the LLM planner returns a search specification $\phi$, including selected variables, allowed operators, engineered transformations and complexity constraints. This induces a restricted symbolic class $\mathcal{F}_{\phi} \subseteq \mathcal{F}_0$. The full-class oracle is $f_0^{\star} \in \arg\min_{f\in\mathcal{F}_0} R(f)$. Because LLM-PySR searches only inside $\mathcal{F}_{\phi}$, it may exclude the full-class optimum. We define the pruning bias as $\mathcal{B}_{\phi} = \inf_{f\in\mathcal{F}_{\phi}} R(f) - \inf_{f\in\mathcal{F}_0} R(f)$, this term is zero only when the LLM-induced restricted class retains a population-risk minimizer, or at least a near-optimal expression, from the full symbolic class.

If $\phi$ is selected from data, an additional model-selection cost must also be paid. We denote this cost by $\Lambda_{\phi}$. For a finite set of possible search specifications, $\Lambda_{\phi}$ can be taken as $\log(1/\pi(\phi))$ under a prior over specifications $\pi(\phi) > 0, \sum_{\phi\in\Phi} \pi(\phi) \leq 1$. This term accounts for the fact that a data-dependent search space is not a free prior.

Combining these terms, LLM-PySR admits the following decomposition:

$$R(\hat{f}_{\phi}) - R(f_0^{\star}) \lesssim \underbrace{\mathcal{B}_{\phi}}_{\text{pruning bias}} + \underbrace{\sqrt{\frac{C(\mathcal{F}_{\phi})+\Lambda_{\phi}+\log(1/\delta)}{n}}}_{\text{statistical estimation error}} + \underbrace{\eta_{\phi}(N_{\text{eval}})}_{\text{finite-budget search error}} \quad (8)$$

where $N_{\text{eval}}$ denotes the expression-evaluation budget used by the symbolic search engine.

This decomposition clarifies why merely reducing the search space is not sufficient. Search-space restriction decreases the statistical complexity term only if

$$C(\mathcal{F}_{\phi}) < C(\mathcal{F}_0). \quad (9)$$

For monotone complexity measures, this follows from $\mathcal{F}_{\phi} \subseteq \mathcal{F}_0$. Another intuitive indicator expression skeleton count number $|S_K| \approx O(K^L)$ shows the number of admissible expression structures grows with the number of possible operators:

$$\left|S_{K_{\phi}}\right| \sim O\left({K_{\phi}}^{L}\right), \left|S_{K_0}\right| \sim O\left({K_0}^{L}\right), K_{\phi} < K_0 \quad (10)$$

where $K_{\phi}$ and $K_0$ represent the number of possible operators of LLM-PySR and PySR, and $L$ represents the skeleton length. However, it may simultaneously increase the pruning bias $\mathcal{B}_{\phi}$. Therefore, LLM-PySR is advantageous only when the reduction in statistical and search complexity outweighs the approximation loss caused by excluding parts of the full symbolic class.

For an unconstrained PySR baseline, the corresponding bound is:

$$R(\hat{f}_0) - R(f_0^{\star}) \lesssim \sqrt{\frac{C(\mathcal{F}_0)+\log(1/\delta)}{n}} + \eta_0(N_{\text{eval}}). \quad (11)$$

Thus, LLM-PySR is theoretically favoured over unconstrained PySR under the sufficient condition:

$$\mathcal{B}_{\phi} + \sqrt{\frac{C(\mathcal{F}_{\phi})+\Lambda_{\phi}+\log(1/\delta)}{n}} + \eta_{\phi}(N_{\text{eval}}) < \sqrt{\frac{C(\mathcal{F}_0)+\log(1/\delta)}{n}} + \eta_0(N_{\text{eval}}). \quad (12)$$

This condition exposes the central bias--complexity--search trade-off. Plain PySR has no pruning bias, because it searches the full symbolic class, but it generally pays a larger statistical complexity penalty and may suffer a larger finite-budget search error. LLM-PySR accepts a nonzero pruning bias in exchange for a lower-complexity symbolic class and a potentially easier search problem. This advantage is most pronounced when semantic priors remove many irrelevant expressions while preserving the region containing a near-optimal formula.

We can view the risk bound of equation author with the same inequality:

$$R(\hat{f}_{author}) - R(f_0^\star) \lesssim \mathcal{B}_{author} + \sqrt{\frac{C(\mathcal{F}_{author}) + \Lambda_{author} + \log(1/\delta)}{n}} + \eta_{author}(N_{eval}) \quad (13)$$

In this role, the LLM directly proposes a finite set of equation skeletons. If $M_{author}$ skeletons are generated, the structural selection component can be much smaller than that of a broad symbolic search space, and $C(\mathcal{F}_{author})$ can be substantially smaller than $C(\mathcal{F}_\phi)$ and $C(\mathcal{F}_0)$, up to the remaining complexity of fitting constants within each skeleton. However, this aggressive compression can substantially increase the pruning bias $\mathcal{B}_{author}$, because the generated skeletons are discrete samples from the LLM prior and may fail to cover the region containing the true or near-optimal formula. Our empirical results are consistent with this interpretation: direct equation authorship often produces formulas close to familiar LLM priors, which reduces structural complexity but can incur large pruning bias when the target relation is atypical. Search control, by contrast, uses the LLM prior only to delimit the operator space while leaving structural enumeration and constant optimization to PySR. This allows the system to retain broader coverage within a restricted subspace, leading to a more favourable balance between pruning bias, statistical complexity and finite-budget search error.

For the candidate decider role, the symbolic search space may remain the same as search controller because PySR still generates candidate expressions. However, the final selection is delegated to the LLM. This introduces an additional selection-noise term:

$$R(\hat{f}_{decider}) - R(f_0^\star) \lesssim \mathcal{B}_\phi + \sqrt{\frac{C(\mathcal{F}_\phi) + \Lambda_\phi + \log(1/\delta)}{n}} + \eta_\phi(N_{eval}) + \zeta_{\mathrm{LLM}}, \quad (14)$$

considering *candidate decider* inherits all other parts from *search controller* other than the term $\zeta_{\mathrm{LLM}}$. This term introduces additional selection noise, which captures the risk gap between the LLM-selected candidate and the numerically best candidate in the pool. This decomposition also explains why the *candidate decider* can exhibit average performance close to that of the *search controller*: in the matched setting, most error terms are shared, and the performance gap is mainly governed by $\zeta_{\mathrm{LLM}}$. When LLM judgment agrees with numerical selection, this gap is small; when the LLM is insensitive to fine residual differences or overweights semantic plausibility, $\zeta_{\mathrm{LLM}}$ can increase and lead to unstable or suboptimal selection.

Thus, the advantage of LLM-PySR is not simply that it reduces the search space. Rather, it uses semantic priors to reduce operator-induced symbolic complexity while preserving numerical enumeration within the restricted class and avoiding LLM-based final selection. This role separation seeks a favourable balance among pruning bias, statistical complexity and finite-budget search error, while preventing semantic plausibility from overriding numerical evidence.

## Data availability

All dataset to conduct the experiments, as well as the experiment result data are available at https://github.com/XieZikai/LLM-PySR.

## Code availability

All Python scripts for LLM-PySR, as well as the equation author variant and candidate decider variant are available at https://github.com/XieZikai/LLM-PySR.

## Acknowledgements

The AI-driven experiments, simulations and model training were performed on the robotic AI-Scientist platform of Chinese Academy of Sciences (CAS). L.C. acknowledges the National Natural Science Foundation of China (Grant 22573101) and the University of Science and Technology of China (USTC) Startup Program for funding. J.J. acknowledges the CAS Strategic Priority Research Program (Grant XDB0450302), the National Natural Science Foundation of China (Grants 22025304, 22033007), and the CAS Project for Young Scientists in Basic Research (Grant YSBR-005) for funding.

## Author contributions

L.C., J.J. and Z.X. conceived the idea. Z.X. and W.L. designed the algorithm workflow, implemented the algorithm and conducted the experiments. M.L. designed and prepared all figures. Z.X. and L.C. led the writing, with contributions from all authors.

## Competing interests

The authors declare no competing interests.

## Additional information

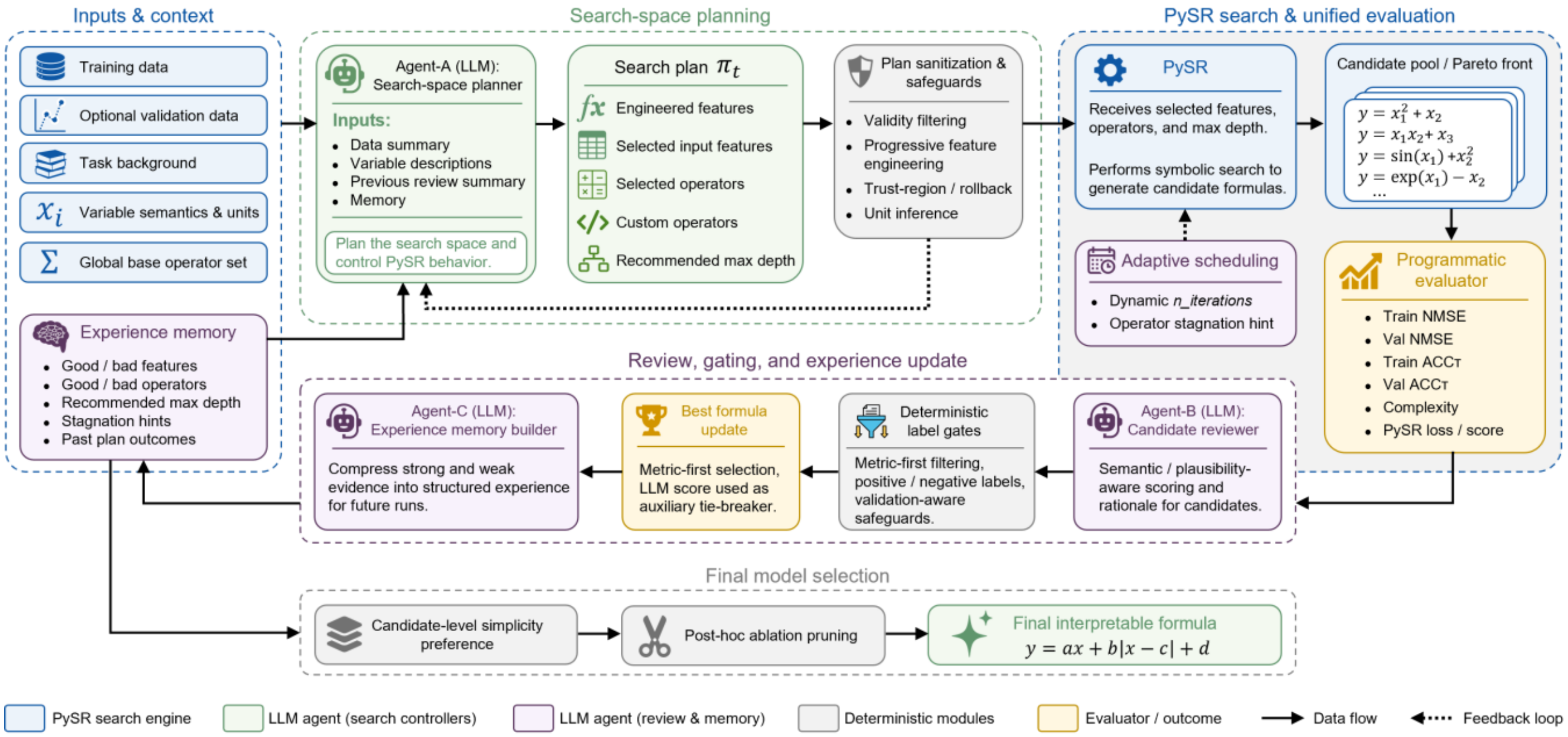

**Extended Data Fig 1. | Detailed workflow of LLM-PySR algorithm.** LLM-PySR operates as an iterative search-controller framework for symbolic regression. Given data, task context, variable semantics, an admissible operator set and accumulated experience memory, an LLM planner proposes a bounded search plan specifying candidate variables, engineered features, operators, transformations and expression depth. This plan is sanitized by deterministic safeguards and executed by PySR, which enumerates symbolic expressions and fits constants. Candidate formulas are assessed by a unified programmatic evaluator, while deterministic metric gates assign retention labels based on predictive error, tolerance accuracy, complexity and stability. LLM-based review is used only as auxiliary semantic feedback, and the accepted numerical outcomes are summarized into experience memory for the next iteration. After the iterative search, simplicity-guided selection and post-hoc ablation pruning yield the final compact interpretable formula.

**Extended Data Table 1 | Summary performance and LLM footprint on the AI-Feynman benchmark.**

| Role | $ACC_{0.1}$ ↑ | NMSE ↓ | Token ↓ |
|---|---|---|---|
| Search controller (LLM-PySR) | 0.958 ± 0.165 | 0.034 ± 0.023 | 91,273 ± 74,873 |
| Equation author | 0.892 ± 0.259 | 0.387 ± 2.050 | 449,627 ± 315,861 |
| Candidate decider | 0.952 ± 0.183 | 0.030 ± 0.135 | 102,527 ± 89,787 |
| No-LLM (PySR) | 0.922 ± 0.237 | 0.104 ± 0.170 | — |
| Full-LLM (LLM-SR) | 0.892 ± 0.270 | 0.051 ± 0.028 | 2,208,135 ± 464,128 |
| Full-LLM (DrSR) | 0.665 ± 0.385 | 0.169 ± 0.068 | 2,835,147 ± 893,329 |

Values are reported as mean ± s.d. across the 74 AI-Feynman equations. ACC0.1 denotes the fraction of test predictions within a relative-error tolerance of 0.1, with higher values indicating better predictive accuracy. NMSE denotes normalized mean squared error, with lower values indicating better predictive accuracy. Total LLM tokens are computed as the sum of input and output tokens across all LLM calls made during the search. The no-LLM PySR baseline makes no LLM calls; therefore, LLM-token usage is not applicable. The best performances are shown in bold.

## Supplementary Information

# Language models guide symbolic equation discovery by controlling search

Zikai Xie[1,2,†,*], Wenmei Li[1,†], Man Luo[1], Jun Jiang[1,2,*] and Linjiang Chen[1,2,3,*]

[1] State Key Laboratory of Precision and Intelligent Chemistry, Hefei National Research Center for Physical Sciences at the Microscale, School of Chemistry and Materials Science, University of Science and Technology of China, Hefei, China.
[2] Center for Scientific Intelligence Innovation, Hefei, China
[3] School of Chemistry, School of Computer Science, University of Birmingham, Birmingham, UK
[†] These authors contributed equally: Zikai Xie, Wenmei Li
* Corresponding authors: linjiangchen@ustc.edu.cn, jiangj1@ustc.edu.cn, zikaix@ustc.edu.cn

## Table of Contents

**Supplementary Note 1.** LLM-PySR implementation details

LLM-PySR is a closed-loop symbolic regression framework that uses LLM agents to control the search behavior of PySR rather than directly generate final equations. This design is related to recent agentic LLM frameworks that use reasoning, feedback, and memory to improve iterative decision making. Given a training set $\mathcal{D}_{\mathrm{tr}}$, an optional validation set $\mathcal{D}_{\mathrm{val}}$, and task context $c$ including variable semantics and units, LLM-PySR maintains an experience memory $\mathcal{M}_t$ over search iterations.

At iteration $t$, a search-space planner proposes a search plan

$$\pi_t = (\mathcal{E}_t, \mathcal{X}_t, \mathcal{O}_t, \mathcal{U}_t, d_t),$$

where $\mathcal{E}_t$ denotes engineered features, $\mathcal{X}_t$ denotes selected input variables, $\mathcal{O}_t$ denotes selected base operators, $\mathcal{U}_t$ denotes custom operators, and $d_t$ denotes the maximum search depth. The raw plan is sanitized by deterministic validity checks and trust-region constraints before being passed to PySR. PySR then searches for candidate expressions under the resulting search configuration and returns a candidate pool $\mathcal{P}_t$.

Each candidate is evaluated by a unified evaluator on both training and validation data. A reviewer agent provides auxiliary semantic scores $r_t$, but final positive and negative feedback labels $z_t$ are assigned by deterministic metric-based gates. These labels are then compressed by a memory-building agent into structured evidence memory $\mathcal{M}_t$ about useful features, operators, and search-depth choices. The updated memory is used to guide the next planning step. This design separates high-level semantic reasoning from low-level symbolic enumeration and numerical optimization, while reducing the risk that LLM-generated feedback overrides numerical evidence. The psuedo-code of the whole algorithm is illustrated in **Supplementary Algorithm 1**.

The five core components of LLM-PySR are introduced below:

**Agent-A: Search space planner agent.** At iteration $t$, Agent A proposes the search configuration for the next PySR run. It receives the task context, variable descriptions, a compact data snapshot, column-level statistics, the current experience memory, the previous candidate-review summary, the maximum-depth recommendation from the memory builder, and the globally admissible base-operator set. It outputs a single structured plan consisting of engineered features, a selected PySR input-feature subset, a selected subset of base operators, optional custom unary or binary operators, and a recommended maximum search depth. Each engineered feature contains a name, an expression, a rationale, and a dimension-check field. The selected base operators must come from the global operator whitelist, while custom operators are restricted to formal arguments only, e.g., $x$ for unary operators and $(x, y)$ for binary operators. In our implementation, we use at most five selected input features, at most $K_{\mathrm{custom}}$ custom operators, and restrict the recommended maximum depth to [6,40].

The raw planner output is not directly executed. Instead, LLM-PySR applies a deterministic sanitization operator $\Gamma$ before passing the plan to PySR:

$$\widetilde{\pi_t} = \Gamma(\pi_t^{\mathrm{raw}}, \mathcal{M}_{t-1}).$$

This sanitization step removes invalid or redundant custom operators, deduplicates engineered features, rejects unsafe or overly composite feature transformations, and materializes a proposed feature only if it can be safely evaluated on all samples with finite outputs. Although Agent A may provide a dimension-check rationale, dimensional validity is determined by deterministic unit-propagation rules rather than by the LLM itself. To prevent abrupt expansion of the search space, we further impose progressive feature engineering, allowing at most $K_{\mathrm{new}}$ new engineered features per iteration, with $K_{\mathrm{new}} = 3$ by default. In addition, once a planner configuration yields the current best formula, it is stored as an anchor configuration; subsequent plans are constrained to make only limited changes around this anchor. When a historically successful planner configuration is available, $\Gamma$ also applies a trust-region constraint that limits excessive deviation from the best previous plan. If the system receives no positive feedback for several consecutive iterations, the planner configuration is rolled back toward the best historical Agent-A plan. This validation-and-control mechanism stabilizes the search loop by ensuring executable plans, limiting combinatorial growth, and preventing noisy LLM feedback from causing large search-space drift.

**PySR search and deterministic evaluation.** Given the sanitized plan, PySR performs low-level symbolic search over the selected variables, primitive operators, custom operators, and complexity constraints. Engineered features are computed using a safe expression evaluator and are included only if they produce finite values with the correct sample dimension. When unit annotations are available, units are propagated through engineered features by deterministic dimensional-analysis rules and passed to PySR as additional constraints. The PySR search budget is scheduled adaptively across iterations: larger selected feature/operator sets and low recent ACC lead to more PySR iterations, whereas high-ACC rounds reduce the search budget.

PySR returns a pool of candidate expressions $\mathcal{P}_t$ together with its native loss, score, and complexity. We then re-evaluate all candidates using a unified evaluator on both training and validation data, if available, computing normalized mean squared error and tolerance accuracy. Constant formulas and invalid expressions are filtered out before review. When engineered features are used, expressions selected for the incumbent or final report are recursively expanded back to the original variables, so that the reported model is expressed in the original problem variables rather than in temporary engineered-feature names.

**Agent-B: Candidate reviewer agent.** The reviewer agent evaluates the candidate expressions returned by PySR and assigns an auxiliary score $r_t$ to each expression. Its input includes the symbolic expression, PySR-native metrics, and the metrics recomputed by our pipeline, including training ACC, validation ACC, NMSE, validation NMSE, PySR loss, PySR score, and expression complexity. Agent-B is prompted to score each candidate according to numerical fit quality, simplicity, interpretability, and consistency with the semantic descriptions of the variables. In particular, validation ACC is treated as the primary reliability signal, followed by training ACC, NMSE, PySR loss, and PySR score as a secondary tie-breaker. The prompt also imposes conservative score-calibration rules: low-accuracy expressions cannot receive high reviewer scores, and scores above a high threshold are only permitted when both training and validation accuracy are sufficiently strong. Thus, Agent-B provides a semantically informed but calibrated quality estimate rather than an unconstrained subjective preference.

The reviewer score is advisory rather than decisive. Final feedback labels $z_t$ are assigned by deterministic gates outside the LLM. In the bootstrap stage, where no reliable incumbent is available, only a small elite subset of numerically top-ranked candidates can be labeled positive, and they must satisfy minimum accuracy and reviewer-score thresholds. In later iterations, each candidate is compared with the current best expression using training ACC, validation ACC, NMSE, PySR loss, and PySR score. A positive label is assigned only when the candidate belongs to the elite numerical subset and receives sufficient metric-based support relative to the incumbent; otherwise the label is forced to negative, even if the LLM reviewer assigns a high score. This hybrid design preserves the LLM's ability to assess interpretability and semantic plausibility, while preventing optimistic or inconsistent LLM judgments from being directly propagated to the memory module.

**Agent-C: Experiment memory agent.** The memory agent converts the outcome of each PySR run into structured search experience for the next planning step. Its input contains the sanitized Agent-A plan $\widetilde{\pi_t}$, together with the reviewed candidate pool and the deterministic positive/negative labels $(\mathcal{P}_t, r_t, z_t)$ from Agent-B. To reduce noise in the feedback signal, we separate the evidence into two channels. Candidates labeled positive and satisfying minimum training and validation ACC thresholds are treated as strong evidence, while low-accuracy positives and the best metric-ranked negative candidates are retained as a weak exploratory channel. The strong channel is used to identify reliable feature combinations, useful primitive operators, and appropriate search-depth ranges. The weak channel is not allowed to dominate the memory, but it can suggest exploratory directions when the current search space is underperforming. Agent-C is therefore prompted to output concise guidance $\mathcal{M}_t$ rather than free-form reflections: good and bad feature combinations, good and bad primitive operators, and an optional recommended maximum expression depth. In addition, when the selected operator set remains nearly unchanged for multiple rounds and the best metrics show little improvement, the memory supplied to Agent-A is augmented with an exploration hint that recommends trying unused primitive operators not marked as harmful by Agent-C. Then this guidance is passed to Agent-A to santize $\pi_t^{raw}$ into $\pi_t$, forming an iterative self-learning loop.

**Formula simplification and final model selection.** To improve interpretability, LLM-PySR applies two levels of simplification. During each iteration, Agent-B prefers simpler candidates when their training and validation tolerance accuracy remain within a small degradation tolerance of the current best expression. After the search terminates, a post-hoc symbolic ablation pruning step greedily replaces subexpressions with simple constants when doing so reduces complexity without materially degrading training or validation performance.

**Supplementary Algorithm 1.** LLM-PySR search-controller algorithm

---

**Algorithm 1** LLM-PySR

---

**Require:** Training data $\mathcal{D}_{tr}$, validation data $\mathcal{D}_{val}$, context $c$, base operator set $\mathcal{O}_{base}$, number of iterations $T$

1: Initialize memory $\mathcal{M}_0 \leftarrow \emptyset$ and best formula $f^\star \leftarrow \emptyset$

2: **for** $t = 1, \dots, T$ **do**

3: $\pi_t^{raw}\text{Planner}(c, \mathcal{D}_{\text{t}}, \mathcal{M}_{t-1}, \mathcal{O}_{\text{bs}})$

4: $\pi_t \leftarrow \text{Sanitize}(\pi_t^{\text{raw}}, \mathcal{M}_{t-1})$

5: Construct engineered features and transformed datasets using $\pi_t$

6: $\mathcal{P}_t \leftarrow \text{PySR}(\mathcal{D}_{\text{t}}; \pi_t)$

7: Evaluate each $f \in \mathcal{P}_t$ on $\mathcal{D}_{tr}$ and $\mathcal{D}_{val}$

8: $r_t \leftarrow \text{Reviewer}(c, \mathcal{P}_t, f^\star)$

9: $z_t \leftarrow \text{Gate}(\mathcal{P}_t, r_t, f^\star)$

10: $f^\star \leftarrow \text{UpdateBest}(f^\star, \mathcal{P}_t, r_t, z_t)$

11: $\mathcal{M}_t \leftarrow \text{MemoryBuilder}(\mathcal{M}_{t-1}, \pi_t, \mathcal{P}_t, z_t)$

12: **end for**

13: $f^\star \leftarrow \text{AblationPrune}(f^\star, \mathcal{D}_{tr}, \mathcal{D}_{val})$

14: **return** $f^\star$

---

**Supplementary Note 2.** Role specifications and controlled variants

To isolate the effect of assigning different scientific-discovery functions to language models, we define symbolic regression as a three-stage closed-loop process: search-space planning, expression enumeration and candidate acceptance. Search-space planning determines which variables, engineered features, primitive operators, custom transformations and expression-complexity constraints are made available to the search. Expression enumeration determines which mathematical structures are generated and fitted. Candidate acceptance determines which expressions are retained as positive evidence, used to update the search memory, or selected as incumbents. The five role specifications compared in this study differ only in how decision authority over these three stages is allocated between LLM-based agents and deterministic numerical procedures.

**Search controller: LLM-PySR.** The search-controller configuration is the proposed LLM-PySR framework, following the algorithm demonstrated in Supplementary Note 1 and Supplementary Algorithm 1. In this role, the LLM has authority only over high-level search-space planning. At each iteration, the LLM planner proposes a bounded search plan containing selected variables, engineered features, allowed operators, optional custom transformations and a recommended maximum expression depth. This plan is sanitized by deterministic validity, safety and trust-region checks before execution. The LLM is not allowed to directly generate candidate equations.

Expression enumeration is performed entirely by PySR. Given the sanitized search configuration, PySR carries out symbolic enumeration and constant optimization, returning a candidate pool with native loss, score and complexity values. Candidate expressions are then re-evaluated by the unified evaluator using the same training and validation metrics across all controlled variants.

Candidate acceptance in LLM-PySR is LLM-assisted but numerically gated. The reviewer agent provides auxiliary semantic and plausibility assessments, but final positive and negative labels are assigned by deterministic metric-based gates. These labels, rather than unconstrained LLM preferences, are passed to the experience-memory agent and used to guide the next iteration. Thus, the search controller tests whether LLMs are most useful when they shape the search space while numerical procedures retain authority over expression generation and retention.

**Equation author controlled variant.** The equation-author variant tests the consequence of granting the LLM direct structural-generation authority. It preserves the same outer-loop interface as LLM-PySR, including the same task context, data summaries, experience memory, validation feedback and unified evaluation protocol. However, instead of using PySR to enumerate candidate expressions, the LLM is asked at each iteration to propose a set of candidate equation skeletons directly.

Each proposed skeleton is parsed and sanitized before numerical fitting. Expressions that cannot be safely evaluated, contain invalid operations or violate basic syntactic constraints are discarded. For the remaining skeletons, numerical optimization is used only to fit scalar constants. The fitted formulas are then evaluated on the same training and validation splits as in the search-controller configuration.

Candidate review and feedback update follow the same LLM-assisted, metric-gated procedure used in LLM-PySR. Therefore, the equation-author variant primarily changes the expression-enumeration stage: structural search is no longer performed by PySR, but by repeated LLM proposal of candidate mathematical forms. This variant tests whether direct LLM equation authorship can replace numerical symbolic enumeration, or whether it prematurely restricts the structural search to formulas favoured by the LLM prior.

**Candidate decider controlled variant.** The candidate-decider variant tests the effect of granting the LLM final evaluative authority. It keeps the search-space planning and expression-enumeration stages identical to LLM-PySR: the LLM planner proposes a bounded search configuration, the configuration is sanitized, and PySR generates and fits candidate expressions within the resulting search space. Thus, the candidate pool available to the decider is generated by the same numerical symbolic-search procedure used in the search-controller configuration.

The difference lies in candidate acceptance. In LLM-PySR, Agent-B provides advisory semantic review and deterministic metric gates assign the final positive and negative labels. In the candidate-decider variant, the deterministic gating module is removed as the final authority. Instead, the LLM examines the candidate-review records produced from Agent-B, together with the numerical metrics reported by the unified evaluator, and determines which candidates should be retained. These LLM-assigned labels are passed directly to Agent-C and used to update the experience memory for subsequent search-space planning.

This design isolates LLM-based evaluative judgment from LLM-based structural generation. Because PySR still enumerates the candidate expressions, any performance difference between the candidate decider and the search controller can be attributed primarily to replacing metric-gated candidate retention with LLM-based acceptance.

**Full-stack LLM search.** The full-stack LLM setting represents the opposite extreme from LLM-PySR. In this role, the language model is responsible for most of the discovery loop, including hypothesis generation, refinement and selection. We use LLM-SR and DrSR as representative full-stack LLM-based symbolic-regression methods. These methods do not merely restrict the search space for a separate symbolic-regression engine; rather, they rely on the LLM to generate and iteratively refine candidate equation structures through repeated prompting and feedback.

The full-stack setting therefore tests whether end-to-end LLM-driven search can match or exceed a division-of-labour architecture in which the LLM controls only the search boundary while numerical procedures perform enumeration and selection. Because these methods are external baselines rather than controlled modifications of LLM-PySR, we use their recommended search settings whenever possible and report their language-model token footprint separately.

**No-LLM numerical search.** The no-LLM configuration represents a purely numerical symbolic-regression baseline. We instantiate this role using PySR without language-model planning, review or memory. The method searches within a fixed admissible space defined by the global input variables, primitive operator set and expression-complexity constraints. Candidate expressions are generated, optimized and selected using numerical procedures only.

This baseline tests whether the benefit of LLM-PySR arises from useful semantic restriction of the search space rather than from PySR alone. Compared with the search controller, the no-LLM baseline has no LLM-induced pruning bias, but it also lacks semantic guidance for selecting variables, operators, engineered transformations and search-depth schedules.

Together, these controlled variants distinguish three forms of LLM authority: planning authority, generative authority and evaluative authority. The search-controller design assigns the LLM only the first of these roles, while leaving expression enumeration and candidate retention under numerical control. The equation-author and candidate-decider variants selectively transfer generative or evaluative authority to the LLM, respectively, allowing the effect of each role assignment to be evaluated under otherwise matched data interfaces and evaluation metrics.

**Supplementary Note 3.** Hyperparameter settings for the experiments

This section summarizes the hyperparameter settings used for LLM-PySR, its controlled role-specification variants, and the external baseline methods. Unless otherwise specified, all LLM-based methods used GPT-5.1 as the base language model. Prediction accuracy was evaluated using the tolerance accuracy metric with relative-error threshold $\tau$=0.1. A test prediction was counted as accurate if its relative error was no larger than 0.1. Normalized mean squared error (NMSE) was computed as MSE/Var(y). Total LLM-token usage was computed as the sum of input and output tokens across all LLM calls made during the search.

**LLM-PySR search-controller configuration.** The main LLM-PySR configuration used a three-agent search-controller pipeline. The language model proposed bounded search specifications, whereas PySR performed symbolic enumeration and constant optimization. Deterministic numerical gates, rather than unconstrained LLM judgments, assigned the final candidate-retention labels. The main outer-loop and LLM settings Supplementary Table 1.

**Supplementary Table 1 | LLM-PySR outer-loop and LLM settings**

| Component | Hyperparameter | Value |
|---|---|---|
| LLM model | model | gpt-5.1 |
| LLM sampling | llm_temperature | 0.1 |
| Outer loop | max_iterations | 10 |
| Early stopping | target_score | 0.99 |
| Early stopping | target_acc | 0.99 |
| Early stopping | min_iterations_before_stop | 2 |
| Accuracy tolerance | acc_tau | 0.1 |
| Custom operators | max_custom_ops | 3 per iteration |
| Engineered features | max_engineered_features | 3 per iteration |
| Progressive feature engineering | progressive_feature_engineering | enabled |
| Post-search pruning | enable_post_ablation_prune | enabled |
| Maximum pruning steps | max_ablation_prune_steps | 12 |

The globally admissible primitive operator pool available to the LLM planner consisted of:

+, -, ×, /, ^, $\sqrt{\cdot}$, log, $log_{10}$, exp, sin, cos, tan, $|\cdot|$, min, max, arcsin, arccos, arctan, sinh, cosh, tanh, asinh, acosh, atanh.

The PySR component used the settings in Supplementary Table 2. The initial PySR search budget was 200 iterations and was adaptively adjusted between 100 and 500 iterations according to the size of the selected search space and recent search performance. The default maximum expression depth was 20, whereas the LLM-recommended maximum depth was constrained to the interval 6–40 before execution.

**Supplementary Table 2 | PySR settings inside LLM-PySR**

| PySR hyperparameter | Value |
|---|---|
| population_size | 128 |
| Initial niterations | 200 |
| Dynamic niterations range | 100–500 |
| fixed_pysr_niterations | False |
| maxsize | 30 |
| Default maxdepth | 20 |
| Allowed LLM-recommended maxdepth | 6–40 |
| Power constraint | ^: (-5, 5) |
| Verbosity | 0 |

The search loop also included deterministic scheduling and stabilization rules to prevent unstable search-space drift. Operator exploration was triggered when the selected operator set remained unchanged and recent accuracy improvement was small. Simplicity-guided post-processing allowed expression simplification only when the reduction in accuracy remained within the specified tolerance. Hyperparameter settings of this part is included in Supplementary Table 3.

**Supplementary Table 3 | LLM-PySR search-scheduling and simplification settings**

| Hyperparameter | Value |
|---|---|
| operator_stagnation_window | 2 |
| llm_b_stagnation_eps | 0.015 |
| val_acc_stagnation_eps | 0.01 |
| operator_same_jaccard_threshold | 1.0 |
| operator_explore_budget_default | 2 |
| operator_explore_budget_hard | 3 |
| operator_explore_hard_iter | 6 |
| operator_explore_hard_score | 0.85 |
| operator_explore_hard_acc | 0.85 |
| operator_explore_cooldown_rounds | 1 |
| operator_last_explore_round | 8 |

| Hyperparameter | Value |
| --- | --- |
| simplify_acc_drop_tol | 0.01 |
| simplify_val_acc_drop_tol | 0.01 |

**Standalone PySR baseline.** The no-LLM baseline used PySR with a fixed numerical search space and no language-model planning, review or memory. Its core search settings were matched to the PySR component used inside the controlled LLM-PySR variants wherever possible. Hyperparameter settings of standalone PySR is included in Supplementary Table 4.

**Supplementary Table 4 | Standalone PySR baseline settings**

| Hyperparameter | Value |
| --- | --- |
| niterations | 200 |
| population_size | 128 |
| populations | 1 |
| maxsize | 30 |
| maxdepth | 20 |
| random_state | 0 |
| Deterministic mode | True |
| Parallelism | serial |
| procs | 0 |
| Verbosity | 0 |
| Progress display | False |
| Timeout | disabled unless explicitly set |

The standalone PySR baseline used the following operator set:

Binary operators = $\{+, -, \times, /, \wedge\}$

Unary operators = $\{\sqrt{\cdot}, \log, \exp, \sin, \cos, \tanh, |\cdot|\}$

The power operator was constrained by (-3, 3).

**Full-stack LLM baselines.** We evaluated two full-stack LLM-based symbolic-regression baselines, LLM-SR and DrSR. These methods use the language model to generate or refine candidate equation structures directly, rather than using the language model only to constrain a separate symbolic-regression engine.

For LLM-SR, we used the original program-skeleton evolution framework with GPT-5.1 as the API model. The maximum number of generated samples was set to 2,500, with four samples generated per prompt. Hyperparameter settings of LLM-SR is included in Supplementary Table 5.

**Supplementary Table 5 | LLM-SR baseline settings**

| **Component** | **Hyperparameter** | **Value** |
|---|---|---|
| LLM model | api_model | gpt-5.1 |
| API mode | use_api | True |
| Maximum generated samples | max_samples | 2,500 |
| Samples per prompt | samples_per_prompt | 4 |
| Samplers | num_samplers | 1 |
| Evaluators | num_evaluators | 1 |
| Evaluation timeout | evaluate_timeout_seconds | 30 s |
| Experience-buffer islands | num_islands | 10 |
| Functions per prompt | functions_per_prompt | 2 |
| Island reset period | reset_period | 4 h |
| Cluster sampling temperature initialization | — | 0.1 |
| Cluster sampling temperature period | — | 30,000 |
| GPT-5.1 endpoint | — | Responses API |
| GPT-5.1 output limit | max_output_tokens | 512 |

For non-GPT-5 chat-completions fallback calls, the output limit was set to max_tokens = 512. GPT-5 output repair was enabled by default using gpt-4o-mini as the repair model with a maximum of 512 output tokens.

DrSR was evaluated using the modified DrSR framework. The main model was GPT-5.1, and the extractor model was GPT-4o-mini. DrSR was run for 10 iterations with four samples per iteration. Hyperparameter settings of DrSR is included in Supplementary Table 6.

**Supplementary Table 6 | DrSR baseline settings.**

| **Component** | **Hyperparameter** | **Value** |
|---|---|---|
| Main LLM model | — | openai/gpt-5.1 |
| Extractor model | — | openai/gpt-4o-mini |

| Component | Hyperparameter | Value |
|---|---|---|
| Iterations | niterations | 10 |
| Samples per iteration | samples_per_iteration | 4 |
| Total generated samples | — | 40 |
| Seed | seed | 0 |
| Samplers | num_samplers | 1 |
| Evaluators | num_evaluators | 1 |
| Evaluation timeout | — | 30 s |
| Islands | — | 10 |
| Functions per prompt | — | 2 |
| Reset period | — | 4 h |
| Cluster sampling temperature initialization | — | 0.1 |
| Cluster sampling temperature period | — | 30,000 |

The main DrSR LLM sampling configuration was:

temperature=0.6, top_p=0.3, frequency_penalty=0.5, n=1.

Streaming was disabled and the request timeout was 120 s. The maximum generation length was set to 4,090 tokens; for GPT-5.1 this value was passed as max_completion_tokens = 4090. The extractor model used temperature = 0, top_p = 0.3 and a maximum output length of 4,090 tokens.

**Equation author variant.** The equation author variant reused the outer LLM-PySR pipeline but replaced PySR-based expression enumeration with direct LLM generation of candidate equation skeletons. At each iteration, the language model proposed a set of candidate skeletons, which were subsequently fitted by numerical least-squares optimization. Thus, this variant transferred structural-generation authority from PySR to the language model while preserving the same general task interface and evaluation protocol.

The outer-loop settings matched LLM-PySR:

model=gpt-5.1, temperature=0.1, max_iterations=10, target_score=0.99, target_acc=0.99, τ=0.1.

The same limits on custom operators and engineered features were used:

max_custom_ops=3, max_engineered_features=3.

All hyperparameter settings of equation author is included in Supplementary Table 7.

**Supplementary Table 7 | Equation author settings**

| Hyperparameter | Value |
|---|---|
| AI-Feynman author_skeleton_count | 200 |
| Battery default author_skeleton_count | 20 |
| Skeleton constants | $(c_0, c_1, \ldots)$ |
| Maximum fitted constants | 8 |
| Constant optimizer | scipy.optimize.least_squares |
| Optimizer initializations | all-ones and all-zeros |
| max_nfev | 2,500 |
| Runner backend | llm_author_numeric_fit |
| PySR usage | disabled |

**Candidate decider variant.** The candidate decider variant reused the LLM-PySR outer loop and PySR-based candidate generation but replaced deterministic candidate-retention gates with direct LLM-based candidate labels. Thus, this variant preserved numerical expression enumeration but transferred final candidate-acceptance authority to the LLM reviewer.

The main settings matched LLM-PySR:

model=gpt-5.1, temperature=0.1, max_iterations=10, target_score=0.99, target_acc=0.99, $\tau$=0.1.

The PySR component used:

population_size=128, initial niterations=200, dynamic niterations $\in$ [100,500], maxsize=30, default maxdepth=20, LLM-recommended maxdepth $\in$ [6,40], maxsize=30, default maxdepth=20, LLM-recommended maxdepth $\in$ [6,40], maxsize=30, default maxdepth=20 with the power operator constrained to (−5,5).

The candidate-decider-specific differences were as follows:

metric hard gate for candidate survival=disabled,

LLM-B labels used directly=enabled,

post-ablation prune=disabled.

The constant-formula hard prefilter was disabled or relaxed relative to the main LLM-PySR pipeline. Consequently, differences between the candidate-decider variant and the search-controller configuration primarily reflect the replacement of deterministic numerical retention with LLM-based candidate acceptance.

**Supplementary Note 4.** Complete evaluation results across role specifications

This section provides the complete numerical results underlying the role-specification comparisons reported in the main text. We report results for six evaluated configurations representing five role specifications: search controller (LLM-PySR), equation author, candidate decider, full-stack LLM search represented by LLM-SR and DrSR, and no-LLM numerical search represented by PySR. The results cover three task categories: the AI-Feynman symbolic-regression benchmark, the complex-formula recovery stress-test suite and the lithium-ion battery cycle-life prediction task. For each task, we report the relevant predictive metrics, expression-level outcomes and language-model token usage where applicable, using the evaluation protocol and computational settings described in the Methods and Supplementary Note 3. These results are intended to make the aggregate trends shown in the main figures directly traceable to the underlying per-task and per-configuration results. Token usages are only recorded for problems in the AI-Feynman dataset group.

All experiment results

==================================================

Dataset group: ai_feynman

--------------------------------------------------------------------------------

Problem: I.10.7

Ground truth: m_0/sqrt(1-v**2/c**2)

Role: search controller

Formula: 0.556511650518845/(m_0*(m_0 + 0.80697)*(m_0**2 + m_0 + 0.45036688)*log(m_0))**0.4325141 + 0.9998649

ACC: 1

NMSE: 0.0330568911242

Token usage: 227493

Role: llmsr

Formula: 1 * m_0 / np.sqrt(np.abs(1 - 1 * m_0 ** 2))

ACC: 1

NMSE: 1.41406943028e-29

Token usage: 2404068

Role: drsr

Formula: (-0.0184493836694)*m_0 + (1.12646772451)

ACC: 0.992666666667

NMSE: 0.417091316677

Token usage: 833057

Role: pysr

Formula: -m_0 + m_0 + (cos(-1.2779886/(m_0*0.81068975 - 0.025795003))**(-2.1473446))**0.18475686 + 8.999759e-5

ACC: 1

NMSE: 1.77575114336e-07

Token usage: 0

Role: equation author

Formula: 0.991774398306 + 0.943184511048/m_0**3

ACC: 1

NMSE: 0.384788829816

Token usage: 176417

Role: candidate decider

Formula: ((m_0 - (1.0 / m_0))/m_0)**(-0.5)

ACC: 1

NMSE: 9.78949967644e-30

Token usage: 222881

Problem: I.11.19

Ground truth: x1*y1+x2*y2+x3*y3

Role: search controller

Formula: (x1 + x2 + x3)*((x1 + x2 + x3)*3.5334787**(y1 + y2)*(-1.0813193e-5) + (y1 + y2)*(1.8386748**(y1 + y2)*0.0020379394 + 0.2760418) + 0.43710488)

ACC: 0.808

NMSE: 0.0193507966018

Token usage: 277505

Role: llmsr

Formula: x1 * y1 + x2 * y2 + x3 * 1

ACC: 1

NMSE: 6.8005060866e-32

Token usage: 2487613

Role: drsr

Formula: x1*y1 + x2*y2 + x3*(1)

ACC: 1

NMSE: 1.1264435255e-27

Token usage: 5370644

Role: pysr

Formula: y1 + Abs(y1*y1) + Abs(y2*(y1 + 0.61043763) - 0.6104376*Abs(y2)) - 1*7.965893e-8

ACC: 0.150666666667

NMSE: 0.337742181591

Token usage: 0

Role: equation author

Formula: (0.0741245527893)*((x1 + x2 + x3) + (y1 + y2) + (1.58813446101))**2 + (-0.264652429514)

ACC: 0.756

NMSE: 0.0255110709491

Token usage: 674723

Role: candidate decider

Formula: 0.415497289192669*(x1 + x2 + x3) - 0.0890421270500248*(y1 + y2)*log(Abs(0.2581268*(y1 + y2) - 1.6454614)) + 0.29641294*((x1 + x2 + x3) * (y1 + y2))

ACC: 0.808

NMSE: 0.0196348161803

Token usage: 367685

Problem: I.12.1

Ground truth: mu*Nn

Role: search controller

Formula: (mu)

ACC: 1

NMSE: 1.98097516372e-32

Token usage: 14083

Role: llmsr

Formula: 1 * mu

ACC: 1

NMSE: 1.6304758773e-32

Token usage: 1409953

Role: drsr

Formula: (1) * mu

ACC: 1

NMSE: 8.42173916209e-26

Token usage: 2817843

Role: pysr

Formula: mu

ACC: 1

NMSE: 1.6304758773e-32

Token usage: 0

Role: equation author

Formula: mu

ACC: 1

NMSE: 1.98097516372e-32

Token usage: 131517

Role: candidate decider

Formula: mu

ACC: 1

NMSE: 1.98097516372e-32

Token usage: 22060

Problem: I.12.11

Ground truth: q*(Ef+B*v*sin(theta))

Role: search controller

Formula: -(-0.24217002)*q*Ef + ((q/Ef)*(-0.20604692) + (q*exp(-Ef)))*(exp(Ef*(Ef - 3.9435272)*(-0.16027664) + Ef) - 0.6608032) + 1.0002487

ACC: 1

NMSE: 8.71293080304e-07

Token usage: 215599

Role: llmsr

Formula: q * (-6.1942977 * Ef + 0.18101422 + 0.60481135 * Ef ** 2 + 15.188266 * np.tanh(0.44891642 * Ef))

ACC: 0.236666666667

NMSE: 0.0341940855638

Token usage: 1996290

Role: drsr

Formula: (-0.636640149892) * q * Ef + (2.00203453924) * q + (0.927181450856)

ACC: 0.237333333333

NMSE: 0.0499924193516

Token usage: 3345142

Role: pysr

Formula: q - (q - 1*1.0000001) + sin(Ef)*Abs(q)

ACC: 1

NMSE: 8.43705562199e-16

Token usage: 0

Role: equation author

Formula: (0.932930192796) + (-1.4779484783)*(q/Ef) + (-0.864629135597)*(q*Ef) + (3.27943370869)*q

ACC: 0.512

NMSE: 0.0244770251052

Token usage: 629566

Role: candidate decider

Formula: (q/Ef)*0.01817552 + (q*Ef)*((Ef*log(1.0/Ef))*(0.01771666*Ef**2*log(1.0/Ef**2) + (Ef*log(1.0/Ef))*(0.013291213*Ef*log(1.0/Ef) - 0.05243208) + 0.26935017) + 0.82348585) + 1.0001794

ACC: 0.996

NMSE: 1.0084349362e-07

Token usage: 100469

Problem: I.12.2

Ground truth: q1*q2*r/(4*pi*epsilon*r**3)

Role: search controller

Formula: 0.04579208*q1 + 0.03378539*(q1^2)/q1

ACC: 1

NMSE: 1.23515038843e-15

Token usage: 92683

Role: llmsr

Formula: q1 * 1 / (4.0 * np.pi * 1 * 1 ** 2)

ACC: 1

NMSE: 1.49666346933e-30

Token usage: 1909260

Role: drsr

Formula: (0.0795774715459) * q1

ACC: 1

NMSE: 1.49390713108e-25

Token usage: 3283374

Role: pysr

Formula: 0.07957747*Abs(Abs(Abs(Abs(Abs(Abs(Abs(Abs(q1))))))))

ACC: 1

NMSE: 1.23175190587e-15

Token usage: 0

Role: equation author

Formula: 0.0795774715459*q1 - 2.76187268551e-17

ACC: 1

NMSE: 1.17320883676e-24

Token usage: 246885

Role: candidate decider

Formula: q1*0.061423615 + (q1^2)*0.018153857*q1/(q1*q1)

ACC: 1

NMSE: 1.06547230581e-16

Token usage: 45541

Problem: I.13.12

Ground truth: G*m1*m2*(1/r2-1/r1)

Role: search controller

Formula: -m1 + (m1/m2)

ACC: 1

NMSE: 5.04776256175e-32

Token usage: 196584

Role: llmsr

Formula: 23.652959 * m1 * m2 * (1.0 / np.where(np.abs(-2.3741587e-05 * m1 + 1.0509055 * m2 + -0.021647818) < 1e-12, 1e-12, np.abs(-2.3741587e-05 * m1 + 1.0509055 * m2 + -0.021647818)) - 1.0 / np.where(np.abs(-2.6909588e-05 * m1 + 1.0061226 * m2 + 0.023154515) < 1e-12, 1e-12, np.abs(-2.6909588e-05 * m1 + 1.0061226 * m2 + 0.023154515)))

ACC: 0.992666666667

NMSE: 6.55776173982e-07

Token usage: 2143314

Role: drsr

Formula: (-1.12657017243)*m1*m2 + (1.57325822431)*m1 + (0.355990288159)*m2 + (-0.463759833238)

ACC: 0.0706666666667

NMSE: 0.499923920097

Token usage: 4278799

Role: pysr

Formula: m1*m2*(-1.290167e-9*(m1*m1*0.85080326/(m2*1.1583954) - (-m1 + exp(m2))) + (1.0 - m2)/m2)/m2

ACC: 1

NMSE: 3.85747604814e-16

Token usage: 0

Role: equation author

Formula: -m1 + m1/m2 - 5.50896127533e-17

ACC: 1

NMSE: 5.1717647379e-32

Token usage: 219369

Role: candidate decider

Formula: -m1 + (m1/m2) + (-m1/m2 + (m1/m2))/(-2.2574642)

ACC: 1

NMSE: 5.04776256175e-32

Token usage: 86182

Problem: I.13.4

Ground truth: 1/2*m*(v**2+u**2+w**2)

Role: search controller

Formula: m*m*0.5 - 1*2.0119255 + 2.5119255 - (-0.5)*(m*v^2)/m

ACC: 1

NMSE: 6.67699995898e-32

Token usage: 34186

Role: llmsr

Formula: -8.3226902e-07 * m ** 0.4549554 * v ** 0.4722072 + 0.50000106 * m ** 1.9999987 + 0.49999945 * v ** 2.0000006 + 0.50000017

ACC: 1

NMSE: 1.11503467753e-13

Token usage: 1998914

Role: drsr

Formula: (0.198480862452) * m * v**2 + (1.34819128178) * m + (0.225414788911)

ACC: 0.356666666667

NMSE: 0.0556956419287

Token usage: 2577666

Role: pysr

Formula: (m*0.6259615*m + (v*0.7910877)**2.0000765 - Abs(m) + Abs(m) + 0.6259769)*0.7988279

ACC: 1

NMSE: 1.93696222306e-09

Token usage: 0

Role: equation author

Formula: (0.5) + (0.5)*m**2 + (0.5)*v**2

ACC: 1

NMSE: 6.47402967001e-32

Token usage: 355577

Role: candidate decider

Formula: (m*m - 1.0331264 + (m*v^2)/m)*0.5 + 1.0165632

ACC: 1

NMSE: 7.05732292762e-32

Token usage: 74429

Problem: I.15.10

Ground truth: N/A

Role: search controller

Formula: (0.9469867 - (1/x1)/1.1138612)**((1/x1)*(-0.4178485))*0.9983383

ACC: 1

NMSE: 2.28878775601e-06

Token usage: 130307

Role: llmsr

Formula: -0.0019433847 * np.power(x1, 0.75006299) + 0.00085939736 * x1 + 1.002487 + 1.6205675 / (-2.6340272 + 3.2928628 * x1 ** 2 + 1e-12)

ACC: 1

NMSE: 9.64240324464e-07

Token usage: 1903344

Role: drsr

Formula: (1.27213808202) + (-0.0843091281038)*x1 + (0.00652560532314)*x1**2

ACC: 0.995333333333

NMSE: 0.170760313448

Token usage: 3753930

Role: pysr

Formula: ((cos(-0.0011446376 + 1.6726881/x1)**(-1.2810452))**0.23908037)**1.1767702 - 1*8.323509e-6

ACC: 1

NMSE: 2.01456063875e-06

Token usage: 0

Role: equation author

Formula: (1.00038474489999*x1**2 - 0.30463280989609)/(x1**2 - 0.797340129815)

ACC: 1

NMSE: 2.69249717446e-05

Token usage: 263457

Role: candidate decider

Formula: exp((((1/x1)*((1/x1) - 1*(-0.22783926))*(-(1/x1) + (1/x1) + (1/x1) - 0.011835406) - 1*(-1.3273357))*0.38285336*((1/x1)*(1/x1) - 0.00029379525))

ACC: 1

NMSE: 6.98856368767e-07

Token usage: 109499

Problem: I.15.3t

Ground truth: (t-u*x/c**2)/sqrt(1-u**2/c**2)

Role: search controller

Formula: 0.548663146504462*c*x**2 + 0.999429*c - 0.999429*x

ACC: 1

NMSE: 2.07033640346e-05

Token usage: 274919

Role: llmsr

Formula: (-39.110938 * x + 39.111138 * c + -3.9571999e-05) / np.sqrt(np.abs(1529.681 + 0.024240109 * (-63105.132 * x ** 2 + 2.5056273e-05 * c ** 2 + -0.0048305271 * x * c)) + 1e-12)

ACC: 1

NMSE: 7.69008813072e-15

Token usage: 3091249

Role: drsr

Formula: (0.775822182987)*x + (1.03007193762)*c + (-0.443431578139)

ACC: 0.966

NMSE: 0.000648099864631

Token usage: 1693471

Role: pysr

Formula: c + 0.18425639*sqrt(((1.1464193**c + c - 0.540304114)**1.7156944)**(2*x)) - tanh(x) - 0.244920245243679

ACC: 0.999333333333

NMSE: 7.44297808586e-06

Token usage: 0

Role: equation author

Formula: 0.792517169426*c*x**3 + 0.988815475691*c - 1.19865571926*x + 0.0364888628601

ACC: 1

NMSE: 0.00271186950732

Token usage: 332919

Role: candidate decider

Formula: (x*(x*((tanh(x))*x*0.9208847 - 1*(-0.005351937)) - 1*(-0.5114914))*(x*x*tanh(tanh(x)) + (x*c) - 1.9730119) + c)*0.99986887

ACC: 1

NMSE: 8.29496519902e-09

Token usage: 81827

Problem: I.15.3x

Ground truth: (x-u*t)/sqrt(1-u**2/c**2)

Role: search controller

Formula: (-(u - 1)*(x**2 - 0.74894035) + 0.348263423466737)/(x**2 - 0.74894035)

ACC: 0.992

NMSE: 0.00413743495921

Token usage: 212853

Role: llmsr

Formula: 0.00028879268 * (x - u * (3187.0399 + 42.855371 * x)) / np.sqrt(np.where(1.0 - u ** 2 / np.log1p(np.exp(331.88164)) <= 0.0, np.finfo(float).eps, 1.0 - u ** 2 / np.log1p(np.exp(331.88164)))) + 1.0155968

ACC: 0.989333333333

NMSE: 0.0232821506335

Token usage: 2930547

Role: drsr

Formula: ((-0.00345838823763)*x + (-1.04886163848)*u + (1.04869610445)) / np.sqrt(1 - (0.206687528127)*u**2)

ACC: 0.986

NMSE: 0.0236721180846

Token usage: 2743501

Role: pysr

Formula: -u + (2.3477335/(u + log(x + log((x/(0.029605785**re(u) + x))**1.469732))))**(0.1958292/x) + 0.00549959999999983

ACC: 1

NMSE: 8.47814285448e-05

Token usage: 0

Role: equation author

Formula: (-0.378373650696)*log(u*u + (0.130532690459)) + (0.141521807637)

ACC: 0.976

NMSE: 0.0425722033714

Token usage: 605511

Role: candidate decider

Formula: (1.785267e-5*x + 0.99928033)*(0.296013441991912*(1 - u)/(x - 0.56212294)**1.767653 + (1 - u))

ACC: 1

NMSE: 2.78743757657e-06

Token usage: 83007

Problem: I.16.6

Ground truth: (u+v)/(1+u*v/c**2)

Role: search controller

Formula: (c + v)/((c*v) + 1.0)

ACC: 1

NMSE: 1.04230336918e-30

Token usage: 41011

Role: llmsr

Formula: -0.55941743 * (2.1369066 + (0.52806949 * v + -0.51268682)) / (1.0 + 2.1369066 * (0.52806949 * v + -0.51268682) / ((1.0986992 * c + 0.99040695) ** 2 + 1e-12)) + 2.1905149

ACC: 0.950666666667

NMSE: 0.09078006754

Token usage: 2172335

Role: drsr

Formula: ((0.999999999993)*c + (1)*v) / (1 + ((0.780136202743)/(-0.883253192886)**2)*c*v)

ACC: 1

NMSE: 1.89237751549e-22

Token usage: 3856387

Role: pysr

Formula: 0.3638521 + 1/((0.909352154008556*((v**1.941972)**log(c))**0.4010189 + 0.664364)*((v**3.3756464)**log(c))**0.08377571)

ACC: 1

NMSE: 0.000513468514087

Token usage: 0

Role: equation author

Formula: (-1.43924116543)*log((0.788081327184) + v) + (0.605623433828)*log((-0.179664403233) + c) + (1.99984482952)*log((1.59740546324) + (v/c))

ACC: 0.952

NMSE: 0.0880965916924

Token usage: 879617

Role: candidate decider

Formula: (-0.00202387462453274*(log(c)*log(v))**2 + 0.035340034*(log(c)*log(v)) + 0.017670017*log((v/c))**2 + 0.0882449107409221)**(0.2058618*(log(c)*log(v)))

ACC: 1

NMSE: 3.04946859755e-08

Token usage: 259783

Problem: I.18.12

Ground truth: r*F*sin(theta)

Role: search controller

Formula: sin(r)

ACC: 1

NMSE: 4.10197511618e-32

Token usage: 37079

Role: llmsr

Formula: r * (-1.028954 * (1.0 + -0.15385866 * r + -0.0009851815 * r ** 2)) * np.sin(4.3720904 + 0.60894871 * r) + 0.003687825

ACC: 0.994666666667

NMSE: 1.30171510295e-06

Token usage: 1640043

Role: drsr

Formula: (-2.04972991961e-12)*r + (-0.999999999999)*np.sin((-1)*r) + (5.55017911494e-12)

ACC: 1

NMSE: 1.04054424712e-23

Token usage: 2664173

Role: pysr

Formula: -4.7519912e-9*(-r*cos(r + 0.94886833) + r)*cos(r*(r - cos(r*(r + 0.88444185)) - 0.34979773)) + sin(r)

ACC: 1

NMSE: 2.33970485833e-16

Token usage: 0

Role: equation author

Formula: (1.17858639619e-17) + (1)*sin((1)*r)

ACC: 1

NMSE: 4.09016030142e-32

Token usage: 135797

Role: candidate decider

Formula: sin(r)

ACC: 1

NMSE: 4.10197511618e-32

Token usage: 57997

Problem: I.18.14

Ground truth: m*r*v*sin(theta)

Role: search controller

Formula: sin((m))

ACC: 1

NMSE: 3.74396362456e-32

Token usage: 35865

Role: llmsr

Formula: 2.9443523 * m + -1.2489133 * m ** 2 + 0.13277453 * m ** 3 + -1.0422548

ACC: 0.994

NMSE: 0.000198786171338

Token usage: 1652161

Role: drsr

Formula: (-0.360299970221)*m + (-0.0490439924564)*m**2 + (1.64985812921)

ACC: 0.202666666667

NMSE: 0.0420795243733

Token usage: 3224210

Role: pysr

Formula: sin(m) - 4.6038623e-10/(-1.79074852751225*sin(cos(m)) - Abs(0.408422788815358*m - exp(cos(cos(m)))))

ACC: 1

NMSE: 1.66410054702e-14

Token usage: 0

Role: equation author

Formula: sin(m)

ACC: 1

NMSE: 3.74396362456e-32

Token usage: 164312

Role: candidate decider

Formula: sin((m))

ACC: 1

NMSE: 3.74396362456e-32

Token usage: 51196

Problem: I.18.4

Ground truth: (m1*r1+m2*r2)/(m1+m2)

Role: search controller

Formula: m2 - (m2 - 1.0)/(m1 + 1.0)

ACC: 1

NMSE: 2.46432192688e-31

Token usage: 34718

Role: llmsr

Formula: (np.abs(1.2630128) * m1 * (9.3389608e-06 * m1 + 1.0000417 * m2 + -6.4190521e-05) + np.abs(1.1394165e-05) * m2 * (0.71286467 * m1 + 0.80955041 * m2 + 0.58653743) + 1.2631996) / (np.abs(1.2630128) * m1 + np.abs(1.1394165e-05) * m2 + (1e-12 + np.abs(1.2631967)))

ACC: 1

NMSE: 3.09615245416e-10

Token usage: 2624077

Role: drsr

Formula: (0.0382975284867)*m1 + (0.506344753968)*m2 + (0.447014992872)

ACC: 0.780666666667

NMSE: 0.082246752346

Token usage: 2352649

Role: pysr

Formula: log((-m1/tanh(tanh(m2) - 2.3420203/tanh(m2)))**(0.21564841*m2 - 0.21712689549896)*(m2 + m2**(0.29388472*m2) + 0.73123384))**1.0610108

ACC: 1

NMSE: 0.000534252358725

Token usage: 0

Role: equation author

Formula: (0.702610726085) + (-0.180988390233)*(m1 + m2) + (0.177150861956)*(m1 * m2) + (0.482618813875)*m2

ACC: 0.98

NMSE: 0.0228337555338

Token usage: 699963

Role: candidate decider

Formula: m2*(1.0 - m2)/(m2 + (m1 * m2)) + m2

ACC: 1

NMSE: 2.68714185819e-31

Token usage: 58293

Problem: I.24.6

Ground truth: 1/2*m*(omega**2+omega_0**2)*1/2*x**2

Role: search controller

Formula: (m^2)*0.25 - 1*(-0.25)

ACC: 1

NMSE: 1.711822564e-31

Token usage: 22311

Role: llmsr

Formula: 0.24999999 * m ** 2 + 0.25

ACC: 1

NMSE: 1.39481347241e-15

Token usage: 1925257

Role: drsr

Formula: ((0.249999999994) + (8.28099787339e-12)*m + (0.249999999997)*m**2)

ACC: 1

NMSE: 1.160956464980e-23

Token usage: 3216269

Role: pysr

Formula: (m*0.49617767)**2.0006974 + tanh(Abs(tanh(((m + m)**1.9554809 - 1*0.3542039)*0.001013153))) + 0.25035742

ACC: 1

NMSE: 4.36505245495e-13

Token usage: 0

Role: equation author

Formula: 0.25*m**2 + 0.25

ACC: 1

NMSE: 1.711822564e-31

Token usage: 268815

Role: candidate decider

Formula: 0.25 - (-0.25)*(m^2)

ACC: 1

NMSE: 1.711822564e-31

Token usage: 37412

Problem: I.26.2

Ground truth: arcsin(n*sin(theta2))

Role: search controller

Formula: n*sin(theta2)/cos(1.51743065000508*n*sin(theta2))**0.117128134

ACC: 1

NMSE: 0.000720026568902

Token usage: 222520

Role: llmsr

Formula: np.arcsin(n * np.sin(theta2))

ACC: 1

NMSE: 6.34687123404e-32

Token usage: 1763281

Role: drsr

Formula: np.arcsin(n * np.sin(theta2))

ACC: 1

NMSE: 6.34687123404e-32

Token usage: 4658929

Role: pysr

Formula: sin(sin(n*sin(theta2)*cos(sin(theta2*sin(n)/n)*tanh(sin(sin(n))*0.29134652))))/cos(n*sin(theta2/1.0063514))

ACC: 1

NMSE: 0.000148532230077

Token usage: 0

Role: equation author

Formula: (112.59425522)*sin((0.814123492775)*(n * sin(theta2))) + (-90.6509581922)*(n * sin(theta2)) + (10.149238147)*(n * sin(theta2))**3

ACC: 1

NMSE: 0.000527363160335

Token usage: 823839

Role: candidate decider

Formula: (n * sin(theta2))/((0.42852435 - cos((n * sin(theta2))/(-0.12562138)))*7.0495684e-5 + ((n * sin(theta2))*(n * sin(theta2))*(n * sin(theta2))*4.833402e-5 + cos((n * sin(theta2))*1.3278909)*cos((n * sin(theta2))*1.5401679))**0.081007995)

ACC: 1

NMSE: 1.4051683118e-05

Token usage: 177334

Problem: I.27.6

Ground truth: 1/(1/d1+n/d2)

Role: search controller

Formula: (d1 / (d1 + d2))

ACC: 1

NMSE: 1.89928442641e-31

Token usage: 20017

Role: llmsr

Formula: (54.96863 * np.where(d1 == 0, 1e-12, d1) + 0.045954705 * np.where(d2 == 0, 1e-12, d2)) / (1.0 / np.where(d1 == 0, 1e-12, d1) + -1.8457258 / np.where(d2 == 0, 1e-12, d2) + 55.147017 * np.where(d1 == 0, 1e-12, d1) + 54.852565 * np.where(d2 == 0, 1e-12, d2))

ACC: 1

NMSE: 5.49708522655e-05

Token usage: 2343543

Role: drsr

Formula: (0.144281140604)*d1 + (-0.0683592277186)*d2 + (0.319294616983)

ACC: 0.554666666667

NMSE: 0.0821572844757

Token usage: 2273857

Role: pysr

Formula: Abs(sin(tanh((d1/d2 + tanh((((d1*1.5744494/d2)**1.3395746)**1.0727283)**0.5388396))*0.3243752)**1.0712736))

ACC: 1

NMSE: 1.88666756255e-05

Token usage: 0

Role: equation author

Formula: (-1.94678438981e-17) + (1)*(d1/(d1 + d2))

ACC: 1

NMSE: 1.63503704318e-31

Token usage: 235876

Role: candidate decider

Formula: (d1 / d2)/((d1 / d2) + 1.0)

ACC: 1

NMSE: 1.88748130879e-31

Token usage: 56104

Problem: I.29.16

Ground truth: sqrt(x1**2+x2**2-2*x1*x2*cos(theta1-theta2))

Role: search controller

Formula: 1.7103646 - cos((x1 * cos(theta1)))

ACC: 0.176

NMSE: 0.956495957259

Token usage: 114389

Role: llmsr

Formula: 0.60991721 * np.sqrt(np.maximum((1.639566 * x1 + 1.0809302e-06) ** 2 + (-1.1137104e-07 * x2 + 1.6395672) ** 2 - 2.0 * (1.639566 * x1 + 1.0809302e-06) * (-1.1137104e-07 * x2 + 1.6395672) * np.cos(theta1 - (0.99999981 * x2 + 4.34131e-07)), 0.0)) + 1.4847075e-07

ACC: 1

NMSE: 2.06920144069e-13

Token usage: 3273284

Role: drsr

Formula: np.sqrt(np.abs((0.0964392826535)*x1**2 + (-0.234820263583)*x2**2 - (-0.722956772092)*x1*x2*np.cos(theta1 + (1.89231968473)) + (2.01585087331)))

ACC: 0.219333333333

NMSE: 0.403531867652

Token usage: 4688748

Role: pysr

Formula: Abs(1.94664609035794*(theta1*(-2.424934*cos(0.7028877*theta1) + 2.424934*Abs(theta1 - Abs(theta1))) + 2.6146412)**0.326922835 - 1.83044825458456)

ACC: 0.0653333333333

NMSE: 2.66547189456

Token usage: 0

Role: equation author

Formula: (1.45792395287) + (-0.806334655287)*cos(x2 - theta1)

ACC: 0.24

NMSE: 0.559348185584

Token usage: 880694

Role: candidate decider

Formula: (-cos((-(x1 * sin(theta1)) + sin(x2))*0.57857496) + 2.0996337 + sin(-(x1 * cos(theta1)) + cos(x2) - 1*1.5824449)/(2.1122534 - cos(((x1 * sin(theta1)) - sin(x2))/0.9239847)))*1.5185791

ACC: 0.896

NMSE: 0.0292010770257

Token usage: 252726

Problem: I.30.3

Ground truth: Int_0*sin(n*theta/2)**2/sin(theta/2)**2

Role: search controller

Formula: (cos(Int_0*theta) - 1.0)/(cos(Int_0) - 1*1.0)

ACC: 1

NMSE: 6.93925988023e-31

Token usage: 42803

Role: llmsr

Formula: 0.13255171 * Int_0 * (np.sin(Int_0 * theta / 2.0) / (np.sin(theta / 2.0) + 1e-12)) ** 2 + 1.5083217 + -0.30216791 * Int_0 + -0.0092870277 * theta

ACC: 0.141333333333

NMSE: 0.727647640734

Token usage: 2637008

Role: drsr

Formula: (-0.228403910323)*Int_0 + (-0.0680540932062)*theta + (1.79882601858)

ACC: 0.121333333333

NMSE: 0.927850850018

Token usage: 2404280

Role: pysr

Formula: (tanh(cos(-Int_0 + sin(tanh(tanh(sin(Abs(-theta + sin(Abs(cos(Int_0))) + 2*cos(Int_0))))))) - 1*1.5134357) + 1.7751732)**6.198205 - 1*(-0.30971614)

ACC: 0.143333333333

NMSE: 0.415449161902

Token usage: 0

Role: equation author

Formula: (1.47809224717) + (-0.0617469671436)*Int_0*theta + (-0.1094294182)*(Int_0*sin(theta))

ACC: 0.176

NMSE: 0.873921839149

Token usage: 921607

Role: candidate decider

Formula: (1.0 - cos(Int_0*theta))/(1.0 - cos(Int_0))

ACC: 1

NMSE: 6.93925988023e-31

Token usage: 67247

Problem: I.30.5

Ground truth: arcsin(lambd/(n*d))

Role: search controller

Formula: asin(1.0/(lambd*d))

ACC: 1

NMSE: 2.87032363959e-31

Token usage: 32790

Role: llmsr

Formula: np.arcsin(np.clip((((1.0 + -0.99996539) * lambd + 4.5968106) / ((np.abs(0.00065282601) + 1e-09 + 3.3891966 * lambd + -0.00010282878 * d) * ((1.0 + 0.35607349) * d + 0.0001445274 + 1e-12) + 1e-12), -1.0, 1.0)) + -1.5990618e-05

ACC: 1

NMSE: 5.91098305512e-09

Token usage: 2701221

Role: drsr

Formula: (-0.081178856795)*lambd + (-0.0759522582068)*d + (0.612588261306)

ACC: 0.251333333333

NMSE: 0.262757153949

Token usage: 3024407

Role: pysr

Formula: 1.04379528134315*Abs(1/((1.05005320094543*lambd - 0.0665561/(d*sqrt(tanh((d - d**(1.0567045/d)/lambd)**1.2292547))))*Abs(d)))

ACC: 0.999333333333

NMSE: 0.00787694904717

Token usage: 0

Role: equation author

Formula: 18.6817256265317/(d*(19.0006057021*lambd - 1.05188144258646))

ACC: 1

NMSE: 0.00219163003111

Token usage: 779912

Role: candidate decider

Formula: asin((1/(d*lambd)))

ACC: 1

NMSE: 2.87032363959e-31

Token usage: 49182

Problem: I.32.17

Ground truth: (1/2*epsilon*c*Ef**2)*(8*pi*r**2/3)*(omega**4/(omega**2-omega_0**2)**2)

Role: search controller

Formula: (-39.16474**(epsilon/c)*(0.0043715024*c**3 - epsilon**3)*(0.06672978*33.0629090916447**(epsilon**2/c**2)*epsilon + 0.53874665*c) + 0.003023708*c**4)/c**4

ACC: 1

NMSE: 0.000317683649541

Token usage: 221805

Role: llmsr

Formula: 1.8033919 * np.power(epsilon, 3.0150625) * np.power(c, -0.14598942) * np.power(np.abs(-1.2846162 * epsilon + 1.4580775 * c), 0.49829235) / np.power((-1.2846162 * epsilon + 1.4580775 * c - 0.10321213) ** 2 + 0.12673456, 1.6521703) + -0.0051034866

ACC: 0.95

NMSE: 2.21648684147e-05

Token usage: 2268525

Role: drsr

Formula: ((-0.459644642796) * epsilon * c + (2.04954091189) * epsilon + (1.83970377306) * c + (-9.7758624815)) * ((-0.0207963937305) + (-0.417501064625) * epsilon**2 + (-0.0340181645577) * c**2 + (0.243876516184) * epsilon * c)

ACC: 0.160666666667

NMSE: 0.0860171550997

Token usage: 3296974

Role: pysr

Formula: c*(0.44629234*epsilon/c + 0.8206506)**20.783756*tanh(epsilon*tanh(sin(sin(0.380985033935125/(c**2*sqrt(tanh(c)))))))

ACC: 0.845333333333

NMSE: 0.000147067503571

Token usage: 0

Role: equation author

Formula: (-1.56178934817) + (14.3158644944)*(epsilon / c) + (-43.712900762)*(epsilon / c)**2 + (46.4456190833)*(epsilon / c)**3

ACC: 0.684

NMSE: 0.00313232731883

Token usage: 632702

Role: candidate decider

Formula: (epsilon / c)*(-0.0016248347 + exp(-1.6351923*(epsilon / c)))*(-0.0003679246*(epsilon / c)/(epsilon + 1.8052094) + 4.21716296695648*(-(epsilon / c)/log((epsilon / c)))**2.1078687 - 0.00124843435079141*exp((epsilon / c)))

ACC: 1

NMSE: 2.59782924006e-09

Token usage: 255741

Problem: I.34.1

Ground truth: omega_0/(1-v/c)

Role: search controller

Formula: c/(c - (c^2))

ACC: 1

NMSE: 7.74411690782e-31

Token usage: 27766

Role: llmsr

Formula: 1 / (1 - 1 * c)

ACC: 1

NMSE: 7.46165550485e-31

Token usage: 1822728

Role: drsr

Formula: (2.28903640374)*c + (0.791800748771)

ACC: 0.985333333333

NMSE: 0.0591951470987

Token usage: 2030302

Role: pysr

Formula: exp(Abs(c*exp(c*(0.24089126*c*exp(0.6692112*0.329027383222828**(0.6192171 - c)*c) + 0.491311286947938)) + 0.00019104838))

ACC: 1

NMSE: 9.8706835275e-08

Token usage: 0

Role: equation author

Formula: ((-1.05562430904) + (-1.05562430904)*c)/((-1.05562430904) + (1.05562430904)*(c^2))

ACC: 1

NMSE: 9.89969584092e-31

Token usage: 194555

Role: candidate decider

Formula: c/(c - (c^2))

ACC: 1

NMSE: 7.74411690782e-31

Token usage: 39390

Problem: I.34.14

Ground truth: (1+v/c)/sqrt(1-v**2/c**2)*omega_0

Role: search controller

Formula: 2.04233791776762**(c**2) + c

ACC: 1

NMSE: 0.0109329606404

Token usage: 174612

Role: llmsr

Formula: 1.6076562 * c + 0.9042283

ACC: 1

NMSE: 0.0193776398921

Token usage: 2328949

Role: drsr

Formula: (((0.855440205872) + (0.8554402059)*c) / np.sqrt(1.0 - (0.999999999967)*c**2)) * (1.16898877692)

ACC: 1

NMSE: 5.08458958379e-23

Token usage: 3262504

Role: pysr

Formula: exp(c/cos(1.5125988*c*cos(cos(0.580287250914152*c*sqrt(cos(cos(1.2790442*c)))))))

ACC: 1

NMSE: 5.98884178947e-09

Token usage: 0

Role: equation author

Formula: (0.35736540923*c + (c**3 + 0.327753283021)*(3.19659074851*c**3 + 0.995972030081))/(c**3 + 0.327753283021)

ACC: 1

NMSE: 0.0194303614114

Token usage: 169251

Role: candidate decider

Formula: (c*c*(c*(c^2)*(1.189076 + (c - 0.54759514)*(-0.028038971))*(c - 0.567854)*exp(c) + (c^2) + 0.89237726) + 2.718544)**c

ACC: 1

NMSE: 5.34011593325e-09

Token usage: 209902

Problem: I.37.4

Ground truth: I1+I2+2*sqrt(I1*I2)*cos(delta)

Role: search controller

Formula: 0.8499987**log(I2) + (I1*I2)**0.51106185*((0.5472272 - 0.02131326*I2**log(I2))*(I2 - 4.3945546)*(I2 - 1.7673781) + log((I1/I2) + 1.0947772))

ACC: 0.924

NMSE: 0.000118856372566

Token usage: 233977

Role: llmsr

Formula: 1.0000021 * I1 + -1.6948353e-06 * I2 + 1.9999746 * (np.power(np.maximum(I1, 0.0), 0.5 + 1.1621737e-06) * np.power(np.maximum(I2, 0.0), 0.5 + -0.49998998)) * np.cos(-1.0518328e-07 * I1 + 1.0000004 * I2) + 1.0000022

ACC: 1

NMSE: 1.16513509224e-11

Token usage: 2795469

Role: drsr

Formula: (2.33020179842)*I1 + (0.579281732805)*I2 + (-2.22855364379)*np.sqrt(I1*I2) + (0.693015851323)

ACC: 0.078

NMSE: 0.667488347802

Token usage: 4370126

Role: pysr

Formula: Abs(Abs(I2 - 3.1757476)**(0.96274704**I1)*(I1 + 0.25996137*Abs((I1 + 3.15051)*(I2 - 3.0613482)))*0.47797874 - 0.024441684)

ACC: 0.586666666667

NMSE: 0.00818527208251

Token usage: 0

Role: equation author

Formula: (-1.00071363812) + (0.0700247652587)*(I1 + (2.81592709965))**2 + (0.826532320592)*(I2 + (-3.12932418793))**2

ACC: 0.316

NMSE: 0.0837689278239

Token usage: 919468

Role: candidate decider

Formula: (I1**0.5259134 + 0.09775204)*(0.09111284*(I2 - 3.14159265)**2 + log(0.00573234570860954*log((I1^2))**2 + 1))*(-0.029026983*I1 + (0.122587875 - 0.0071255304*log(0.12356315*(I2 - 3.14159265)**2))*(I2 - 3.14159265)**2 - 3.14159265)**2

ACC: 1

NMSE: 5.48465297099e-06

Token usage: 88372

Problem: I.39.11

Ground truth: 1/(gamma-1)*pr*V

Role: search controller

Formula: (1.0/(gamma - 1.0))

ACC: 1

NMSE: 2.51841152098e-31

Token usage: 22242

Role: llmsr

Formula: 1 / (gamma - 1.0)

ACC: 1

NMSE: 1.99613236239e-31

Token usage: 1642267

Role: drsr

Formula: (1) / (gamma - 1.0)

ACC: 1

NMSE: 1.67398001365e-27

Token usage: 3277150

Role: pysr

Formula: gamma/(gamma*(gamma - 0.99999994))

ACC: 1

NMSE: 1.09870206913e-14

Token usage: 0

Role: equation author

Formula: (-1) + (1)*gamma/((-1) + gamma)

ACC: 1

NMSE: 4.01920908435e-31

Token usage: 174544

Role: candidate decider

Formula: (1/gamma) + ((1/gamma)/(gamma - 1.0))

ACC: 1

NMSE: 2.78196621504e-31

Token usage: 54299

Problem: I.39.22

Ground truth: n*kb*T/V

Role: search controller

  Formula: n

  ACC: 1

  NMSE: 1.47059121614e-31

  Token usage: 14018

Role: llmsr

  Formula: 1 * n

  ACC: 1

  NMSE: 1.37309181238e-31

  Token usage: 1525558

Role: drsr

  Formula: (1) * n

  ACC: 1

  NMSE: 1.13021140276e-25

  Token usage: 2767068

Role: pysr

  Formula: n

  ACC: 1

  NMSE: 1.37309181238e-31

  Token usage: 0

Role: equation author

  Formula: n

  ACC: 1

  NMSE: 1.47059121614e-31

  Token usage: 233884

Role: candidate decider

  Formula: n

  ACC: 1

  NMSE: 1.47059121614e-31

  Token usage: 21957

Problem: I.40.1

Ground truth: n_0*exp(-m*g*x/(kb*T))

Role: search controller

Formula: n_0/exp(1.0/m)

ACC: 1

NMSE: 7.59328730685e-32

Token usage: 102144

Role: llmsr

Formula: n_0 * np.exp(-0.15856177 * m) * (1.0 + -0.17682292 * n_0 + 0.53108426 * m) + -1.3970285

ACC: 0.165333333333

NMSE: 0.128512794894

Token usage: 2046053

Role: drsr

Formula: (0.16185394918)*n_0 + (0.641687479383)*m + (-0.407196344329)

ACC: 0.088

NMSE: 0.269686620197

Token usage: 2292219

Role: pysr

Formula: 0.900577179914607*n_0*tanh(0.88789976*(sqrt(m) - 0.5256067)*tanh(1.875866*m) + 0.032344017)

ACC: 0.712

NMSE: 0.000132640222382

Token usage: 0

Role: equation author

Formula: -0.0419471674014*m**2 + 0.432180017484*m*n_0 - 0.0178534691714*m**(n_0 + 1) - 0.0111337902285*n_0**2 - 0.092133244996449

ACC: 0.356

NMSE: 8.05784950922

Token usage: 875229

Role: candidate decider

Formula: (m*n_0)*exp(-1.0/m)/m

ACC: 1

NMSE: 7.47731245521e-32

Token usage: 56671

Problem: I.41.16

Ground truth: h/(2*pi)*omega**3/(pi**2*c**2*(exp((h/(2*pi))*omega/(kb*T))-1))

Role: search controller

Formula: 0.10132118*omega - 0.008062881 + 0.000213472140718468/omega

ACC: 1

NMSE: 9.91990425776e-10

Token usage: 184790

Role: llmsr

Formula: 0.44947308 * omega ** 0.99517794 / (np.expm1(np.clip(-0.00045630863 * omega + 1.6876938 + 0.072332023 / (omega + 1e-12), -700.0, 700.0)) + 1e-12)

ACC: 1

NMSE: 3.06890531576e-07

Token usage: 2954518

Role: drsr

Formula: (0.101209191285)*omega + (-0.00757790637475)

ACC: 0.987333333333

NMSE: 2.6892793848e-06

Token usage: 2324283

Role: pysr

Formula: 0.10130378*omega - 0.007927855*tanh(omega + sqrt(tanh(tanh((omega - 0.07782935)/(omega**sin(omega) + 0.043954488)))))**0.20084158

ACC: 1

NMSE: 4.97164615697e-09

Token usage: 0

Role: equation author

Formula: 0.10118961867*omega - 0.00772087802327 + 0.00308010977163659*exp(-6.68379420741*omega)

ACC: 1

NMSE: 9.21759301119e-07

Token usage: 175465

Role: candidate decider

Formula: omega*0.10132118 - (0.008062888 - 0.0004277676/(omega + omega + 0.0004319953/(omega + 0.007408765) + 0.00042778463/omega))

ACC: 1

NMSE: 3.10452451729e-15

Token usage: 194721

Problem: I.43.43

Ground truth: 1/(gamma-1)*kb*v/A

Role: search controller

Formula: (1/gamma) + gamma*(1/gamma)/(gamma*(gamma - 1*1.0))

ACC: 1

NMSE: 3.11749575446e-31

Token usage: 90036

Role: llmsr

Formula: 1 / (gamma - 1)

ACC: 1

NMSE: 2.56576597277e-31

Token usage: 1653157

Role: drsr

Formula: (1) / (gamma - (1))

ACC: 1

NMSE: 6.69519418628e-26

Token usage: 3419375

Role: pysr

Formula: 1/((gamma/(gamma - 1*1.0530558))**4.210868e-8*(gamma - sqrt(cos(-gamma + gamma))))

ACC: 1

NMSE: 3.41420545403e-15

Token usage: 0

Role: equation author

Formula: (-6.01931137497e-17) + (1)*(1/gamma)/(1 + (-1)*(1/gamma))

ACC: 1

NMSE: 4.22422283344e-31

Token usage: 112227

Role: candidate decider

Formula: (1/gamma) + gamma*(1/gamma)*(1/gamma)/(gamma*(1/gamma)*(gamma - 1*1.0))

ACC: 1

NMSE: 3.06923730625e-31

Token usage: 41960

Problem: I.44.4

Ground truth: n*kb*T*ln(V2/V1)

Role: search controller

Formula: 0.25*n*(log(kb) + log(kb*(kb^2)))

ACC: 1

NMSE: 5.75523686309e-32

Token usage: 39652

Role: llmsr

Formula: -160.06388 * n * kb * np.log(np.abs((1.0143068e-05 * n ** -0.23695255 + -3.4399947) / (0.0092690854 * kb ** -1.9537035 + -3.4490342)) + 1e-12)

ACC: 0.365333333333

NMSE: 0.00870277531041

Token usage: 1889874

Role: drsr

Formula: (0.715603651869)*n*kb + (-0.858254710425)*n + (-0.0192160198676)*kb + (-0.000498695890964)

ACC: 0.126666666667

NMSE: 0.131165586336

Token usage: 3421113

Role: pysr

Formula: sqrt(n)*(9.95173256370919e-9*n - 2.8968663e-9*kb)*tanh(sqrt(n)*log(kb)) + log(kb**n)

ACC: 1

NMSE: 3.60175543412e-16

Token usage: 0

Role: equation author

Formula: (5.39969139883e-17) + (1)*n*log(kb)

ACC: 1

NMSE: 5.65241928955e-32

Token usage: 341964

Role: candidate decider

Formula: n*(0.5*log(kb) + 0.25*log((kb^2)))

ACC: 1

NMSE: 5.6140031334e-32

Token usage: 65378

Problem: I.47.23

Ground truth: sqrt(gamma*pr/rho)

Role: search controller

Formula: gamma**0.5

ACC: 1

NMSE: 6.3693895988e-31

Token usage: 46951

Role: llmsr

Formula: 1 * np.sqrt(gamma)

ACC: 1

NMSE: 6.16733387942e-31

Token usage: 1684319

Role: drsr

Formula: (0.999999999999)*np.sqrt(gamma) + (9.94507376182e-13)

ACC: 1

NMSE: 3.26389485548e-24

Token usage: 2983960

Role: pysr

Formula: sqrt(gamma)

ACC: 1

NMSE: 6.16733387942e-31

Token usage: 0

Role: equation author

Formula: (1)*gamma**(0.5)

ACC: 1

NMSE: 6.3693895988e-31

Token usage: 85767

Role: candidate decider

Formula: sqrt(gamma)

ACC: 1

NMSE: 6.3693895988e-31

Token usage: 61875

Problem: I.48.20

Ground truth: N/A

Role: search controller

Formula: x1**2 + 0.51011044

ACC: 1

NMSE: 4.89118947431e-06

Token usage: 34009

Role: llmsr

Formula: 0.49923605 + 0.00014503097 * x1 + 0.99999167 * x1 ** 2 + 0.38813843 / ((x1 - -0.060946761) ** 2 + -0.99433131 + 1e-12)

ACC: 1

NMSE: 9.34120072975e-14

Token usage: 2174663

Role: drsr

Formula: (-379.535003783) * np.sin((-0.0753763320694) * x1 + (-10.9774634559)) + (380.883453656)

ACC: 1

NMSE: 1.54269057981e-05

Token usage: 3824677

Role: pysr

Formula: x1*x1 + 0.49999908 - 0.37485936/(0.8411416 - (Abs(x1**2) - 0.10129087/(x1**4.8704863 + cos(-x1 - x1 + 1.456858))))

ACC: 1

NMSE: 6.91014055428e-14

Token usage: 0

Role: equation author

Formula: (1.15938601543469*x1**4 - 0.419973236789545*x1**2 - 0.0689138786948)/(1.15939590561*x1**2 - 1)

ACC: 1

NMSE: 4.18798317923e-11

Token usage: 167277

Role: candidate decider

Formula: 0.16231978**((x1^2)*((x1^2) - 0.05716716)) + (x1^2) - 1*(-0.49999878) + 0.3751118/((x1^2) - 1*0.99935615 + 0.39309254/(1.3060099 + 0.12215829**(-0.35041705/(x1^2))))

ACC: 1

NMSE: 1.5654330192e-14

Token usage: 61115

Problem: I.50.26

Ground truth: x1*(cos(omega*t)+alpha*cos(omega*t)**2)

Role: search controller

Formula: (omega*cos(omega*(x1/omega)) + 1.0)*cos(omega*(x1/omega))

ACC: 1

NMSE: 3.26108535728e-31

Token usage: 40332

Role: llmsr

Formula: 0.11335305 * x1 * (np.cos(-0.1921522 * omega + 6.652209) + 1.4766192 * np.cos(-0.1921522 * omega + 6.652209) ** 2) + 0.52343757 * omega + -1.3924861

ACC: 0.0513333333333

NMSE: 0.692739899093

Token usage: 2393308

Role: drsr

Formula: (0.277806607845)*x1 + (0.517658722967)*omega + (-1.38092618052)

ACC: 0.0533333333333

NMSE: 0.693343068419

Token usage: 2359583

Role: pysr

Formula: (cos(omega)*Abs(omega) + 1.0)*(cos(omega) - 1.5759296e-8 - 1.5759296e-8)

ACC: 0.0586666666667

NMSE: 1.27082560127

Token usage: 0

Role: equation author

Formula: (0.668477116001) + (0.0394376806463)*(x1^2)*cos(x1)

ACC: 0.08

NMSE: 0.506781595951

Token usage: 600497

Role: candidate decider

Formula: (x1*omega)*0.09522436

ACC: 0.056

NMSE: 0.655605520178

Token usage: 241964

Problem: I.6.2

Ground truth: exp(-(theta/sigma)**2/2)/(sqrt(2*pi)*sigma)

Role: search controller

Formula: 0.349663064670883*log(Abs(0.21498056**(sigma/theta)*(-0.010379398) + (sigma/theta)**(0.21537578 + 1.1628271/(sigma/theta))) + 1)*Abs(theta*(1/(sigma*theta)))

ACC: 1

NMSE: 1.24611774675e-06

Token usage: 236231

Role: llmsr

Formula: 1.0 / np.sqrt(2.0 * np.pi * np.where(sigma == 0, 1e-12, sigma) ** 2) * np.exp(-0.5 * (theta / np.where(sigma == 0, 1e-12, sigma)) ** 2)

ACC: 1

NMSE: 1.19684124305e-30

Token usage: 2535981

Role: drsr

Formula: np.exp(-0.5*(theta/sigma)**2)/(np.sqrt(2*np.pi)*sigma)

ACC: 1

NMSE: 1.503851621e-30

Token usage: 3841066

Role: pysr

Formula: 0.239578709464241*Abs(sin((theta + tanh((sigma*(-0.16107258) + theta)**theta))/(0.799245*(sigma + 0.39649627)))))/Abs(theta)

ACC: 0.957333333333

NMSE: 0.00350923251231

Token usage: 0

Role: equation author

Formula: (0.208658759264) + (-0.0192634109085)*(sigma + theta) + (0.233515077)*(cos(theta) / sigma) + (-0.0725964205601)*cos(theta)

ACC: 0.904

NMSE: 0.0295762168048

Token usage: 792097

Role: candidate decider

Formula: ((theta/sigma)*(-0.0020556143)*Abs((theta/sigma)**(theta/sigma) - 2.618656) + 0.535066)/((((1.3683693*(theta/sigma))**((theta/sigma) - 1*(-0.19024964)) + 0.7439906)*Abs(theta)/Abs((theta/sigma))))

ACC: 1

NMSE: 3.66355918471e-06

Token usage: 228678

Problem: I.6.2a

Ground truth: exp(-theta**2/2)/sqrt(2*pi)

Role: search controller

Formula: 0.60653067**(theta^2)*0.39894226

ACC: 1

NMSE: 1.3527743597e-15

Token usage: 28753

Role: llmsr

Formula: 1 * (1.0 / np.sqrt(2.0 * np.pi * np.where(np.abs(1) < 1e-12, 1e-12, 1) ** 2)) * np.exp(-0.5 * (theta / np.where(np.abs(1) < 1e-12, 1e-12, 1)) ** 2)

ACC: 1

NMSE: 5.47532743918e-31

Token usage: 1933901

Role: drsr

Formula: np.exp(-theta**2/2)/np.sqrt(2*np.pi)

ACC: 1

NMSE: 5.92127054177e-31

Token usage: 3255750

Role: pysr

Formula: -tanh(0.101039540241464*sqrt(1/theta) - (0.05163811*theta + 0.295673031987181*theta/theta**theta)/theta)

ACC: 1

NMSE: 5.39862998812e-06

Token usage: 0

Role: equation author

Formula: (0.367512991053) + (-0.358879467846)*theta**4/((0.946303096681) + theta**2)**2 + (-0.820493073719)*theta**2/((25.6279513856) + theta**4)

ACC: 1

NMSE: 2.54136551914e-08

Token usage: 228479

Role: candidate decider

Formula: 0.39894763/exp((theta^2)*0.5000064 + (1.3877332 + (theta^2)*(-0.6600442))*9.680921e-6)

ACC: 1

NMSE: 4.69554979458e-15

Token usage: 68660

Problem: I.6.2b

Ground truth: exp(-((theta-theta1)/sigma)**2/2)/(sqrt(2*pi)*sigma)

Role: search controller

Formula: (((theta - theta1)/sigma)*exp(-0.5*((theta - theta1)/sigma)**2))*0.69914126/((theta - theta1)/0.57061756)

ACC: 1

NMSE: 1.78474958542e-17

Token usage: 257911

Role: llmsr

Formula: 1.0 / np.sqrt(2.0 * np.pi) * np.exp(-0.5 * ((theta - theta1) / np.where(sigma <= 0, np.finfo(float).eps, sigma)) ** 2) / np.where(sigma <= 0, np.finfo(float).eps, sigma)

ACC: 1

NMSE: 7.23007843233e-31

Token usage: 2552386

Role: drsr

Formula: np.exp(-((theta-theta1)/sigma)**2/2)/(np.sqrt(2*np.pi)*sigma)

ACC: 1

NMSE: 7.79501777449e-31

Token usage: 4636229

Role: pysr

Formula: Abs(0.04209158*sigma + Abs((0.15624918*sigma - 0.44546032)*sin(theta1/(((0.9262945*theta)**log(theta))**log(theta))**0.83251613)))

ACC: 0.846666666667

NMSE: 0.0750868672819

Token usage: 0

Role: equation author

Formula: (0.262320949825) + (-0.114135091954)*(theta - theta1)*(theta - theta1) + (0.0409899964853)*(theta - theta1)*(sigma * (theta - theta1)) + (0.133646123213)*cos(sigma)

ACC: 0.9

NMSE: 0.035681502449

Token usage: 1036822

Role: candidate decider

Formula: (cos((theta - theta1)*(-1.4598972)/sigma)*0.11056169 + cos(((theta - theta1)/sigma)*((theta - theta1)/sigma)*(-2.206705) - 1*(-0.29471588))*(-0.0004061885) + 0.28875753)*cos((theta - theta1)*(-0.6375771)/sigma)/sigma

ACC: 1

NMSE: 6.97587746105e-09

Token usage: 76334

Problem: I.8.14

Ground truth: sqrt((x2-x1)**2+(y2-y1)**2)

Role: search controller

Formula: sqrt((x1 + x2 - y1) + (x1 + x2 - y1)*((x1 + x2 - y1) - 2.5882716) + 2.5615807) - 0.9050158

ACC: 0.224

NMSE: 0.474643751657

Token usage: 220284

Role: llmsr

Formula: 0.29592448 * np.sqrt((-3.9024619e-07 * x2 + 4.3792406 * x1 + -3.3792403 - x1) ** 2 + (-3.3792409 * x2 + 4.3792404 * y1 + 3.4078505e-07 - y1) ** 2) + 4.4316231e-08

ACC: 1

NMSE: 1.87470187776e-14

Token usage: 2409189

Role: drsr

Formula: (0.508789662814)*x1 + (0.161547093229)*x2 + (0.0842874904063)*y1 + (-0.0486600006276)

ACC: 0.138666666667

NMSE: 0.36597103579

Token usage: 2430387

Role: pysr

Formula: (exp(tanh(y1 - 3.89025)) + Abs(y1 - 1*1.0124472) + Abs((y1 - 1.601952)*0.83152944 + cos(y1)))**0.8837622 - 0.44668505

ACC: 0.086

NMSE: 1.1153928699

Token usage: 0

Role: equation author

Formula: (-2.11998257062) + (1.02957659116)*(sqrt(x1^2 + x2^2)) + (2.08724420673)*(y1 / (x1 + x2)) + (-0.989308440334)*log((x1 + x2)*(y1 / (x1 + x2)))

ACC: 0.3

NMSE: 0.295810159139

Token usage: 1014278

Role: candidate decider

Formula: (y1 - (x1 + x2))*(0.22330594*log(Abs(log(y1)) + 1) - 0.20505172) + 0.48109448*Abs(y1 - 0.92410713) + Abs(0.528278392305401*(y1 - (x1 + x2)) + 0.6020189)

ACC: 0.304

NMSE: 0.304861244597

Token usage: 297724

Problem: I.9.18

Ground truth: G*m1*m2/((x2-x1)**2+(y2-y1)**2+(z2-z1)**2)

Role: search controller

Formula: ((G * m1 * m2) + 0.39258987)/tanh((sqrt((x2 - x1)^2 + y1^2))) - 1*0.30494136

ACC: 0.232

NMSE: 0.755552264664

Token usage: 308686

Role: llmsr

Formula: 1.443474 * G * m1 * m2 / np.maximum((1.9018032 * (x2 - x1)) ** 2 + (1.8314187e-05 * y1) ** 2 + 0.35596786, 1e-12) + (0.60676198 * m1 + 0.25685787 * m2 + -1.6987076 * G + 1.2735414 * y1 + 1.0418419)

ACC: 0.833333333333

NMSE: 0.0418580067334

Token usage: 2862857

Role: drsr

Formula: (1.01705453649)*m1 + (1.25689517291)*m2 + (-1.40540552362)*G + (0.87785534435)

ACC: 0.418

NMSE: 0.263875372694

Token usage: 2826814

Role: pysr

Formula: 1.0557821*m1/(G**G*sqrt(G**m2/sqrt(m2 + 0.32477924))*log(x1**m2*x2/(sqrt(y1)*tanh(x1))))

ACC: 0.980666666667

NMSE: 0.013703214401

Token usage: 0

Role: equation author

Formula: (0.167835711029) + (5.69438917005) * (m1 * m2) + (-3.66309434405) * (G * m1 * m2)

ACC: 0.44

NMSE: 0.287537227538

Token usage: 1138455

Role: candidate decider

Formula: (m1 * m2)*((m1 * m2)*((m1 * m2)**9.0085 - ((sqrt((x2 - x1)^2 + y1^2))**4.334156 - ((m1 * m2) + 2.8437583)))/(G * m1 * m2) - 1.5008265) - (((m1 * m2)*(G * m1 * m2))**6.365207 - 1*0.36297327)

ACC: 0.388

NMSE: 0.333559123759

Token usage: 303318

Problem: II.10.9

Ground truth: sigma_den/epsilon*1/(1+chi)

Role: search controller

Formula: (1/sigma_den)/((1/sigma_den) + 1.0)

ACC: 1

NMSE: 4.18695096224e-31

Token usage: 24894

Role: llmsr

Formula: 1 / np.where(1.0 + 1 * sigma_den == 0, np.finfo(float).eps, 1.0 + 1 * sigma_den)

ACC: 1

NMSE: 4.09161992719e-31

Token usage: 2237227

Role: drsr

Formula: (-0.0744791419163)*sigma_den + (0.499458680168)

ACC: 0.694666666667

NMSE: 0.0772641462235

Token usage: 2096748

Role: pysr

Formula: 0.8020135*sin(sin(sin(sin(sigma_den**(-0.6242026724042)))))/sigma_den**0.28605792 - 0.003627948

ACC: 1

NMSE: 1.33422925602e-06

Token usage: 0

Role: equation author

Formula: (-3.30536316635e-17) + (1)*(1/sigma_den)/(1 + (1/sigma_den))

ACC: 1

NMSE: 3.96888059962e-31

Token usage: 161481

Role: candidate decider

Formula: (1/sigma_den)/((1/sigma_den) + 1.0)

ACC: 1

NMSE: 4.18695096224e-31

Token usage: 45751

Problem: II.11.20

Ground truth: n_rho*p_d**2*Ef/(3*kb*T)

Role: search controller

Formula: (1.0 / n_rho)*0.10884799 - (-1)*0.22448534/n_rho

ACC: 1

NMSE: 1.75646513168e-16

Token usage: 22364

Role: llmsr

Formula: 1.4329977e-07 * n_rho + 0.33333323 * n_rho ** -0.99999986

ACC: 1

NMSE: 7.63904844984e-13

Token usage: 1903417

Role: drsr

Formula: (-0.30516085828)*n_rho + (0.0148291731474)*n_rho**2 + (0.897618072521)

ACC: 0.0733333333333

NMSE: 0.6410805432

Token usage: 2836636

Role: pysr

Formula: 0.35261306/(-(-0.057839174)*n_rho + n_rho)

ACC: 1

NMSE: 6.30607875489e-17

Token usage: 0

Role: equation author

Formula: (0.333333333333)*(1.0 / n_rho)**(1)

ACC: 1

NMSE: 1.75529561674e-24

Token usage: 453244

Role: candidate decider

Formula: -0.8783115*(-0.37951607)/n_rho

ACC: 1

NMSE: 3.37055321797e-16

Token usage: 47794

Problem: II.11.27

Ground truth: n*alpha/(1-(n*alpha/3))*epsilon*Ef

Role: search controller

Formula: alpha*n*(0.33319047*alpha*n + 1)

ACC: 1

NMSE: 0.00421945947128

Token usage: 114389

Role: llmsr

Formula: 1 * (n * alpha) / (1.0 - n * alpha / 3.0)

ACC: 1

NMSE: 4.99550892713e-32

Token usage: 2290561

Role: drsr

Formula: (1) * (n * alpha) / (1.0 - (0.333333333333) * n * alpha)

ACC: 1

NMSE: 1.29369611976e-24

Token usage: 4515516

Role: pysr

Formula: Abs((alpha*n*0.9311839/(0.44296482**alpha)**(n/1.6962692))**0.9687117 - 0.001360667)

ACC: 0.953333333333

NMSE: 7.8227369999e-06

Token usage: 0

Role: equation author

Formula: 0.285829885786*alpha**2*n**2 + 1.00779973134*alpha*n - 0.000247116665699

ACC: 1

NMSE: 0.00787533714368

Token usage: 1029505

Role: candidate decider

Formula: 0.13021027*(n*alpha)*((n*alpha)^2) + (n*alpha) + ((n*alpha)^2)*(0.035034955*(n*alpha)*(((n*alpha)^2) + 0.16328543*log(((n*alpha)^2))) + 0.33448017)

ACC: 1

NMSE: 1.35524180065e-11

Token usage: 119218

Problem: II.11.28

Ground truth: 1+n*alpha/(1-(n*alpha/3))

Role: search controller

Formula: exp(alpha*n)

ACC: 1

NMSE: 0.020416525601

Token usage: 295777

Role: llmsr

Formula: 1.0 + n * alpha / (1.0 - n * alpha / 3.0)

ACC: 1

NMSE: 2.04674519593e-31

Token usage: 2513226

Role: drsr

Formula: 1 + n*alpha/(1 - (n*alpha/3))

ACC: 1

NMSE: 2.04674519593e-31

Token usage: 4464641

Role: pysr

Formula: Abs((((0.6222025**alpha)**(0.5308141**alpha) + exp(1.2706684 + n*(-1.6547433))*0.07250163 + 1.6704609)**n)**alpha)

ACC: 1

NMSE: 4.23160849035e-06

Token usage: 0

Role: equation author

Formula: (1) + (1)*(n*alpha)/(1 + (-0.333333333333)*(n*alpha))

ACC: 1

NMSE: 1.60744226791e-25

Token usage: 240992

Role: candidate decider

Formula: 2.1197479**(n*alpha)*(n*alpha)/(n*alpha) + (n*alpha)*(((n*alpha)*0.674423 + 3.6452482)**(n*alpha)*0.039642718 + 0.20902234)

ACC: 1

NMSE: 4.26005958477e-11

Token usage: 174926

Problem: II.11.3

Ground truth: q*Ef/(m*(omega_0**2-omega**2))

Role: search controller

Formula: (q^2)/(1.0 - (q^2)) + 1.0

ACC: 1

NMSE: 3.09100539571e-30

Token usage: 29077

Role: llmsr

Formula: 1 / np.where(1 - 1 * q ** 2 == 0.0, np.finfo(float).eps, 1 - 1 * q ** 2)

ACC: 1

NMSE: 3.70104373288e-30

Token usage: 2264080

Role: drsr

Formula: (1.23508949576)*q + (0.72195947694)

ACC: 0.996666666667

NMSE: 0.0695712824651

Token usage: 2137423

Role: pysr

Formula: (q + (((q*0.8367879 - 1*0.0019004461 + 0.95834386)**(q*2.1072059)*1.7762294)**(q*0.5894595))**(q*1.8297918))**q

ACC: 1

NMSE: 1.4937710353e-06

Token usage: 0

Role: equation author

Formula: (1) + (-0.951010049396)*q**2/((-0.951010049396) + (0.951010049396)*q**2)

ACC: 1

NMSE: 2.91928287373e-30

Token usage: 219005

Role: candidate decider

Formula: (q^2) + (q + (q^2))**(q + 0.39922282)*(q^2)**4.8208995*(-0.23417933) + exp(q*q*q*q)**exp((q^2))

ACC: 1

NMSE: 4.91239159038e-13

Token usage: 124658

Problem: II.11.7

Ground truth: N/A

Role: search controller

Formula: (x1 * cos(x2))*(x3 + 0.20311688*(x1 * cos(x2))) + (x1*sin(x2)) + (0.1303695*(x1 * cos(x2))*(log(0.6628589*(x1*sin(x2))) + sin((x1 * cos(x2)))) + 0.023124570356925)*(sin(cos((x1*sin(x2)))) + 0.6377462)

ACC: 0.972

NMSE: 0.00013488358764

Token usage: 343119

Role: llmsr

Formula: (1.1342027e-07 + -0.42966935 * x1) * (-1.1699957e-07 + 2.3273787 * x3) * (np.sin(0.9999951 * x2 + 4.7123946) + -2.5647214e-06 * np.cos(1.7169189 * x2)) + (0.99999995 * x1 + -1.4445861e-07 * x3)

ACC: 1

NMSE: 2.54089149241e-13

Token usage: 3200505

Role: drsr

Formula: (-1.04233802528)*x1*x2 + (-0.332927166865)*x1*x3 + (-1.6033626133)*x2*x3 + (3.06740790282)*x1 + (2.01466593007)*x2 + (3.21411281515)*x3 + (-4.03598441094)

ACC: 0.472666666667

NMSE: 0.0334236011197

Token usage: 5335548

Role: pysr

Formula: x3*(log(exp(x3)**cos(x3)/(x3*x3*sqrt(x3**exp(x3))))**2.81629165e-9) + 1.0)

ACC: 0.0653333333333

NMSE: 1.72130453649

Token usage: 0

Role: equation author

Formula: (0.0815874580305) + (0.919727387027)*x1 + (3.10308827352)*x3 + (-0.186739819939)*(x1 * x3) + (-1.67857575559)*(x2 * x3)

ACC: 0.444

NMSE: 0.0921527685745

Token usage: 754452

Role: candidate decider

Formula: Abs(0.0060911006**((x3 - x2)*(x3 - x2))*((x1 * x3 / x2)*(x3 - x2)*((x1 * x3 / x2) + 2.3150325) - 1.6446009) + (x1 * x3 / x2) - 1.9149394) - 1*0.37669808

ACC: 0.088

NMSE: 0.869158542856

Token usage: 297982

Problem: II.13.23

Ground truth: rho_c_0/sqrt(1-v**2/c**2)

Role: search controller

Formula: ((rho_c_0 - (1.0 / rho_c_0))/rho_c_0)**(-0.5)

ACC: 1

NMSE: 8.87291389861e-30

Token usage: 31975

Role: llmsr

Formula: 2.8338404 * rho_c_0 / np.sqrt(-8.0309515 + 8.0306352 * rho_c_0 ** 2 + 1e-12)

ACC: 1

NMSE: 7.27302362307e-09

Token usage: 2117955

Role: drsr

Formula: (-0.020968558921)*rho_c_0 + (1.13883924238)

ACC: 0.985333333333

NMSE: 0.423615648958

Token usage: 2273816

Role: pysr

Formula: sqrt(Abs(exp((tanh((rho_c_0 - 0.0098555675)**1.4661278) - 1.8749789)/(rho_c_0*(rho_c_0 - 0.40171865)))**(-1.038369) + 0.0012541421))

ACC: 1

NMSE: 4.78878649504e-06

Token usage: 0

Role: equation author

Formula: ((-303.118056718) + (4.2841737941)*rho_c_0 + (-2.12724860853)*(1.0 / rho_c_0) + (304.089004021)*rho_c_0*(1.0 / rho_c_0))/(1 + (4.28251757901)*rho_c_0 + (-4.4497485269)*(1.0 / rho_c_0))

ACC: 1

NMSE: 4.19363352973e-08

Token usage: 112273

Role: candidate decider

Formula: ((rho_c_0 - (1.0 / rho_c_0))/rho_c_0)**(-0.5)

ACC: 1

NMSE: 8.87291389861e-30

Token usage: 52751

Problem: II.13.34

Ground truth: rho_c_0*v/sqrt(1-v**2/c**2)

Role: search controller

Formula: ((rho_c_0 - (1/rho_c_0))/rho_c_0)**(-0.5)

ACC: 1

NMSE: 8.4042436323e-30

Token usage: 173087

Role: llmsr

Formula: -0.082735998 * rho_c_0 + 0.0062995923 * rho_c_0 ** 2 + 1.2712868

ACC: 0.999333333333

NMSE: 0.154405814558

Token usage: 2545612

Role: drsr

Formula: (-0.0195958370168)*rho_c_0 + (1.13241089528)

ACC: 0.991333333333

NMSE: 0.414325446635

Token usage: 2599445

Role: pysr

Formula: 1.0024616*sqrt(tanh(sqrt(rho_c_0 - sqrt((rho_c_0*tanh(1.001247*rho_c_0**rho_c_0))**(-0.8247827*rho_c_0))))**(-1.5475187))

ACC: 1

NMSE: 2.91057447111e-06

Token usage: 0

Role: equation author

Formula: 0.992524110931 + 0.91144659732/rho_c_0**3

ACC: 1

NMSE: 0.325833336528

Token usage: 174234

Role: candidate decider

Formula: 7.231414e-7 + exp(-(1/rho_c_0)/((1/rho_c_0) + 0.1084044/(exp(rho_c_0)*exp(-((1/rho_c_0) - 0.5450837)/(-2.57727162912072*(log(rho_c_0 - (1/rho_c_0)) - 0.005175262)*exp(-1/(2.36993421820989*(1/rho_c_0) - 0.42195264/(1/rho_c_0))) - 2/((1/rho_c_0) - 0.5450837))) - 2.8671832) - 2/(1/rho_c_0)))

ACC: 1

NMSE: 2.69472541362e-08

Token usage: 73079

Problem: II.15.4

Ground truth: -mom*B*cos(theta)

Role: search controller

Formula: (-(cos(mom)))

ACC: 1

NMSE: 1.56848931077e-31

Token usage: 112914

Role: llmsr

Formula: -mom * (1.2362663 * (1.0 + -0.16428351 * mom)) * np.cos(4.9348984 + -0.55878896 * mom) + -0.87860011

ACC: 0.988666666667

NMSE: 1.31301404193e-06

Token usage: 1668926

Role: drsr

Formula: -(1.3681885469)*mom*np.cos((-0.279826169434)*mom + (-3.09001715086)) + (-1.94946368881)

ACC: 0.654666666667

NMSE: 0.00772920402183

Token usage: 3073954

Role: pysr

Formula: -cos(mom) + tanh(-cos(exp(tanh(Abs(cos((0.71458316 - 1*(-0.9718074))*Abs(mom*Abs(mom))))))) - 0.32430276)*(-7.2541284e-10)*Abs(mom**2)*3.7359478

ACC: 1

NMSE: 6.37151181009e-16

Token usage: 0

Role: equation author

Formula: (2.42796135307e-17) + (-1)*cos(mom)

ACC: 1

NMSE: 1.56335404857e-31

Token usage: 223455

Role: candidate decider

Formula: (cos(mom))*(-1.0)

ACC: 1

NMSE: 1.56848931077e-31

Token usage: 15935

Problem: II.15.5

Ground truth: -p_d*Ef*cos(theta)

Role: search controller

Formula: cos((p_d - 3.141592653589793))

ACC: 1

NMSE: 3.27920407997e-31

Token usage: 35969

Role: llmsr

Formula: -(6.3193652e-06 * p_d + -4.9706884) * -0.20117993 * np.cos(-9.8602352e-08 + 1.0000002 * p_d)

ACC: 1

NMSE: 6.09481364285e-12

Token usage: 1998913

Role: drsr

Formula: (-(-1.60105241781)*np.cos((0.312191846838)))*p_d + (-0.367701366604)*p_d**2 + (0.776474409867)*p_d + (-2.64982001795)

ACC: 0.590666666667

NMSE: 0.0115839851421

Token usage: 3516932

Role: pysr

Formula: sin(p_d - 1.8282704 - 1*(-0.2574741)) + 2.3462569e-8

ACC: 1

NMSE: 5.16747702849e-15

Token usage: 0

Role: equation author

Formula: 7.22446934981e-16 - cos(p_d)

ACC: 1

NMSE: 3.9080777771e-30

Token usage: 140858

Role: candidate decider

Formula: (cos(p_d))*(-1.0)

ACC: 1

NMSE: 2.78509786929e-31

Token usage: 16110

Problem: II.2.42

Ground truth: kappa*(T2-T1)*A/d

Role: search controller

Formula: (kappa - 1.0)/T1

ACC: 1

NMSE: 2.98881470634e-30

Token usage: 129493

Role: llmsr

Formula: (-0.44823751 * kappa + 0.46521929) * (-2.5394371 - -0.59228079 * T1) + 0.013412413

ACC: 0.238

NMSE: 0.0662508915808

Token usage: 2190994

Role: drsr

Formula: (0.61881910405)*kappa + (-0.04193820411)*T1 + (-0.548773052855)

ACC: 0.164

NMSE: 0.207094584563

Token usage: 2266008

Role: pysr

Formula: -(-3.3614231e-9)*(Abs(kappa/(sin(kappa)*(-2.2582293) - 0.81800246)) + 0.72826165 + 0.82826084/((T1*T1*tanh(T1)))) + (kappa - 1*1.0)/T1

ACC: 1

NMSE: 1.55138272905e-15

Token usage: 0

Role: equation author

Formula: (kappa - 1)/T1

ACC: 1

NMSE: 2.98881470634e-30

Token usage: 633457

Role: candidate decider

Formula: (kappa / T1)*(kappa - 1.0)/kappa

ACC: 1

NMSE: 3.01017624691e-30

Token usage: 39659

Problem: II.21.32

Ground truth: q/(4*pi*epsilon*r*(1-v/c))

Role: search controller

Formula: -1*0.0016667007 + 0.08124417 - 0.079577476/(1.0 - q)

ACC: 1

NMSE: 3.77871986412e-15

Token usage: 25457

Role: llmsr

Formula: -0.079578163 * q / (1.0 - 1.0000027 * q + 1e-12) + -1.1938452e-07 * q

ACC: 1

NMSE: 1.18388035122e-10

Token usage: 2090306

Role: drsr

Formula: ((0.249782207644) * q) / ((-3.13885579415) + (3.13885579413) * q)

ACC: 1

NMSE: 1.90433831633e-22

Token usage: 3014636

Role: pysr

Formula: 0.06542736 + 0.024694113/log(sqrt(q) - 0.18625198 + 0.34281272/(q**2 + log(2)))

ACC: 1

NMSE: 1.70920940144e-05

Token usage: 0

Role: equation author

Formula: (-3.12101540374e-16) + (1.96839337451)*log((-0.0396213317436) + q*(1/q))/((-1) + (1/q))

ACC: 1

NMSE: 9.91033721454e-25

Token usage: 113406

Role: candidate decider

Formula: 0.07957747 - (-1)*0.079577476/(q - 1*1.0)

ACC: 1

NMSE: 3.24597079131e-15

Token usage: 33640

Problem: II.24.17

Ground truth: sqrt(omega**2/c**2-pi**2/d**2)

Role: search controller

Formula: sqrt(1.0 - 9.869604*(1/omega)**2)

ACC: 1

NMSE: 6.48031761004e-15

Token usage: 114389

Role: llmsr

Formula: 0.45893149 + -0.0013039628 * omega + 0.70181001 * np.sqrt(np.maximum((omega - 0.91419714) ** 2 / np.where((1.1152685 if 1.1152685 != 0 else 1e-12) * (1.0 + 1.0704315 * (omega - 0.91419714)) == 0, 1e-12, (1.1152685 if 1.1152685 != 0 else 1e-12) * (1.0 + 1.0704315 * (omega - 0.91419714))) ** 2 - np.pi ** 2 / np.where((1.156889 if 1.156889 != 0 else 1e-12) * (1.0 + 1.0853206 * (omega - 0.91419714)) == 0, 1e-12, (1.156889 if 1.156889 != 0 else 1e-12) * (1.0 + 1.0853206 * (omega - 0.91419714))) ** 2, 0.0))

ACC: 1

NMSE: 0.000311066735768

Token usage: 2693099

Role: drsr

Formula: (0.0106266466667)*omega + (0.826605708096)

ACC: 0.986

NMSE: 0.317179293548

Token usage: 2833657

Role: pysr

Formula: tanh(Abs(log(Abs(0.034888964 + log(log(0.450615879252969*omega)/(omega + 0.720247926622294 - 1.222391/omega**2))/omega))))

ACC: 1

NMSE: 1.15753820622e-05

Token usage: 0

Role: equation author

Formula: 0.972489756363965 - 2.8114765124/omega**2

ACC: 0.996

NMSE: 0.232948410403

Token usage: 178162

Role: candidate decider

Formula: 0.8768211*(1/omega)/(log(2*(1/omega)) - 7.9118305e-5) + ((1/omega) - 7.9118305e-5)*(omega - 1.8557299*(1/omega) + 0.20698188) + 0.0024875354

ACC: 1

NMSE: 1.08265287994e-07

Token usage: 6890288

Problem: II.34.11

Ground truth: g_*q*B/(2*m)

Role: search controller

Formula: (g_/2)

ACC: 1

NMSE: 1.83264288161e-31

Token usage: 54983

Role: llmsr

Formula: g_ * 1 * 1 / (2.0 * 1)

ACC: 1

NMSE: 1.9859961275e-31

Token usage: 1582029

Role: drsr

Formula: (0.5) * g_

ACC: 1

NMSE: 8.9340957928e-26

Token usage: 2915901

Role: pysr

Formula: g_*0.5

ACC: 1

NMSE: 1.9859961275e-31

Token usage: 0

Role: equation author

Formula: (g_/2)

ACC: 1

NMSE: 1.83264288161e-31

Token usage: 169092

Role: candidate decider

Formula: (g_/2)

ACC: 1

NMSE: 1.83264288161e-31

Token usage: 14498

Problem: II.34.29b

Ground truth: g_*mom*B*Jz/(h/(2*pi))

Role: search controller

Formula: g_*(h*4.8975224e-8 + h)*0.7581056 + (g_*h)*5.5250797

ACC: 1

NMSE: 6.14649705646e-17

Token usage: 51492

Role: llmsr

Formula: 6.2831854 * g_ * h

ACC: 1

NMSE: 4.1964271312e-16

Token usage: 1887493

Role: drsr

Formula: (7.65956480439)*g_ + (18.9350736092)*h + (-22.9780134531)

ACC: 0.452

NMSE: 0.0968571823586

Token usage: 2097736

Role: pysr

Formula: Abs(h*6.2831855*g_ - 3.82200000000005e-8)

ACC: 1

NMSE: 2.34359139629e-15

Token usage: 0

Role: equation author

Formula: (-1.24857809314e-14) + (6.28318530718)*(g_**2*h - g_*h**2)/(g_ - h)

ACC: 1

NMSE: 1.1615111575e-26

Token usage: 219106

Role: candidate decider

Formula: h*g_*0.626294 + (g_*h)*(-0.00028343804) + (g_*h)*1.5148949e-7 + (g_*h)*2.8119666e-5 + (g_*h)*5.6571465

ACC: 1

NMSE: 4.60968692556e-17

Token usage: 99856

Problem: II.35.18

Ground truth: n_0/(exp(mom*B/(kb*T))+exp(-mom*B/(kb*T)))

Role: search controller

Formula: n_0*(-0.0040454967 - 0.9025022/(-kb*(kb + (kb*(-0.95930755)*kb - 0.51001626)*(kb - 1.2810785)*(-0.12665503)) - 1.8035959))

ACC: 0.996

NMSE: 7.71563627297e-07

Token usage: 200410

Role: llmsr

Formula: 1 * n_0 / (2.0 * np.cosh(1 * kb) + 1e-12)

ACC: 1

NMSE: 6.97144638745e-25

Token usage: 2585319

Role: drsr

Formula: (17.0505872585)*n_0/(np.exp((22.6851957992)/(kb+(-6.71051627716)))+np.exp(-(22.6851957992)/(kb+(-6.71051627716))))

ACC: 0.935333333333

NMSE: 0.00243839146932

Token usage: 2606517

Role: pysr

Formula: sqrt(Abs((n_0**2.0401013/((kb**2.6913178 + 2.4662342)**1.5121171*sqrt(2.276619**kb)))**0.9828742))

ACC: 0.950666666667

NMSE: 0.000229699076238

Token usage: 0

Role: equation author

Formula: (1.07023961643) + (1.02852327862)*tanh((0.503967662359)*n_0 + (-0.812762701547)*kb + (-0.995483857206)*(n_0 / (kb + n_0)))

ACC: 0.9

NMSE: 0.0057228141093

Token usage: 695794

Role: candidate decider

Formula: n_0*exp((n_0 * kb)/n_0)/(exp(kb)*exp((n_0 * kb)/n_0) + n_0/n_0)

ACC: 1

NMSE: 1.40672264334e-31

Token usage: 66053

Problem: II.35.21

Ground truth: n_rho*mom*tanh(mom*B/(kb*T))

Role: search controller

Formula: ((n_rho^2)/(0.20421058 + ((n_rho^2)*0.5014894 + 0.05563015)/n_rho) + 0.8442708)/((n_rho^2) + (n_rho^2) + 0.8336322)

ACC: 1

NMSE: 5.87280809877e-07

Token usage: 197497

Role: llmsr

Formula: (-2.7933362 * n_rho + 1.5588823) * np.tanh(0.0027747606 + 0.28028064 * n_rho / (1.0 + 0.94045256 * n_rho ** 2)) + 0.94342679

ACC: 0.997333333333

NMSE: 0.000143027768706

Token usage: 2545999

Role: drsr

Formula: (-7.89812184221)*n_rho*np.tanh((-0.00308515518114)*n_rho + (0.0404127018655)) + (1.06721153776)

ACC: 0.736666666667

NMSE: 0.124814391012

Token usage: 2340754

Role: pysr

Formula: tanh(n_rho**(-1.0))

ACC: 1

NMSE: 1.17063503446e-31

Token usage: 0

Role: equation author

Formula: (0.946814853492) + (0.440285253404)*n_rho + (0.000865726925344)*(n_rho^2) + (-0.791988466139)*((n_rho^2)/(1 + (n_rho^2))) + (-0.45979039203)*n_rho*((n_rho^2)/(1 + (n_rho^2)))

ACC: 1

NMSE: 5.0014849136e-05

Token usage: 525326

Role: candidate decider

Formula: -0.009366577 + 1.1047633/(n_rho + (exp(-n_rho))/exp(n_rho*1.2697895) - (exp(-n_rho))/(n_rho/0.017294278 - ((exp(-n_rho)) + 0.01729428) + 0.00030370615/((n_rho^2)*(n_rho^2))) + 0.3353261)

ACC: 1

NMSE: 2.35367776328e-06

Token usage: 288641

Problem: II.36.38

Ground truth: mom*H/(kb*T)+(mom*alpha)/(epsilon*c**2*kb*T)*M

Role: search controller

Formula: (mom/(mom / (H * kb)) + 1.0)/mom

ACC: 1

NMSE: 5.36099677297e-31

Token usage: 38582

Role: llmsr

Formula: -0.0048736757 * (mom * H / kb) + -1.9448027 * mom + 0.21907026 * H + 2.4823608 * kb + 0.46062155 * mom ** 2 + 0.078168316 * H ** 2 + 1.4419124 * kb ** 2 + -0.20374351 * mom * H + -1.292378 * mom * kb + 2.6137311

ACC: 0.232

NMSE: 0.25126947868

Token usage: 2708726

Role: drsr

Formula: (-1.0182383937)*mom + (0.288575230881)*H + (2.08384879215)*kb + (1.88976429594)

ACC: 0.150666666667

NMSE: 0.463771113861

Token usage: 2030078

Role: pysr

Formula: ((H*kb)**0.99999994 - 1*(-1.0))/mom

ACC: 1

NMSE: 6.67565755866e-16

Token usage: 0

Role: equation author

Formula: (-0.453076958089) + (1.23530363698)*H*kb + (1.35498413392)/(mom)

ACC: 0.608

NMSE: 0.060337368354

Token usage: 847141

Role: candidate decider

Formula: (mom/(mom / (H * kb)) + 1.0)/mom

ACC: 1

NMSE: 5.36099677297e-31

Token usage: 40801

Problem: II.37.1

Ground truth: mom*(1+chi)*B

Role: search controller

Formula: (mom + 1)

ACC: 1

NMSE: 2.25301410706e-31

Token usage: 77071

Role: llmsr

Formula: 1 * mom + 1

ACC: 1

NMSE: 1.69902168709e-31

Token usage: 1763252

Role: drsr

Formula: (0.999999999997)*mom + (1.00000000001)

ACC: 1

NMSE: 6.55399748286e-24

Token usage: 1977546

Role: pysr

Formula: mom + (mom**0.9103938 + sin(mom/((-0.7802015)*8.57926e-9*(mom - 0.015092924)**0.76457417)))*8.57926e-9 + 0.084805004 + 0.9151949

ACC: 1

NMSE: 4.16535744446e-15

Token usage: 0

Role: equation author

Formula: (1)*mom + (1)

ACC: 1

NMSE: 2.25301410706e-31

Token usage: 150515

Role: candidate decider

Formula: (mom + 1)

ACC: 1

NMSE: 2.25301410706e-31

Token usage: 14167

Problem: II.38.14

Ground truth: Y/(2*(1+sigma))

Role: search controller

Formula: 0.5/(Y + 1.0)

ACC: 1

NMSE: 1.34821910862e-30

Token usage: 26253

Role: llmsr

Formula: -11.922254 * ((2.0663098 * Y + 1.4088918) / (2.0 * (1.0 + (1.5014892 + 3.798454 * (2.0663098 * Y + 1.4088918))))) + 1.5693336

ACC: 1

NMSE: 3.10259231481e-09

Token usage: 1922485

Role: drsr

Formula: ((3.68517212071) * Y) / ((18.9300425285) * (1 + (-6.14193413825))) + (0.250653121785)

ACC: 0.803333333333

NMSE: 0.0772217762739

Token usage: 2056725

Role: pysr

Formula: tanh(tanh(tanh(tanh(tanh(sin(0.349351199894916*(1/(Y - 0.120679565*Abs((Y*(0.25221637*Y - 0.0263304797879862))**0.127195225) + 0.30881733))**0.77290756))) - 0.013649782)))

ACC: 1

NMSE: 4.07585168461e-07

Token usage: 0

Role: equation author

Formula: (-8.80318089194e-17) + (0.5)/(Y + (1))

ACC: 1

NMSE: 1.47214052456e-30

Token usage: 98883

Role: candidate decider

Formula: (1/(Y + 1))*0.5

ACC: 1

NMSE: 1.34821910862e-30

Token usage: 23645

Problem: II.38.3

Ground truth: Y*A*x/d

Role: search controller

Formula: 1.0/(Y / A)

ACC: 1

NMSE: 1.44776839875e-29

Token usage: 28856

Role: llmsr

Formula: 1.0000003 * Y ** -0.99999987 * A ** 0.99999966

ACC: 1

NMSE: 5.24607117665e-14

Token usage: 1895239

Role: drsr

Formula: (0.49264317898) + (-0.242831912186)*Y + (1.04521926663)*A + (-0.217083615655)*Y*A

ACC: 0.216666666667

NMSE: 0.179023557197

Token usage: 2513132

Role: pysr

Formula: A*Abs(tanh(A/(0.10172727 + 0.43854696/((A*(Y*Y - A + cos(8.805438 - A))))) + 163.1792))/Y

ACC: 1

NMSE: 1.17125876008e-29

Token usage: 0

Role: equation author

Formula: A/Y - 9.38573228887e-17

ACC: 1

NMSE: 1.46354582948e-29

Token usage: 913912

Role: candidate decider

Formula: 1.0/(Y / A)

ACC: 1

NMSE: 1.44776839875e-29

Token usage: 34774

Problem: II.6.11

Ground truth: 1/(4*pi*epsilon)*p_d*cos(theta)/r**2

Role: search controller

Formula: epsilon*(epsilon - 0.05003169)*(-epsilon*(epsilon + (epsilon*(1/epsilon) - 0.027247299)*(-0.066837445))*(-(epsilon*(-0.0005778098) - 0.023440387) - 0.027832272) - 0.041208483) + 0.078751706

ACC: 1

NMSE: 5.67019367626e-10

Token usage: 65869

Role: llmsr

Formula: 0.13321451 * (1.0 / epsilon) + 0.15372763 + -0.10816708 * (1.0 / epsilon) ** 2 + -0.16291407 * epsilon + 0.02472239 * epsilon ** 2

ACC: 0.968

NMSE: 0.000270886977025

Token usage: 1790902

Role: drsr

Formula: (-0.065188330282)*epsilon + (0.102547925819)

ACC: 0.281333333333

NMSE: 0.0168954757643

Token usage: 2079140

Role: pysr

Formula: epsilon*(sin(tanh(tanh(epsilon*8.539856)))*cos(epsilon)*0.11531695 + 7.0793665e-10)/(epsilon - 1*1.089012e-9)

ACC: 1

NMSE: 5.21593256589e-15

Token usage: 0

Role: equation author

Formula: (0.430570923596) + (-3.71263975347)*epsilon/((16.0136938384) + (epsilon - 2)**2) + (8.18615910048)*(epsilon - 2)/((47.3480695722) + epsilon**2)

ACC: 1

NMSE: 6.49553193355e-08

Token usage: 141187

Role: candidate decider

Formula: (epsilon - 2)*(-0.006948465*epsilon - (epsilon*(epsilon - 2))*(0.00047556768*(epsilon*(epsilon - 2)) - 0.0021964777*log(epsilon) - 0.0032781037) + 0.006948465*(epsilon*(epsilon - 2)) - 0.0584607713082) - 0.033115566

ACC: 1

NMSE: 1.05928265932e-10

Token usage: 203000

Problem: II.6.15a

Ground truth: p_d/(4*pi*epsilon)*3*z/r**5*sqrt(x**2+y**2)

Role: search controller

Formula: epsilon*r*(0.068664305 + 0.1726109/exp(p_d/((epsilon*0.6986411)))) - (-0.22769476)*(r * p_d)

ACC: 1

NMSE: 9.70721057981e-09

Token usage: 270603

Role: llmsr

Formula: 3.8760532 * (p_d + 1e-12) ** 0.4644257 / (4.0 * np.pi * (np.abs(epsilon) + (abs(0.2706958) + 1e-12)) ** -0.58781928 * (np.abs(r) + (abs(-2.5943337e-09) + 1e-12)) ** -1.004464)

ACC: 0.914666666667

NMSE: 0.00537335289251

Token usage: 2622711

Role: drsr

Formula: (0.203297989505)*epsilon + (0.206745921074)*p_d + (0.445901948409)*r + (-0.484032731062)

ACC: 0.474

NMSE: 0.0434027466223

Token usage: 2453418

Role: pysr

Formula: p_d*r*(sin(tanh((epsilon**2.3513312*0.25633404)**0.6708354*0.26718318/p_d**1.5760373)) + 0.23083165)

ACC: 1

NMSE: 4.88175362122e-06

Token usage: 0

Role: equation author

Formula: (-0.000979741712237) + (0.174546465935)*epsilon*r + (0.171251822986)*p_d*r

ACC: 1

NMSE: 0.00120689470785

Token usage: 770368

Role: candidate decider

Formula: 0.16881745*r*(epsilon + (0.24756354 - 0.02800463*log(p_d/epsilon)**2)*(-epsilon + p_d)*log(p_d/epsilon)) + 0.16881745*(r * p_d)

ACC: 1

NMSE: 4.3180888305e-09

Token usage: 52464

Problem: II.6.15b

Ground truth: p_d/(4*pi*epsilon)*3*cos(theta)*sin(theta)/r**3

Role: search controller

Formula: (epsilon * ((1/epsilon) - (epsilon - 2))) + ((epsilon * ((1/epsilon) - (epsilon - 2))) + 0.16375236)*(-(epsilon * ((1/epsilon) - (epsilon - 2)))*((epsilon * ((1/epsilon) - (epsilon - 2)))*((epsilon * ((1/epsilon) - (epsilon - 2))) - 1*0.5600756)*(-0.00020434124) + ((epsilon - 2) - 1.656179)*0.02198359) - 1.01416) - 1*0.069794156 + 0.117264226

ACC: 1

NMSE: 5.38110808193e-07

Token usage: 204931

Role: llmsr

Formula: 2.5519377 / epsilon + -70.000235 / (epsilon + 2.3220606) + 57.846789 / (epsilon ** 2 + 4.4226143) + 7.9614827

ACC: 0.993333333333

NMSE: 5.60651501598e-05

Token usage: 1865531

Role: drsr

Formula: ((-5.05947273316)/epsilon + (8.69106884539)/(epsilon**2) + (-4.37579758766)/(epsilon**3) + (0.813052294933))

ACC: 0.302666666667

NMSE: 0.0352132017498

Token usage: 3738102

Role: pysr

Formula: 1.3660393e-9 + epsilon*(cos(epsilon)*0.23873241 - 9.0095836e-10/epsilon)*sin(epsilon)/epsilon

ACC: 1

NMSE: 1.04935946624e-15

Token usage: 0

Role: equation author

Formula: (-0.11223983946) + (0.518357400068)*epsilon + (-0.187700387601)*epsilon*epsilon + (0.131747553457)*(epsilon^2)*(epsilon^2) + (-0.22375956084)*epsilon*(epsilon^2) + (-0.0181792496546)*epsilon*(epsilon^2)*(epsilon^2)

ACC: 1

NMSE: 4.2585302675e-06

Token usage: 945654

Role: candidate decider

Formula: (epsilon - 2)*((-(epsilon^2) - 0.85825807)*(epsilon - 2)*(-0.25161344) - (epsilon^2))*(-0.036121774*(epsilon - 2)*(1.9518888 - 0.6759537*(epsilon - 2)) - 0.0016113544*(epsilon - 2) + 0.038986344) - 0.09031534

ACC: 1

NMSE: 1.53414461327e-07

Token usage: 207448

Problem: III.10.19

Ground truth: mom*sqrt(Bx**2+By**2+Bz**2)

Role: search controller

Formula: (sqrt(mom^2 + Bx^2 + 1))

ACC: 1

NMSE: 1.62180451309e-31

Token usage: 28798

Role: llmsr

Formula: -0.016284986 * mom * np.abs(Bx) + 0.33558711 * mom + 0.34642896 * np.abs(Bx) + 0.83008722 + 0.13492292 * mom ** 2 + 0.16591828 * np.abs(Bx) ** 2 + 0.026285501 * mom ** 2 * np.abs(Bx) + -0.072237053 * mom * np.abs(Bx) ** 2 + 0.014786006 * mom ** 2 * np.abs(Bx) ** 2 + -0.012723183 * mom ** 3 * np.abs(Bx)

ACC: 1

NMSE: 0.000720571172939

Token usage: 2023966

Role: drsr

Formula: (0.642935141318)*mom + (0.629481375058)*Bx + (0.545179420615)

ACC: 0.976666666667

NMSE: 0.0103037885986

Token usage: 2288281

Role: pysr

Formula: sqrt(mom**2.0 + Bx**2 + Abs(Bx/Abs(Bx)))

ACC: 1

NMSE: 1.47439400817e-31

Token usage: 0

Role: equation author

Formula: 0.0184825002219*Bx*mom**2 - 0.276968576887202*Bx*mom + 1.19124815776*Bx + 1.19124815776*mom - 0.222834171547*log(Bx*mom*(Bx + mom)) - 0.228874700005

ACC: 0.996

NMSE: 0.00934231035038

Token usage: 156504

Role: candidate decider

Formula: sqrt(mom*(mom - Bx) + Bx*(mom + Bx) + 1.0)

ACC: 1

NMSE: 1.68035341248e-31

Token usage: 41356

Problem: III.12.43

Ground truth: n*(h/(2*pi))

Role: search controller

Formula: n*0.11889101 + 0.040263932/(1.0 / n)

ACC: 1

NMSE: 3.7282789198e-16

Token usage: 52196

Role: llmsr

Formula: 1 * n / (2.0 * np.pi)

ACC: 1

NMSE: 1.21448528707e-31

Token usage: 1563015

Role: drsr

Formula: (0.159154943092) * n

ACC: 1

NMSE: 1.20742521141e-24

Token usage: 2862514

Role: pysr

Formula:
0.159154940581905*Abs(Abs(Abs(Abs(Abs(Abs(Abs(Abs(Abs(Abs(Abs(Abs(Abs(Abs(Abs(n)))))))))))))))

ACC: 1

NMSE: 1.85804052988e-15

Token usage: 0

Role: equation author

Formula: 0.159154943092*n

ACC: 1

NMSE: 3.42641393954e-24

Token usage: 137668

Role: candidate decider

Formula: n*0.028447693 + n*0.13070725

ACC: 1

NMSE: 2.64078690479e-18

Token usage: 27544

Problem: III.13.18

Ground truth: 2*E_n*d**2*k/(h/(2*pi))

Role: search controller

Formula: E_n*0.00012561145 + E_n + E_n + E_n*10.566245 - 1.4653083e-6

ACC: 1

NMSE: 5.27357684609e-16

Token usage: 114389

Role: llmsr

Formula: 4 * np.pi * 1 ** 2 * 1 / 1 * E_n

ACC: 1

NMSE: 1.52626860793e-31

Token usage: 1758706

Role: drsr

Formula: (12.5663706144)*E_n

ACC: 1

NMSE: 1.12051855163e-28

Token usage: 3243617

Role: pysr

Formula: -E_n + E_n*1.0254036 + E_n*12.540967

ACC: 1

NMSE: 5.32599790318e-18

Token usage: 0

Role: equation author

Formula: (2.05309031715e-14) + (6.28318530718)*(E_n + (E_n^2)) + (6.28318530718)*(E_n - (E_n^2))

ACC: 1

NMSE: 1.90989258647e-26

Token usage: 541144

Role: candidate decider

Formula: (E_n - 9.2482395 + (E_n^2)/E_n)*6.2831855 + 58.108402

ACC: 1

NMSE: 1.37325657347e-15

Token usage: 27148

Problem: III.14.14

Ground truth: I_0*(exp(q*Volt/(kb*T))-1)

Role: search controller

Formula: exp((1.0 / I_0)) - 1.0

ACC: 1

NMSE: 2.984442244835e-31

Token usage: 47103

Role: llmsr

Formula: I_0 * np.expm1(-8.4808132 * I_0 + 6.466813) + 2.5616541

ACC: 0.804666666667

NMSE: 0.0292848226061

Token usage: 2088485

Role: drsr

Formula: (-2.88427266938)*I_0 + (5.3516156076)

ACC: 0.123333333333

NMSE: 0.550040417499

Token usage: 2128012

Role: pysr

Formula: exp(I_0/(I_0*I_0)) - 1.7585578 - 1*(-0.7585578)

ACC: 1

NMSE: 4.68421985037e-31

Token usage: 0

Role: equation author

Formula: (I_0*(I_0 + 0.494524519177) + 0.112605801901)/(I_0*(I_0**2 - 0.064426717090446))

ACC: 0.992

NMSE: 0.00801436092806

Token usage: 161729

Role: candidate decider

Formula: exp((1.0 / I_0)) - 1.2649703 - 1*(-0.2649703)

ACC: 1

NMSE: 3.04052057751e-31

Token usage: 51020

Problem: III.15.12

Ground truth: 2*U*(1-cos(k*d))

Role: search controller

Formula: 2.0 - ((cos(U)) + (cos(U)))

ACC: 1

NMSE: 3.70252248654e-31

Token usage: 26156

Role: llmsr

Formula: (-3.8274329e-08 * U + 2.0000001) * (1.0 - np.cos(-1.0 * U + 3.3690978e-07)) + -1.1396977e-07

ACC: 1

NMSE: 1.55002210288e-13

Token usage: 2096072

Role: drsr

Formula: ((-0.333237745717) * U * (1 - np.cos((0.567434073996)))) + (2.62791925286)

ACC: 0.088

NMSE: 0.999327784374

Token usage: 2945235

Role: pysr

Formula: (1.9459958*Abs(Abs(U)/Abs(Abs(U))) + 0.054004263) + cos(Abs(U))*(-1.9999999)

ACC: 1

NMSE: 4.18950072854e-15

Token usage: 0

Role: equation author

Formula: 2 - 2*cos(U)

ACC: 1

NMSE: 3.70252248654e-31

Token usage: 222705

Role: candidate decider

Formula: 2.0 - ((cos(U)) + (cos(U)))

ACC: 1

NMSE: 3.70252248654e-31

Token usage: 27457

Problem: III.15.27

Ground truth: 2*pi*alpha/(n*d)

Role: search controller

Formula: (alpha / n)*1.5708168 + (alpha / n)*4.7123685

ACC: 1

NMSE: 4.44712620233e-18

Token usage: 89014

Role: llmsr

Formula: 1 * 2 * np.pi * alpha / n

ACC: 1

NMSE: 9.66282813769e-32

Token usage: 1755505

Role: drsr

Formula: (6.28318530718) * alpha / n

ACC: 1

NMSE: 2.5861231436e-30

Token usage: 3881386

Role: pysr

Formula: alpha*6.2831855/n

ACC: 1

NMSE: 3.2429862356e-15

Token usage: 0

Role: equation author

Formula: 6.28318530718*alpha/n

ACC: 1

NMSE: 1.47854212676e-26

Token usage: 178064

Role: candidate decider

Formula: (alpha/n) + (alpha/n)*4.9167905 + ((alpha/n)*(-1.6379939) + (alpha/n))*(-0.57429206)

ACC: 1

NMSE: 4.93584124722e-17

Token usage: 6890288

Problem: III.17.37

Ground truth: beta*(1+alpha*cos(theta))

Role: search controller

Formula: beta + (alpha*beta)*cos(theta)

ACC: 1

NMSE: 1.40499482024e-31

Token usage: 44327

Role: llmsr

Formula: 1 * beta * (1.0 + alpha * np.cos(theta))

ACC: 1

NMSE: 1.19891596399e-31

Token usage: 1711960

Role: drsr

Formula: beta * (1 + alpha * np.cos(theta))

ACC: 1

NMSE: 1.19891596399e-31

Token usage: 4838578

Role: pysr

Formula: beta*alpha*(-1.6430067e-9*beta**log(beta)*theta*log((theta*(theta - 0.92251469))**log(log(beta))) + cos(theta)) + beta

ACC: 1

NMSE: 6.50919159571e-15

Token usage: 0

Role: equation author

Formula: (2.07665555712) + (0.448672002161)*(alpha * cos(theta)) + (0.00759824602557)*(alpha * beta)*(alpha * beta) + (0.548880049974)*(alpha * beta)*cos(theta) + (0.0818523560697)*(alpha * cos(theta))*(alpha * beta)

ACC: 0.44

NMSE: 0.0205139599915

Token usage: 667064

Role: candidate decider

Formula: beta + alpha*(beta*cos(theta))

ACC: 1

NMSE: 1.33337107633e-31

Token usage: 6890288

Problem: III.19.51

Ground truth: -m*q**4/(2*(4*pi*epsilon)**2*(h/(2*pi))**2)*(1/n**2)

Role: search controller

Formula: N/A

ACC: -1

NMSE: N/A

Token usage: 95012

Role: llmsr

Formula: -0.0061082866 * m ** -2.0003658 / np.where(0.22105621 == 0.0, 1.0, 0.22105621) ** 2 + -5.267681e-06

ACC: 1

NMSE: 4.63339260601e-09

Token usage: 1911648

Role: drsr

Formula: (0.019147858082)*m + (-0.0832552187081)

ACC: 0.0773333333333

NMSE: 0.306978069973

Token usage: 2164536

Role: pysr

Formula: (0.2593472/Abs(m**2))/(-2.0747776)

ACC: 1

NMSE: 4.78616232518e-30

Token usage: 0

Role: equation author

Formula: -0.125/m**2

ACC: 1

NMSE: 4.46966647675e-30

Token usage: 712870

Role: candidate decider

Formula: (1.0 / (m^2))*(-0.125)

ACC: 1

NMSE: 4.46966647675e-30

Token usage: 6890288

Problem: III.4.32

Ground truth: 1/(exp((h/(2*pi))*omega/(kb*T))-1)

Role: search controller

Formula: 6.2831855*h - 0.4999999 + 0.0132425021783717/h

ACC: 0.996

NMSE: 1.15835499236e-08

Token usage: 114389

Role: llmsr

Formula: 1.0 / (np.exp((-1.5276451e-08 * h + 0.43410997) / (2.7275926 * h + 4.1508781e-07 + 1e-12)) - 1.0 + 1e-12)

ACC: 1

NMSE: 2.51900567569e-14

Token usage: 2100791

Role: drsr

Formula: (6.27582293318)*h + (-0.469836893771)

ACC: 0.983333333333

NMSE: 2.72939763842e-06

Token usage: 2144855

Role: pysr

Formula: Abs(-(h - (-5.2831855)*h - 0.500003) + tanh(tanh(tanh(tanh(sin(tanh(tanh(tanh(-0.013270048/h)))))))))

ACC: 1

NMSE: 5.12149738845e-12

Token usage: 0

Role: equation author

Formula: (6.28320047298)*(h**2 + (-0.0752811917256)*h + (0.0017920913526))/(h + (0.00431349656036))

ACC: 1

NMSE: 1.46988359128e-09

Token usage: 126657

Role: candidate decider

Formula: ((h^2)/h)*6.2831855 - 1*0.5000009 + (-1*(-0.013263675) + 4.4928882e-8/(h^3))/((h^2)/h) - 5.76498e-6/(h^3)

ACC: 1

NMSE: 5.69081020626e-12

Token usage: 6890288

Problem: III.4.33

Ground truth: (h/(2*pi))*omega/(exp((h/(2*pi))*omega/(kb*T))-1)

Role: search controller

Formula: h - 0.0604476900785013

ACC: 0.996

NMSE: 0.000132552870313

Token usage: 114389

Role: llmsr

Formula: 0.0014019339 * h / (np.exp(1.4979437 * h) - 1) + 0.99994537 * h + -0.078866942 + 0.0021598181 / (np.exp(1.10063 * h) - 1)

ACC: 1

NMSE: 1.90740816598e-09

Token usage: 3286611

Role: drsr

Formula: ((1.05708704765)*h)/(np.exp(((1.08843303815)*h+(0.0310004792342))/((1.50909393246)*h+(-0.0430262911969)))-1)

ACC: 0.999333333333

NMSE: 8.84247970418e-09

Token usage: 2604446

Role: pysr

Formula: Abs(Abs(Abs(h) - 0.060125876) + (h - 0.02576486)*(Abs(h - 0.018276904) - 0.08591131)*(-0.0194713)/(Abs(h - 0.019902864)*h))

ACC: 1

NMSE: 4.54958034647e-11

Token usage: 0

Role: equation author

Formula: (1.00001017873)*(h + (-0.0388956804704))**2/(h + (0.0077542758959)) + (0.00590792811082)

ACC: 1

NMSE: 1.44756177698e-10

Token usage: 590609

Role: candidate decider

Formula: (-h + (h) + ((h) - (h + 1.3035443e-6))*((h)**(-1.8216) + (h) + 0.0021109716) + 0.0021108019)/h**1.0006716 + (h) - 0.07957566

ACC: 1

NMSE: 2.06222702592e-13

Token usage: 6890288

Problem: III.8.54

Ground truth: sin(E_n*t/(h/(2*pi)))**2

Role: search controller

Formula: (E_n^2)*(E_n^2)*0.0069326395 + (0.28619504*(E_n*(E_n^2)*(E_n^2) + (exp(-E_n)))*exp(-0.28619504*E_n*(E_n^2)**2 - 0.28619504*(exp(-E_n))))*3.091553 - 27.967654484871*((E_n^2)*3.8721023 - 2.9844203)*exp(-3.8721023*(E_n^2))*exp(27.967654484871*((E_n^2)*3.8721023 - 2.9844203)*exp(-3.8721023*(E_n^2)))

ACC: 0.58

NMSE: 0.20869128382

Token usage: 114389

Role: llmsr

Formula: 0.35639401 + 0.48716266 * np.sin(2.0 * np.pi * 0.86062854 * E_n / np.where(1.2363879 == 0, 1e-12, 1.2363879) + 1.7376258) ** 2 + -0.090127909 * E_n

ACC: 0.064

NMSE: 0.786891508824

Token usage: 2445134

Role: drsr

Formula: np.sin((0.105791510407) * E_n + (-0.908601984159))**2

ACC: 0.0566666666667

NMSE: 0.977604662102

Token usage: 2858862

Role: pysr

Formula: Abs(sin(-6.283186/E_n))**2.000011

ACC: 1

NMSE: 1.38503194071e-11

Token usage: 0

Role: equation author

Formula: (0.606830999158) + (E_n^2)*cos(E_n)

ACC: 0.268

NMSE: 15.8430742072

Token usage: 781103

Role: candidate decider

Formula: sin(-1.5707963 - 12.566371/((E_n^2)/(-E_n + (E_n^2))) + (-6.041159 + (exp(-E_n))/(E_n^2))*(E_n - (E_n^2)/E_n)/(E_n^2))*0.5 + 0.5

ACC: 1

NMSE: 3.69829727814e-14

Token usage: 6890288

Problem: III.9.52

Ground truth: (p_d*Ef*t/(h/(2*pi)))*sin((omega-omega_0)*t/2)**2/((omega-omega_0)*t/2)**2

Role: search controller

Formula: (1.2337629 - sin((-Ef + t + 5.142917)/(-3.2749314)))*1.4057425*(1.0008206 - sin((-Ef + t + 3.144172)/(-2.0014355)))/p_d

ACC: 0.972

NMSE: 3.17520513824e-05

Token usage: 114389

Role: llmsr

Formula: 5.1377461 * p_d ** -0.78618795 * Ef ** -0.1725622 * t ** 0.28574052 * (np.sin(0.26758702 * t / 2.0) ** 2 / (0.26758702 * t / 2.0 + -0.02240812) ** 2)

ACC: 0.13

NMSE: 0.687543004695

Token usage: 3295109

Role: drsr

Formula: (-32.5508339784)*p_d + (-0.357124004602)*Ef + (-0.330409877734)*t + (30.2955046113)

ACC: 0.140666666667

NMSE: 0.846541844867

Token usage: 2761334

Role: pysr

Formula: Abs(Abs(Abs(t - Abs(13.468766 + 6.12928*sin(log(1.482173*Ef*(Ef/t - 0.74737805) + 2.9269912))/p_d)) - 5.203321) - 3.7916138)

ACC: 0.374

NMSE: 0.0854526648114

Token usage: 0

Role: equation author

Formula: (32.6101731828) + (-16.3252446917)*p_d*exp(-(-0.106583558307)*t) + (-5.70458859044)*Ef*exp(-(0.306999401491)*t)

ACC: 0.236

NMSE: 0.72058000746

Token usage: 765444

Role: candidate decider

Formula: -1*0.0012978212 + (-(-0.0002298771)*((Ef - t)^2)/exp(p_d) + (6.6732526 - 0.17271689*((Ef - t)^2))*(exp((-(-0.006307087)*((Ef - t)^2)*((Ef - t)^2) + ((Ef - t)^2))*(-0.053582754)) - 0.05865095))/p_d

ACC: 0.94

NMSE: 7.79571330274e-05

Token usage: 6890288

Dataset group: complex_formula

--------------------------------------------------------------------------------

Problem: CRK0

Ground truth: -0.1899*A^2 + (0.459688*A^2)/(0.7498*A^4 + 1)

Role: search controller

Formula: -0.623587514510912*A**2 + 0.941978882215093*A - 0.246802528140829

ACC: 0.997428571429

NMSE: 0.00030446440905

Role: llmsr

Formula: 0.79183693 + 3.8209172e-08 * t - (1.1126865 + (1.095212 - 1.1126865) * (1 - np.exp(-t / np.clip(1.0183726, 1e-12, None))) * A ** np.clip(0.99736705, 0.0, None)) - 1.1025406 * (A - 0.7614884) + 0.65655283 * A ** np.clip(0.86891292, 0.0, None)

ACC: 0.984

NMSE: 3.96969083688433E-07

Role: drsr

Formula: -(-2.50011634821)*A - (2.05806017639)*A**2 - (-0.437215087889)*A**3 + (-2.35047465035e-05)*np.exp(-(0.3521279243)*t) + (-0.80641424975)

ACC: 0.999

NMSE: 9.20524136990111E-10

Role: pysr

Formula: Abs(A*exp(-t*A/(2.8321362/A) + exp(A*A*(-sin(t) - 4.6407003)) - 2.6322055) - 1*1.4102863e-5)

ACC: 0.222

NMSE: 0.000206037101513

Role: equation author

Formula: (-0.813479460233) + (2.51956385973)*A + (-2.07603042632)*A**2 + (0.442778275143)*A**3

ACC: 1

NMSE: 1.48453768138e-09

Role: candidate decider

Formula: ((A - 1.173) - ((A - 1.173)^2)*((A - 1.173) + 0.023408754) + ((A - 1.173)^2) - (-((A - 1.173)^2) + exp(((A - 1.173)^2))))*(-0.52292943) - 1*0.52280223

ACC: 1

NMSE: 1.46566576007e-10

Problem: I.37.4_0_1

Ground truth: 2*I2*cos(delta)^2 + I2 + Int + 2*sqrt(I2*(I2*cos(delta)^2 + I2 + Int))*cos(delta)

Role: search controller

Formula: I2 + Int + (cos(delta))*(I2*((cos(delta))*((cos(delta)) + 3.3952599) + 3.542877) + Int - (Int*(0.43900788 + (cos(delta))*(-0.07017853)) - (sqrt(Int*I2))))*0.5667753

ACC: 0.987509433962

NMSE: 0.000274543224467

Role: llmsr

Formula: (np.sqrt(np.clip(I2, 0.0, None)) * np.cos(delta) + np.sqrt(np.clip(0.99999999 * (np.clip(Int, 0.0, None) - np.clip(I2, 0.0, None) * (1.0 - np.cos(delta) ** 2)) + 2 * np.clip(I2, 0.0, None), 0.0, None))) ** 2

ACC: 1

NMSE: 2.37337863874433E-14

Role: drsr

Formula: (Int + (1.39168679436)*I2*np.cos(delta) - (-1.15647141611)*I2) / ((1.05612422142) + (-0.664309580321)*np.cos(delta) + 1e-8)

ACC: 0.9054375

NMSE: 0.0037220484666894

Role: pysr

Formula: Abs((Int + 0.623474458136165*exp(sqrt(I2) + cos(delta)) - 0.022444502)*exp(cos(delta))**0.6476931 - 0.63017774)

ACC: 0.83135

NMSE: 0.00548658357172

Role: equation author

Formula: (-0.037035976986) + (1.01068270465)*(Int + I2) + (1.24595133436)*(2*sqrt(Int*I2))*(cos(delta)) + (0.535839646511)*(2*sqrt(Int*I2))*(cos(delta))*(cos(delta))

ACC: 0.878

NMSE: 0.0120336922502

Role: candidate decider

Formula: (I2*(cos(delta))*(0.27563432 - 0.17236868*(cos(delta))**2) + (Int + I2))*sqrt((I2/(0.0559349498906751*Int + 0.0337412091234378*I2) + 1.5054215)**(cos(delta))) + 0.003649249

ACC: 1

NMSE: 3.84704558979e-06

Problem: III.4.33_3_0

Ground truth: (h*omega)/(2*pi*kb*log(1 + (h*omega)/(2*pi*E_n)))

Role: search controller

Formula: (7.7758526494585*E_n**2 + 1.27258722575714*E_n*h*omega + 0.0379624692664019*h**2*omega**2)/(kb*(7.7758526494585*E_n + 0.7545174*h*omega))

ACC: 1

NMSE: 0.000196735361542

Role: llmsr

Formula: h / (2.0 * np.pi) * omega / kb / np.maximum(np.log1p(h / (2.0 * np.pi) * omega / (np.maximum(E_n - (0.5 + -0.5) * (h / (2.0 * np.pi)) * (omega * (1.0 + 0.58094473)), 1e-12) + 1e-12) + 1e-12), 1e-12)

ACC: 1

NMSE: 5.47404150472047E-15

Role: drsr

Formula: ((4.05397374512)*E_n + (0.261618170739)*h*omega + (0.0156415308531)*E_n*h*omega + (0.196186895535)) / (kb*((4.08044670215) + (0.00700417604028)*h*omega + (-0.000711519323664)*E_n + (0.000246064465768)*h*omega*E_n))

ACC: 0.997125

NMSE: 3.88030566420042E-06

Role: pysr

Formula: (E_n + log(log(log(1.9311807**h + sin(sqrt(1.9311807**(h + 0.44728914) - 0.302587354901568)) + 0.44728914)**omega + 4.213333) - 0.5485824))/kb

ACC: 0.9879

NMSE: 2.76932477566e-05

Role: equation author

Formula: (1.00063503216*E_n + 0.0715201182899*h*omega)/kb

ACC: 0.99

NMSE: 2.91191450899e-05

Role: candidate decider

Formula: (h * omega / kb)*((E_n / (h * omega)) + (-0.0013812685 - 4.828934e-6/(E_n / (h * omega)))/((E_n / (h * omega)) + 0.107747674) - 1*(-0.07957804) - 0.0007284028/((E_n / (h * omega)) + 0.030991424) - 4.7824376e-8/(0.057898674 - 0.0013812516/(E_n / (h * omega))))

ACC: 1

NMSE: 3.21664624125e-14

Problem: bactgrow

Ground truth: mu_max*b*(s/(Ks+s))*(tanh(k*(temp-x0))/(1 + c*(temp-x_decay)^4))*exp(-abs(pH-pH_opt))*sin(((pH-pH_min)*pi/(pH_max-pH_min))^2)

Role: search controller

Formula: (b + (b * s)*0.10697548)*log(log(1 + exp(-Abs((5.603636 + temp*(-0.16584201))*(17.702116 + temp*(-0.48881134))))) + 1)*log(log(1 + exp(-Abs(1.53724996354776*pH - 10.7648392909123))) + 1)

ACC: 0.138142857143

NMSE: 0.00432752351084

Role: llmsr

Formula: (0.046499216 + 0.33539561 * s / (np.abs(0.84932334) + 1e-12 + s) * (1 - s / (np.abs(-1.0798344) + 1e-12))) * (1 / (1 + np.exp(-(np.abs(1.2447839) + 1e-12) * (temp - 1.0056442))) - 1 / (1 + np.exp(-(np.abs(1.2447839) + 1e-12) * (0.73312607 - 1.0056442))) / (1 - 1 / (1 + np.exp(-(np.abs(1.2447839) + 1e-12) * (0.73312607 - 1.0056442))) - 1 / (1 + np.exp(-(np.abs(1.2447839) + 1e-12) * (1.2658531 - 1.0056442))))) * np.exp(-(pH - -0.36901177) ** 2 / 2) * (((np.abs(-0.047172079) + 1e-12) / (np.abs(-0.047172079) + 1e-12 + b)) ** (np.abs(1.8654913) + 1e-12) * (1 - (np.abs(1.2738282) / (np.abs(1.2738282) + b)) ** (np.abs(0.36464351) + 1e-12)))

ACC: 0.00133333333333333

NMSE: 1.077955185286

Role: drsr

Formula: (0.0165524935353) * b * (s / ((1.20219110741) + s)) * np.exp(-(-0.00125549919395) * (temp - (-10.7585972103))**2) * np.exp(-(0.774915960402) * (pH - (7.03320309842))**2) - (-5.08179863601e-05) * b - (0.000375217919815) * b**2

ACC: 0.0101333333333333

NMSE: 0.312387558034066

Role: pysr

Formula: (sqrt(temp) - 3.9351313)*Abs(0.03407653*Abs((sin(pH + 0.84856886) + 0.5643238)*(-sqrt(temp) + sin(b) + 1.0768099)) - 0.0318680754124849)

ACC: 0.018

NMSE: 0.405795378728

Role: equation author

Formula: (0.00906311441171) * b * exp((0.0625724138888) * temp) * exp((0.0284401623164) * (temp * (1 - abs(pH - 7))))

ACC: 0.048

NMSE: 0.27627138061

Role: candidate decider

Formula: (-0.06592953 + 1.2631224*exp(-Abs(0.214705877299155*temp - 8.000169)))*(b*exp(-1.1210316*Abs(pH - 7.0004935)) - 0.033562835)*exp(-1.5256155*exp(-Abs(s)))*exp(-Abs(0.233876425329309*temp - 7.656467))

ACC: 0.358

NMSE: 0.00167709666715

Problem: oscillator1

Ground truth: 0.8*sin(x) - 0.5*v^3 - 0.2*x^3 - 0.5*x*v - x*cos(x)

Role: search controller

Formula: -0.499016539205744*v*(v**2 + x) + 0.151556785639324*x**3 - 0.19849712*x

ACC: 0.998210526316

NMSE: 2.53857070382e-06

Role: llmsr

Formula: (0.36725823 * x - 0.11316885 * x ** 3 + (0.00019286678 * v - 0.94674756 * v ** 3) + -0.94265558 * x * v + -1.055715e-05 * (1.0 + 0.1 * np.abs(x) + 0.05 * np.abs(v)) * np.cos(x + 0.90867438) + 0.37186031 * (1.0 + 0.1 * x ** 2) * np.cos(2.0 * x + 1.5707295)) / 1.8844166

ACC: 0.9992

NMSE: 1.3302684870906E-07

Role: drsr

Formula: -(-0.894452906022)*x - (-3.9056718967e-08)*v - (0.191816100413)*x**3 - (0.500001417226)*v**3 - (0.499999881871)*x*v + (1.65519555271e-08) + (0.780663043147)*np.sin((-1.40195509155)*x)

ACC: 1

NMSE: 1.67374488970997E-12

Role: pysr

Formula: (-v + cos(x*1.1915406)*(-1.3737239))*(x*tanh(tanh(v) - 1*(-0.42587212)) + v*v*(v - 0.009031496))*0.34999835

ACC: 0.9744

NMSE: 0.000194098552143

Role: equation author

Formula: -0.49711038403*v**3 - 0.500211023945*v*x + 0.153061890717*x**3 - 0.198893072711*x

ACC: 0.998

NMSE: 1.94924320828e-06

Role: candidate decider

Formula: -0.048358284*x + (x - (x^3))*(-v*(v^2)*(v*((x*v) - 1*(-0.8564168)) - 0.13683422) - 0.14960459) + (v*(v^2) + (x*v))*(-0.50056773)

ACC: 1

NMSE: 4.35926373117e-07

Problem: oscillator2

Ground truth: 0.3*sin(t) - 0.5*v^3 - 1.0*x*v - 5.0*x*exp(0.5*x)

Role: search controller

Formula: -0.99975324*v*(x - cos(v)) - v - 5.000083*x + 0.30000252995933*sin(t) - 2.5000415*tan(x**2)

ACC: 0.999052631579

NMSE: 9.80782121794e-06

Role: llmsr

Formula: 0.94692288 * (-0.31684273 * np.sin(0.99998977 * t + 9.4252672) - (5.2788896 * x + 2.6424689 * x ** 2 + 0.73923909 * x ** 3) - (-0.0004869684 * v + 0.53634821 * v ** 3) - 1.0556832 * x * v)

ACC: 0.9992

NMSE: 4.41383910893228E-08

Role: drsr

Formula: (-5.00005584482)*x + (-0.622171533691)*x**3 + (9.16119541255e-08)*v + (-0.500052307074)*v**3 + (-0.99997079799)*x*v + (-2.50284285053)*x**2 + (3.22054641656e-05)*v**2 + (-0.300000225672)*np.sin((-1.00000017885)*t) + (6.39506807991e-06)

ACC: 1

NMSE: 4.22206564715512E-10

Role: pysr

Formula: (sin(x) + 0.55058265*cos(2.23104495412801*t - 0.426400591269029))*cos(log(t - sin(t) - cos(0.986122425667481*t - 1.48349793979224*x)))

ACC: 0.5806

NMSE: 0.00875670136085

Role: equation author

Formula: -0.0318665366944*v - 2.3478917222*x**2 - 5.03457658153*x + 0.298050820308*sin(0.999980436038*t)

ACC: 0.858

NMSE: 0.00150428390217

Role: candidate decider

Formula: (0.0005796547 - (v*v*(-(x^3)/(x + 0.25876948) + (x + 0.021030162)*0.36042473) - (x*0.054585215 - (x^3))))*(-135.58333)

ACC: 0.328

NMSE: 0.132180763971

Problem: stressstrain

Ground truth: (A + B*strain^n)*(1 - ((temp-Tr)/(Tm-Tr))^m)

Role: search controller

Formula: -0.0582393 + exp(-temp**2*exp(-(strain * (1 - temp))**2)*exp((strain^2)))*exp(-(0.5816915 - exp(-exp(2*strain*(strain * (1 - temp)) - 2*temp)) + 1.5569849*exp(-67.5881248690533*strain))**2)

ACC: 0.920529801325

NMSE: 0.0153913678069

Role: llmsr

Formula: np.clip(np.where(np.abs(strain) <= (-1.9159115 + -0.88395036 * (temp - 3.5133818)) / (32.89433 + -0.91930236 * (temp - 3.5133818)), (32.89433 + -0.91930236 * (temp - 3.5133818)) * strain, np.sign(strain) * (-1.9159115 + -0.88395036 * (temp - 3.5133818) + (-0.82158661 + -0.21471593 * (temp - 3.5133818)) * (np.abs(strain) - (-1.9159115 + -0.88395036 * (temp - 3.5133818)) / (32.89433 + -0.91930236 * (temp - 3.5133818))))), -0.85056205, 0.85056205)

ACC: 0.813975011568718

NMSE: 0.0210373011174167

Role: drsr

Formula: ((35.9883662212) * strain * (1 + (-0.0115129087436) * temp + (0.00218407613697) * temp**2) + (-35.643631154) * np.maximum(strain - ((0.0231713232655) + (0.0016443854171) * temp + (-

0.018145977602) * temp**2), 0.0) + (-0.360595550616) * np.maximum(strain - ((0.0231713232655) + (0.0016443854171) * temp + (-0.018145977602) * temp**2), 0.0)**2 + (0.000539209353519) + (0.0151926215581) * temp)

ACC: 0.755668671911152

NMSE: 0.0179449135777797

Role: pysr

Formula: sin(exp(tanh(sin(strain))))*sin(exp(re(0.0019494138**tanh(strain))))*cos(temp*Abs(sin(exp(sin(tanh(strain)))))/0.73116386)

ACC: 0.70319001387

NMSE: 0.0906428680881

Role: equation author

Formula: (1.08003035643) + (-0.952860297193)*strain*temp + (0.128259304369)*log(strain + (0.0007392910429)) + (-0.315858818736)*temp**2

ACC: 0.492

NMSE: 0.0903257425823

Role: candidate decider

Formula: 0.86867570081697*exp(-2*Abs(exp(-153.95023*Abs(strain))/temp**0.34385886 + temp**2.7684977*((strain**2*((strain^2) - 1*(-0.5469036))**2*exp(-Abs((strain * (1 - temp)) - 1*0.094919465)))**((strain^2) - 1*(-0.44512773)) - 1*(-0.50284195))))

ACC: 0.838

NMSE: 0.00907740710997

Dataset group: battery

--------------------------------------------------------------------------------

Problem: battery_cycle_life_log_zscore

Ground truth: N/A

Role: search controller

Formula: 0.18800634*delta_q_100_10_min + 0.08823834*Abs(delta_q_100_10_min - 0.3875072) + 2.7426913

ACC: 1

NMSE: 0.169492688492

Role: llmsr

Formula: 2.7939691 - (0.009369131 * delta_q_100_10_min + 0.11576148 * delta_q_100_10_var + (-0.0014997437 * capacity_fade_slope_2_100 + 0.048894435 * capacity_fade_intercept_2_100) + -

0.081122356 * discharge_capacity_cycle_2 + 0.03383006 * temp_integral_2_100 + 0.018372048 * min_ir_2_100 + 0.017604721 * ir_diff_100_2 + -0.031506652 * avg_charge_time_first_5)

ACC: 1

NMSE: 0.0346581964098319

Role: drsr

Formula: ((0.00150675242993) * capacity_fade_slope_2_100 + (-0.0489142351767) * capacity_fade_intercept_2_100 + (0.0811423116135) * discharge_capacity_cycle_2 + (0.0315088619397) * avg_charge_time_first_5 + (-0.0338293499066) * temp_integral_2_100 + (-0.018374757752) * min_ir_2_100 + (-0.0176064605374) * ir_diff_100_2 + (-0.00938111727564) * delta_q_100_10_min + (-0.115764199563) * delta_q_100_10_var + (2.79396979625))

ACC: 1

NMSE: 0.0346581960281725

Role: pysr

Formula: (1.5561504 - (delta_q_100_10_var - (-temp_integral_2_100 + sin(delta_q_100_10_min - temp_integral_2_100 - (delta_q_100_10_min - (-temp_integral_2_100 + ir_diff_100_2 - sin(ir_diff_100_2*cos(ir_diff_100_2))))))))*0.025488654*(delta_q_100_10_min - 1*(-1.945337)) + 2.6897624

ACC: 1

NMSE: 0.114347693146

Token usage: 0

Role: equation author

Formula: 0.500244570077*exp(0.319825440139*delta_q_100_10_min) + 2.27362903988

ACC: 1

NMSE: 0.152638377207

Role: candidate decider

Formula: Abs(delta_q_100_10_min*((capacity_fade_slope_2_100 * discharge_capacity_cycle_2)**3*exp(-3.4*exp(-3.4*Abs((capacity_fade_slope_2_100 * discharge_capacity_cycle_2)**3))*Abs((capacity_fade_slope_2_100 * discharge_capacity_cycle_2)**3))*exp(-3.4*Abs((capacity_fade_slope_2_100 * discharge_capacity_cycle_2)**3))) + ((delta_q_100_10_min*delta_q_100_10_min*(delta_q_100_10_min - 0.5631909) + (capacity_fade_slope_2_100 * discharge_capacity_cycle_2))*exp(-3.4*Abs(delta_q_100_10_min*delta_q_100_10_min*(delta_q_100_10_min - 0.5631909) + (capacity_fade_slope_2_100 * discharge_capacity_cycle_2))))*1.1395235 + exp(delta_q_100_10_min)*0.092722535) + 2.6687772

ACC: 1

NMSE: 0.235584506435

**Supplementary Note 5.** Detailed prompt templates in LLM-PySR

Agent A (Search controller)

Prompt template:

Task: You are LLM-A in a symbolic-regression loop.

Context: {context or "Unknown physical background. Use variable meanings + data patterns only."}

Variable meanings: {json.dumps(variable_desc, ensure_ascii=False)}

Data snapshot (first up to 100 rows): {json.dumps(data_snapshot if isinstance(data_snapshot, dict) else {}, ensure_ascii=False)}

Experience memory: {json.dumps(memory, ensure_ascii=False)}

Previous round reviews_with_formula: {json.dumps(reviews_with_formula or [], ensure_ascii=False)}

LLM-C recommended maxdepth (previous round): {llm_c_recommended_maxdepth if llm_c_recommended_maxdepth is not None else "null"}

Allowed base operators: {global_operators}

Return strict JSON with keys:

- engineered_features: [{"name":"u1","expression":"...","reason":"...","dimension_check":"pass/unknown"}]

- selected_input_features: ["x1","x2","u1"]

- selected_operators: ["+","*","log", "..."]  # choose from allowed_base_operators

- custom_unary_operators: {"f1":"..."}

- custom_binary_operators: {"f2":"..."}

- recommended_maxdepth: <int_or_null>

- decision_evidence: {

    "operators": [{"op":"sin","source":"pattern|memory|context","evidence":"..."}],

    "features": [{"name":"u1","source":"pattern|memory|context","evidence":"..."}]

  }

Rules:

1) Prefer only high-confidence engineered features.

2) A custom operator should combine >=2 base operators.

3) Do not exceed {max_custom_ops} custom operators total.

4) Only use operators from allowed list in expressions.

4.1) Custom operator constraints:

  - Unary custom operator must be expression over x only (form f(x)).

- Binary custom operator must be expression over x,y only (form f(x,y)).

- Do not reference dataset variable names (u1, v, E_n, etc.) inside custom operator definitions.

- Do not call custom operators inside custom operator definitions (no f1(...), f2(...)).

- Do NOT output assignments like "f1(x)+x" as an operator definition. The definition itself must be only the RHS expression.

- Valid examples:

  unary: "exp(-abs(x))", "sin(x) + x^2"

  binary: "x*log(1+abs(y))", "sin(x) + cos(y)"

- Invalid examples:

  "f1(x)+x", "f2(u1, x)", "x + z", "f3(x)"

- If uncertain, output empty custom operators: {}.

5) selected_input_features are the final variables to pass to PySR. Keep them minimal.

5.2) selected_input_features must contain at most 5 variables total.

  If uncertain, choose fewer variables instead of adding weak ones.

5.1) Feature-engineering budget:

- Hard limit: at most 3 engineered features.

- Output only the top-3 most important engineered features ranked by expected impact on PySR fit quality.

- If you are uncertain, output fewer features instead of filling to 3.

- Start from atomic/low-granularity features first (pairwise distance, ratio, product, square).

- Avoid directly constructing an almost-complete final law in one step.

6) Use Experience memory as hard guidance:

- Prioritize operators marked effective.

- Consider expanding to suggested operators when current fit is poor.

- Avoid operators marked ineffective unless there is clear new evidence.

- Prefer recommended feature-engineering patterns and custom unary/binary operators.

7) Analyze internal relation patterns directly from the provided first-100-row data snapshot.

8) For selected_operators: prefer a stable compact set.

  If memory already supports prior operators, change at most 1-2 operators this round.

9) Dimension-check policy:

- Do NOT output "fail". Use only "pass" or "unknown".

- If variable meanings do NOT provide explicit physical units, prefer "unknown".

- For sin/cos/tan/log/exp on normalized or unit-unknown variables, use "pass" or "unknown".

10) selected_operators must be a concise subset of allowed_base_operators.

11) Prefer operators that are repeatedly supported by data_snapshot and experience_memory.

11.1) Never add "re"/"im" as operators. They are exporter wrappers and should not increase complexity.

12) Progressive update policy:

- Keep high-performing engineered features/operators from prior rounds unless strong contradictory evidence appears.

- Avoid replacing many features/operators at once.

13) Single-plan policy:

- Return only one final plan.

- Do not output conservative_plan or expansion_plan.

14) Every selected operator/engineered feature should have a traceable evidence source in decision_evidence.

15) Decide recommended_maxdepth in [6, 40]. Prefer LLM-C recommendation unless strong contrary evidence appears.

Agent B (Candidate reviewer)

Prompt template:

Task: You are LLM-B. Analyze and score candidate symbolic formulas.

Context: {context or "Unknown background"}

Variable meanings: {json.dumps(variable_desc, ensure_ascii=False)}

Current best score: {current_best_score}

Current best metrics: {json.dumps(current_best_metrics, ensure_ascii=False)}

Candidates:

{json.dumps(candidates, ensure_ascii=False)}

Return strict JSON object with this schema (no extra text):

{

"reviews": [

{

"index": <int>,

"subscores": {

"interpretability": <float_0_to_1>,

"acc_quality": <float_0_to_1>,

"nmse_quality": <float_0_to_1>,

"pysr_metrics_quality": <float_0_to_1>,

"fit_quality": <float_0_to_1>,

"simplicity": <float_0_to_1>,

"variable_semantics_consistency": <float_0_to_1>

},

"score": <float_0_to_1>,

"label": "positive|negative",

"reason": "<short>"

}

]

}

Rules:

1) Score each candidate directly; do not hard-reject whole formulas by a global dimension gate.

2) Metric meanings:

- ACC: higher is better (accuracy under tolerance).
- val_acc: higher is better; treat this as the primary reliability signal for final score.
- NMSE: lower is better (normalized mean square error).
- loss (from PySR): lower is better.
- sr_score (from PySR): higher is better, but only a secondary tie-breaker.

3) Score based on:

- physical interpretability,
- ACC/val_acc value (most important): larger is better,
- NMSE value: smaller is better,
- PySR metrics (loss lower; use sr_score only when other metrics are close).

4) Reason must explicitly mention ACC/NMSE/loss/sr_score and why the score is assigned.

5) The final positive/negative label is determined by system-side deterministic gates.

You still need to return a reasonable label and score, but the system may override label.

6) In comparison reasoning, prioritize val_acc/ACC first, then NMSE/loss, then interpretability.

Use sr_score only for tie-breaking when the above metrics are close.

7) Score direction must follow metric direction:

- if ACC/val_acc are higher while NMSE and loss are lower, assign a higher score.
- if ACC/val_acc are lower while NMSE and loss are higher, assign a lower score.

8) Strict anti-overrating guard (must follow):

- Let reliability = val_acc when available, otherwise ACC.
- If reliability < 0.10, score must be <= 0.45.
- If 0.10 <= reliability < 0.20, score must be <= 0.55.
- If 0.20 <= reliability < 0.40, score must be <= 0.65.
- If 0.40 <= reliability < 0.60, score must be <= 0.75.
- If 0.60 <= reliability < 0.80, score must be <= 0.85.
- If 0.80 <= reliability < 0.90, score must be <= 0.92.
- If reliability >= 0.90, score can be in [0.0, 1.0] according to other metrics.

9) High-score gate:

- score > 0.85 is only allowed when BOTH ACC >= 0.60 and val_acc >= 0.60 (or ACC >= 0.60 if val_acc missing).

10) Never give high score to low-quality fit:

- If reliability < 0.60 and (NMSE is large or loss is large relative to better candidates),
  keep score conservative (typically <= 0.75).

Agent C (Experience reflector)

Prompt template:

Task: You are LLM-C. Produce concise, correctable experience for the next LLM-A round.

Context: {context or "Unknown physical background"}

Variable meanings: {json.dumps(variable_desc, ensure_ascii=False)}

Candidate briefs from this iteration:

{json.dumps(candidate_briefs or [], ensure_ascii=False)}

Strong-evidence channel:

{json.dumps(strong_pos, ensure_ascii=False)}

Weak-evidence channel:

{json.dumps(weak_channel, ensure_ascii=False)}

Return strict JSON:

```
{
  "memory": [],
  "guidance": {
    "good_feature_combinations": ["..."],
    "bad_feature_combinations": ["..."],
```

"good_operators": ["..."],

"bad_operators": ["..."],

"recommended_maxdepth": <int_or_null>

}

}

Rules:

1) Keep guidance concise and high precision. Only output these 4 categories.

2) good_feature_combinations: at most 2 combinations.

3) Use strong channel as hard evidence; weak channel only as tentative hints.

4) Experience can be wrong. You must revise previous conclusions if current evidence contradicts them.

5) If no reliable evidence exists for an item, do not output it.

5.1) Treat re(expr)/im(expr) as exporter wrappers (real/imag part), not as extra operator complexity.

5.2) good_operators and bad_operators must be operator names from this exact set only:

{json.dumps(GLOBAL_BASE_OPERATORS, ensure_ascii=False)}.

Do NOT output natural-language phrases such as "avoid division_by_xxx".

5.3) good_feature_combinations and bad_feature_combinations must be expression snippets over known variables only.

Do NOT output plain-language advice in these fields.

6) If evidence is enough, set guidance.recommended_maxdepth in [6, 40].

6) Previous good combo memory (may contain stale ideas):

{json.dumps(prior_combos, ensure_ascii=False)}